\documentclass{article}

\usepackage[final,main]{neurips2026}
\setcitestyle{numbers,square}

\usepackage[utf8]{inputenc} 
\usepackage[T1]{fontenc}    
\usepackage{hyperref}       
\usepackage{url}            
\usepackage{booktabs}       
\usepackage{amsfonts}       
\usepackage{nicefrac}       
\usepackage{microtype}      
\usepackage{xcolor}         

\usepackage{amsmath,amsfonts,bm}

\def\eqref#1{equation~\ref{#1}}

\def\1{\bm{1}}

\DeclareMathAlphabet{\mathsfit}{\encodingdefault}{\sfdefault}{m}{sl}
\SetMathAlphabet{\mathsfit}{bold}{\encodingdefault}{\sfdefault}{bx}{n}

\newcommand{\E}{\mathbb{E}}

\newcommand{\R}{\mathbb{R}}

\DeclareMathOperator*{\argmin}{arg\,min}

\usepackage{hyperref}
\usepackage{url}

\usepackage{subcaption}
\usepackage{graphicx}
\usepackage{algorithm, algorithmic}
\usepackage{amsthm, amssymb}
\usepackage{enumitem}
\usepackage{mathtools}

\newtheorem{lemma}{Lemma}

\definecolor{colin}{RGB}{128,0,128}

\newcommand{\passat}[1]{\mathrm{pass}\text{@}#1}
\newcommand{\passiat}[1]{\mathrm{pass}_i\text{@}#1}
\newcommand{\passhat}[1]{\widehat{\mathrm{pass}_i@#1}}
\newcommand{\passdat}[1]{\textrm{pass}_\mathcal{D}@#1}
\newcommand{\dto}{\overset{d}{\to}}
\newcommand{\N}{\mathcal{N}}

\title{Efficient Prediction of pass@$k$ Scaling \\ in Large Language Models}

\author{
\textbf{Joshua Kazdan}\textsuperscript{1$\dagger$},
\textbf{Colin Sullivan}\textsuperscript{1$\dagger$}\thanks{Correspondence to: \url{jkazdan@stanford.edu}, \url{cpsullivan@cs.stanford.edu}, \url{sanmi@cs.stanford.edu}.},\\
\textbf{Rylan Schaeffer}\textsuperscript{1$\ddagger$},
\textbf{Youssef Allouah}\textsuperscript{1$\ddagger$},\\
\textbf{Kyssen Yu}\textsuperscript{2},
\textbf{Noam Levi}\textsuperscript{3},
\textbf{Sanmi Koyejo}\textsuperscript{1} \\
\textsuperscript{1}Stanford University \quad
\textsuperscript{2}University of Toronto \quad \textsuperscript{3} EPFL \\
\textsuperscript{$\dagger$}Equally contributing first author.
\textsuperscript{$\ddagger$} Equally contributing second author.
}

\begin{document}

\maketitle

\begin{abstract}
Assessing the capabilities and risks of frontier AI systems is a critical area of research; recent work has shown that repeated sampling can dramatically increase both.
For instance, repeated sampling allows models to solve difficult math and coding problems, but it may also increase their risk of being jailbroken.
Such results raise a crucial question:
how can one accurately predict a model's behavior when scaled to a massive number of attempts, given a vastly smaller sampling budget?
This question is directly relevant to model providers, who serve hundreds of millions of users daily, and to governmental regulators, who seek to prevent harms.
To answer this question, we make three contributions.
First, we find that standard methods for fitting these laws suffer from statistical shortcomings that hinder predictions, especially in data-limited scenarios.
Second, we remedy these shortcomings by introducing a robust estimation framework, which uses a beta-binomial distribution to generate more accurate predictions from limited data.
Third, we propose a dynamic sampling strategy that allocates a greater budget to harder problems.
Combined, these innovations enable more reliable prediction of rare risks and capabilities at a fraction of the computational cost.

\end{abstract}

\section{Introduction}

Prompt-based attacks against frontier (multimodal) AI systems often fail when attempted only once \citep{anil2024manyshot,panfilov2025capabilitybasedscalinglawsllm, howe2025scalingtrendslanguagemodel,kazdan2025nocourseican}. 
Likewise, many hard math \citep{glazer2024frontiermathbenchmarkevaluatingadvanced} and software engineering \citep{jimenez2024swe} tasks are too difficult for models to solve reliably on the first attempt. 
Through repeated attempts, however, the success rate of these models can climb rapidly to near-100\% \citep{brown2024largelanguagemonkeysscaling, hughes2024bestofnjailbreaking, kwok2025robomonkeyscalingtesttimesampling}.
Consequently, predicting changes in capabilities and/or risks when a user is allowed many attempts to accomplish a task has become an important problem for companies, researchers, and governmental regulators alike.
The relevance of this problem is only underscored by the massive scale at which these frontier AI systems are deployed, with some experiencing billions of daily interactions.
However, making such predictions is challenging because sampling from language models at such scale can be prohibitively expensive.
How can one predict the behavior of frontier AI systems in this repeated attempts regime using only a limited number of samples?

In this work, we approach this problem through estimation of the widely used $\passat{k}$ metric ~\citep{kulal2019spoc, chen2021evaluatinglargelanguagemodels}, which measures the expected pass rate given $k$ attempts at solving each problem, where a problem is solved if any attempt is successful.
Unfortunately, direct estimation at high $k$ is often difficult.
While prior work has shown that $\passat{k}$ can follow predictable power laws across a range of domains including jailbreaking, mathematical problem-solving, and code generation~\citep{hughes2024bestofnjailbreaking, brown2024largelanguagemonkeysscaling,du2024evaluatingcode}, we find that standard methods for fitting these laws~\citep{chen2021evaluatinglargelanguagemodels, brown2024largelanguagemonkeysscaling,hughes2024bestofnjailbreaking} suffer from statistical shortcomings that hinder predictive accuracy, especially in data-limited scenarios.

We argue that the shortcomings of prior prediction methods stem from statistical approximations that do not hold in sample-limited regimes.
By carefully modeling the data-generating process and developing faithful estimators, we demonstrate that predictions can be substantially improved.

\begin{figure*}
    \centering
    \begin{subfigure}[t]{0.65\linewidth}
        \centering
        \includegraphics[width=\linewidth]{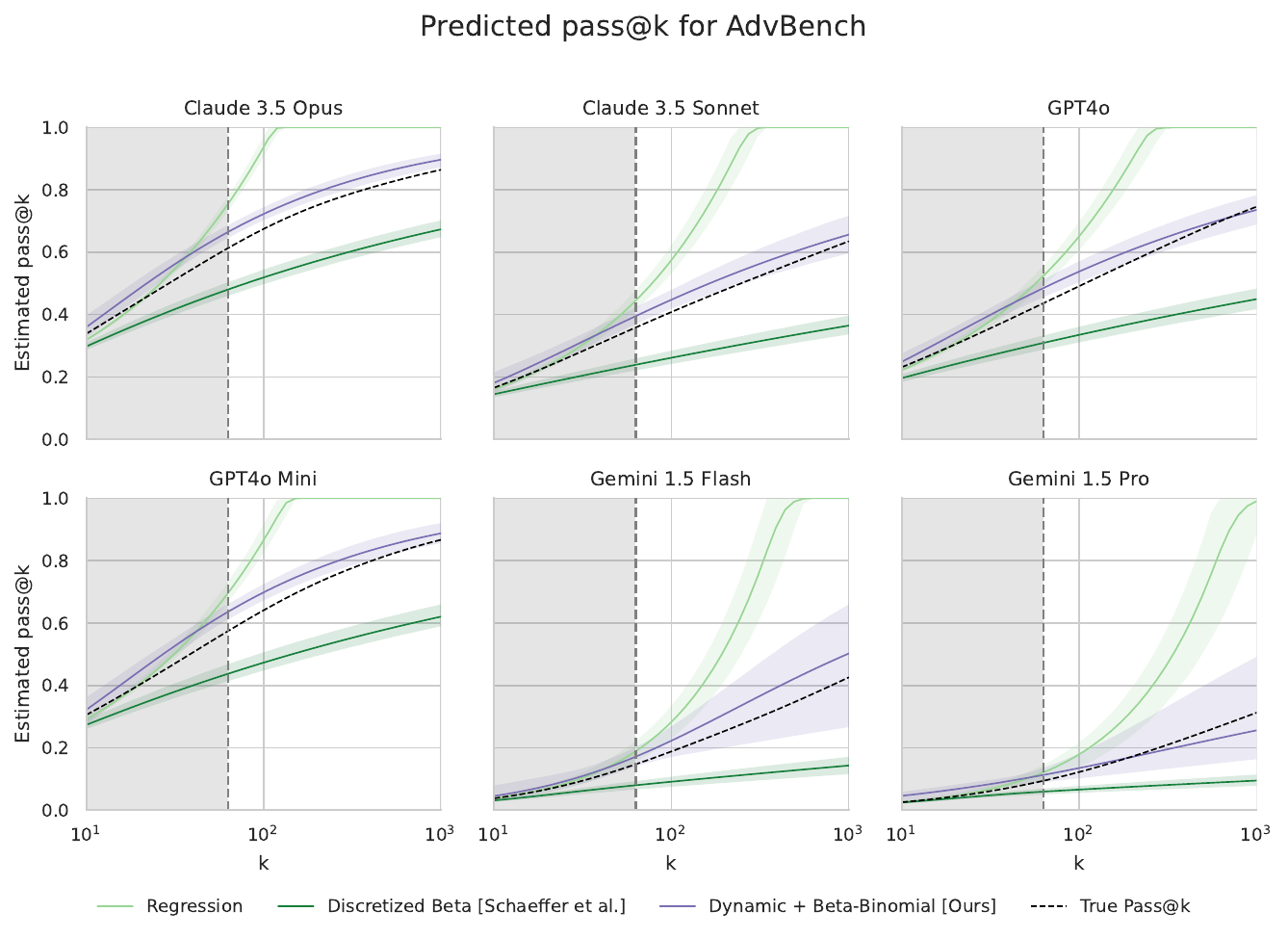}
        \label{fig:jailbreaking_results}
    \end{subfigure}\\[-2em]
    \begin{subfigure}[t]{0.65\linewidth}
        \centering
        \includegraphics[width=\linewidth]{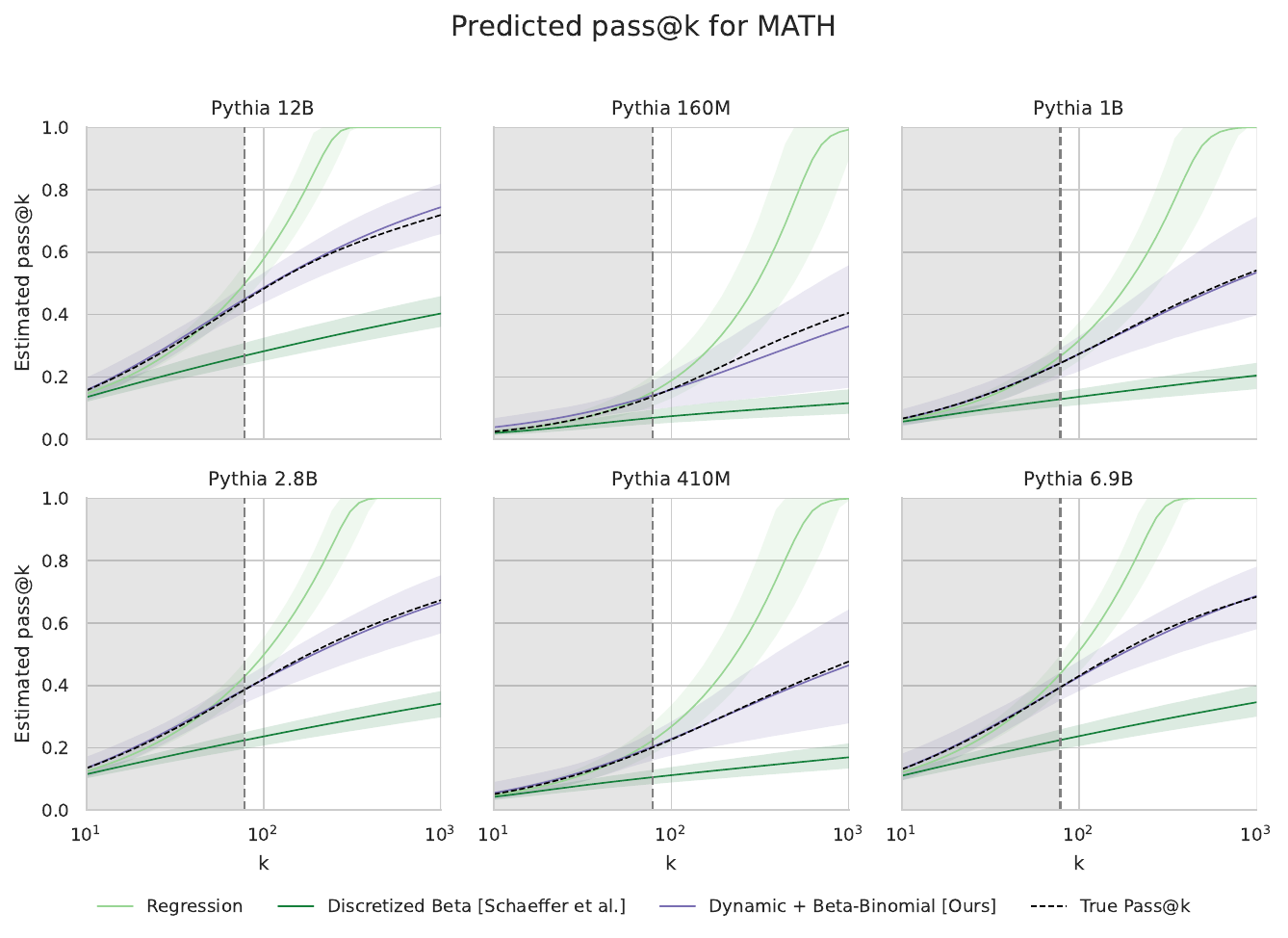}
        \label{fig:math_results}
    \end{subfigure}\\[-2em]
    \begin{subfigure}[t]{0.65\linewidth}
        \centering
        \includegraphics[width=\linewidth]{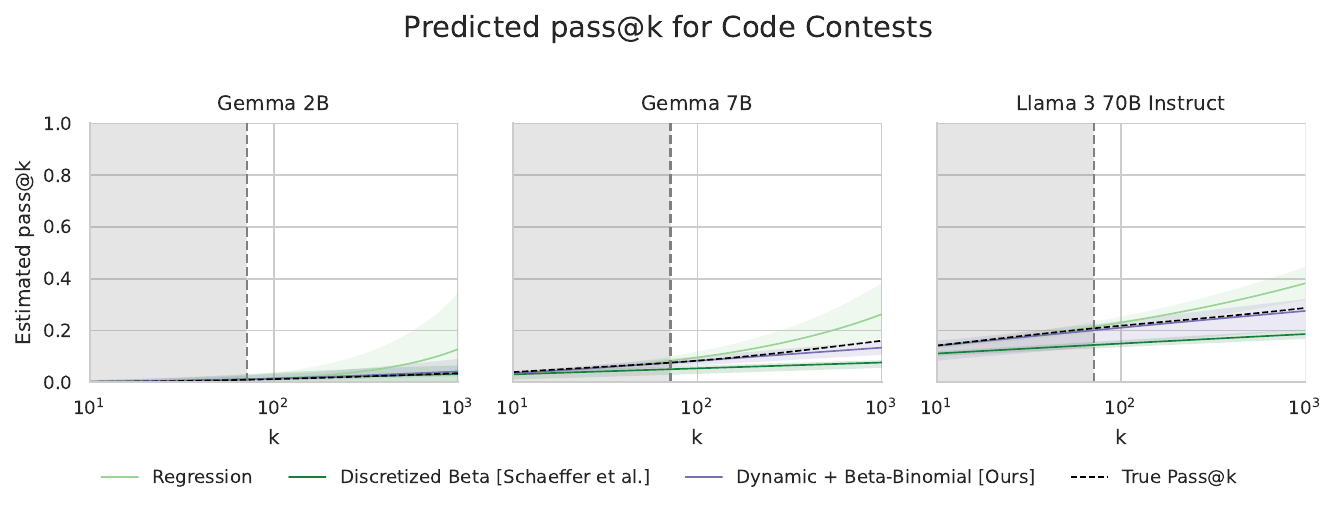}
        \label{fig:code_contests_results}
    \end{subfigure}
    \caption{\textbf{Comparing Forecasting Methods for $\mathrm{pass}_{\mathcal{D}}\text{@}k$ Across Different Datasets.}
    The ground truth is computed based on $10\,000$ actual samples per problem.
    All predictive models are trained on data from a budget of $10\,000$ total samples.
    \textbf{The gray region} shows $k$ for which $\passat{k}$ can be directly estimated given the available budget, while \textbf{the white region} shows $k$ for which the $\passat{k}$ must be extrapolated given the budget.
    Our estimator tracks the ground truth far better than prior methods.
    Error bars represent a bootstrap-averaged 95\% percentile range of estimates.
    }
    \label{fig:prediction_results}
\end{figure*}

\subsection{Contributions}

To address the challenge of efficient prediction, this paper makes the following contributions:
\begin{enumerate}[leftmargin=6mm]
    \item \textbf{Rigorous critique of prior prediction methods.} We discuss statistical flaws that have led to poor prediction accuracy in approaches such as log-log-linear regression and existing distributional fitting techniques.
    \item \textbf{Robust estimation framework for prediction.} We remedy the shortfalls of previous methods by employing a more suitable distributional model---the beta-binomial---and deriving an improved predictor for $\passat{k}$ that faithfully accounts for the data generating process in order to deliver more accurate predictions.
    \item \textbf{Efficient dynamic sampling strategy.} We show empirically that by allocating our fixed compute budget adaptively to focus on more difficult problems, we achieve more accurate predictions than the standard approach of uniform sampling.
\end{enumerate}

The insights from this work are important for both AI safety and capabilities research.
For AI safety, reliable forecasts for the scaling of vulnerability rates is crucial for assessing the societal risk posed by models deployed to millions of users.
For capabilities, such predictions are vital for efficiently applying methods like Reinforcement Learning from Verified Rewards, where training on difficult problems requires correctly sizing batches to ensure non-zero success rates.
Thus, efficient prediction for $\passat{k}$ scaling represents a critical step toward developing aligned and powerful AI systems.

Code for reproducing our experiments is available on GitHub.\footnote{\url{https://github.com/jlkazdan/monkeys_revisited}}
\section{Problem Statement: Efficient Prediction of Rare Model Behaviors from Repeated Sampling}
\label{sec:problem_statement}

We consider the performance of AI systems on some problem, defined as a set of prompts with verifiable binary outcomes: each attempt either produces the (un)desirable outcome for that prompt, or does not.
For example, we may want our AI system to solve a Millennium Problem, or to not launch a cyberattack on a nation's infrastructure.
Our goal is to predict the success rate of an AI system, given many repeated attempts at the problem.
To quantitatively measure the system's behavior, we use the widely-adopted $\passat{k}$ metric \citep{kulal2019spoc, chen2021evaluatinglargelanguagemodels}: For a single prompt, indexed by $i$, from a distribution of prompts $\mathcal{D}$, let $\passiat{1}$ be the model's true probability of success in one attempt. The probability of achieving at least one success in $k$ attempts is then $\passiat{k}$:

\begin{equation}
    \passiat{k} = 1 - (1-\passiat{1})^k.
\end{equation}
For the entire dataset $\mathcal{D}$ of $m$ problems, the overall pass rate $\passdat{k}$ is the expected fraction of problems solved within $k$ attempts:

\begin{equation}
    \passdat{k} = \mathbb{E}_{i \sim \mathcal{D}} [\passiat{k}].
\end{equation}

Our goal is to predict performance given many attempts using data from an economically feasible, small-scale experiment.
This leads to our formal research question:

\begin{quote}
    {\em
    Given a compute budget of $B$ samples to be distributed across a dataset $\mathcal{D}$ containing $m$ problems, how should one allocate samples to build a model that predicts
    $\passdat{k}$ for $k \gg B/m$?
    }
\end{quote}

In this work, we use a small budget (e.g., $B/m \in [10^0, 10^2]$) to predict performance for $\passat{k}$ at large scale (e.g., $k \in [10^1, 10^3]$).
We evaluate predictions by comparing them to a ground truth estimate of $\passat{k}$ computed using a withheld dataset of $10\,000$ samples per problem.
To evaluate performance, we compute mean squared error (MSE) relative to the ground truth $\passat{k}$ values.

The product of our contributions is an estimator that provides consistently more accurate predictions than existing methods (see Figure \ref{fig:prediction_results}).

\section{Critiquing past methods of predicting
\texorpdfstring{\boldmath$\passat{k}$\unboldmath}{pass@k}}

We now examine past methods of predicting $\passat{k}$ scaling and identify their shortcomings.

\subsection{Combinatorial Estimation}

Directly measuring pass$_{\mathcal{D}}\text{@}k$ for a large $k$ is often computationally expensive.
While unbiased estimators exist, such as that of \cite{chen2021evaluatinglargelanguagemodels}, they are only defined when the number of samples taken for each problem is greater than or equal to the number of attempts $k$.
Given $b_i$ samples on problem $i$ with $s_i$ successes, this estimator is:
\begin{equation}\label{pass_i_at_k_estimate}
    \widehat{\passiat{k}} = 1-\frac{\binom{b_i-s_i}{k}}{\binom{b_i}{k}}.
\end{equation}

In this paper, we focus on the regime where $B/m < k < B$.
As the size and quantity of benchmarks continues to grow, we may often find ourselves in such constrained contexts.
Here, given that $k > B/m$, we cannot allocate the required minimum of $k$ samples for each of $m$ problems.
This means the standard unbiased estimator (Equation~\ref{pass_i_at_k_estimate}) cannot be directly applied, so we must instead rely on extrapolation and predictive modeling.

\subsection{Linear Regression} \label{regression_issues}

The first and most common extrapolation of $\passat{k}$ uses linear regression \citep{ brown2024largelanguagemonkeysscaling, hughes2024bestofnjailbreaking}.
Specifically, given $b$ samples per problem, one first estimates $\passdat{k}$ for $k$ between $1$ and $b$ and then fits a least squares regression of the form:
\begin{equation}
    \label{least_squares_relation}-\log(\passdat{k}) \sim a\log(k) + c.
\end{equation}
Fixing $C=e^{-c}$ corresponds to the power law: 
\begin{equation}\label{power_law}
    \passdat{k} \sim C \cdot k^{-a}.
\end{equation}
Explicitly, the regression loss takes the form:
\begin{equation}
    \label{least_squares_loss}
    \frac{1}{b}\sum_{k=1}^b \left(-\log \left(\widehat{\passdat{k}}\right) - a\log(k) - c\right)^2.
\end{equation} 

There are several problems with this approach, leading to poor estimates of $\passat{k}$ for high $k$ values (see Figure \ref{fig:prediction_results}):
\begin{enumerate}[leftmargin=6mm]
    \item Estimates of $\passdat{k}$ are not independent for different $k$ when computed using the same dataset of samples.
    \item Estimates of $\passdat{k}$ are not homoskedastic, i.e. they have different variances for each value of $k$.
    \item $\passat{k}$ may not follow a power law for some datasets.
    \item Power laws typically apply only for large values of $k$. Therefore, if the computation budget for sampling is limited, non-leading terms can dominate, resulting in poor fits of the data.
\end{enumerate}

As a concrete example of the fourth point, suppose that
\begin{equation}
    1 - \mathrm{pass}\text{@}k = \frac{A}{k^{\alpha}} + \frac{B}{k^{\beta}}
    \label{example_power_law}
\end{equation}
where $A \gg B$ but $\alpha > \beta$. For small values of $k$, the first term of Equation~\ref{example_power_law} dominates. However, for large values of $k$, the second term, which supplies the true asymptotic power law, dominates. If we lack a sufficient budget to observe samples for large $k$, then least squares will incorrectly fit to the first term. We quantify statements 1 and 2 more precisely with proofs in Appendix~\ref{regression_proofs}.

Our work directly remedies these issues by moving away from regression on aggregate statistics, instead modeling the underlying distribution of problem difficulties.

\subsection{Discretized-Beta Distributional Fitting} \label{rylan_distributional}

\cite{schaeffer2025largelanguagemonkeyspower} use a variant of empirical Bayes to estimate $\passat{k}$ for high $k$.
To describe their method, we first introduce some notation.
As before, let $\mathcal{D}$ denote a data set of questions.
Define $\mathcal{U}$ to be the distribution of per-problem success probabilities $\passiat{1}$ for $i\in \mathcal{D}$:

\begin{equation}
    \passiat{1} \sim \mathcal{U}, \quad i\in \mathcal{D}.
\end{equation}

For the $i$-th question in our dataset, we observe $b$ samples, of which we say that $s_i$ are successful. 
\cite{schaeffer2025largelanguagemonkeyspower} fit scaled beta distributions to $\passhat{1} = \frac{s_i}{b}$ and leveraged this distribution to estimate $\passat{k}$:

\paragraph{Step 1: Fit the scale $\theta$.} Recall the probability density function of a scaled beta distribution:
\begin{equation}
    \label{scaled_beta} \textrm{Beta}(p; \alpha, \beta, \theta) = \frac{1}{\mathrm{Be}(\alpha, \beta)} \left(\frac{p}{\theta}\right)^{\alpha-1} \left(1-\frac{p}{\theta}\right)^{\beta-1}\frac{1}{\theta},
\end{equation}

\cite{schaeffer2025largelanguagemonkeyspower} provide the following estimate for the scale parameter $\theta$:

\begin{equation} \label{scale_estimator} 
\hat{\theta} = \frac{b+1}{b}\textrm{max}_{i\in \mathcal{D}} \left(\widehat{\passiat{1}}\right).\end{equation}  
They use this estimator because it resembles the uniformly minimum variance unbiased estimator (UMVUE) for the parameter $B$ of a uniform distribution $\mathrm{Uniform}(0, B)$ \cite{l1983theory}.
Unfortunately, the scaled beta distribution is not an exponential family distribution.
In particular, the UMVUE for $\theta$ in a scaled beta distribution is unknown.
As such, this is not a principled estimator for $\theta$.
We provide details for how to estimate $\theta$ using a stabilized MLE in Appendix \ref{more_flexible}, but we find empirically that using the scale parameter does not improve predictions.

\paragraph{Step 2: Fit $\alpha$ and $\beta$ by discretizing.} \cite{schaeffer2025largelanguagemonkeyspower} first divide the interval $(0, 1)$ into log-scale bins with endpoints $0=e_0, e_1, ..., e_\ell=1$, where the bin widths decrease ($e_{i} - e_{i-1} > e_{i+1} - e_i$).
They then numerically compute the probability mass in each bin and fit $\alpha$ and $\beta$ by maximizing the multinomial likelihood over the number of problems whose estimated success rate falls into each bin.  Specifically, if we assign the estimated probability: \begin{equation} A_i(\alpha, \beta, \theta):= \int_{e_i}^{e_{i+1}} \textrm{Beta}(p; \alpha, \beta, \theta)dp,\end{equation} then \cite{schaeffer2025largelanguagemonkeyspower} fit $\alpha$ and $\beta$ by optimizing

\begin{align}
\argmin_{\alpha, \beta} \; & 
- \log \Biggl( 
    \prod_{i=1}^\ell 
    A_i(\alpha, \beta, \theta)^{\sum_{j=1}^{m} 
        \mathbf{1} \{\widehat{\textrm{pass}@1} \in [e_i, e_{i+1})\}
    } 
\Biggr) \\
= \argmin_{\alpha, \beta} \; & 
- \sum_{i=1}^\ell 
    \Biggl(
        \sum_{j=1}^{m} \mathbf{1} \{\widehat{\textrm{pass}@1} \in [e_i, e_{i+1})\}
    \Biggr) 
    \log \bigl( A_i(\alpha, \beta, \theta) \bigr)
\end{align}

This complex discretized beta estimator was used to support the common case when $s_i = 0$.
The estimate $\widehat{\mathrm{pass}_i\text{@}1}$ is also $0$, so the scaled beta density is not supported.

\paragraph{Step 3: Predict $\passat{k}$} \cite{schaeffer2025largelanguagemonkeyspower} use the fit distribution to approximate the asymptotic slope of the $\passat{k}$ scaling curve and do not attempt to extrapolate $\passat{k}$ beyond the provided number of trials.
To extend this approach to the high-$k$ regime, we take the expectation over the success probability $\passiat{1} \sim \mathrm{Beta}(\hat{\alpha}, \hat{\beta}, \hat{\theta})$:

\begin{equation}
    \widehat{\passdat{k}} = \E_{\passiat{1} \sim \mathrm{Beta}(\hat{\alpha}, \hat{\beta}, \hat{\theta})} \left[ 1 - (1 - \passiat{1})^k \right].
\end{equation}

\paragraph{Analysis of the Discretized-Beta Estimator}
Because the bins are wider for smaller values, this fitting method consistently produces \textbf{downward-biased} estimates of the distribution $\mathcal{U}$.
We demonstrate this phenomenon in Figure~\ref{fig:beta-fit} where the discretized beta distribution is fit on problem success probabilities drawn from a uniform distribution.
The fit is visibly skewed, incorrectly up-weighting the left tail of the distribution.
\section{Better Estimation of pass@k} \label{better_estimator}

In this section, we develop a novel predictor of $\passdat{k}$, inspired by \cite{levi2024simplemodelinferencescaling} that improves predictive accuracy for large $k$. 
As shown in Figure \ref{fig:math_heatmap}, our method provides equivalent or better estimates across all models, values of $k$, and sampling budgets tested.
We no longer assume a fixed sampling budget per question: $b_i$ denotes the budget for the $i$-th question.
Our improvements involve two steps: 
\begin{enumerate}[leftmargin=6mm]
\item We develop an alternative distributional fitting method for the problem-difficulty distribution $\mathcal{U}$.
\item We propose a simple dynamic sampling strategy to allocate the sample budget more efficiently.
\end{enumerate}

\subsection{Fitting the Problem-Difficulty Distribution \texorpdfstring{$\mathcal{U}$}{U}}

\begin{figure}[ht] \label{uniform_example}
    \centering
    \includegraphics[width=0.6\linewidth]{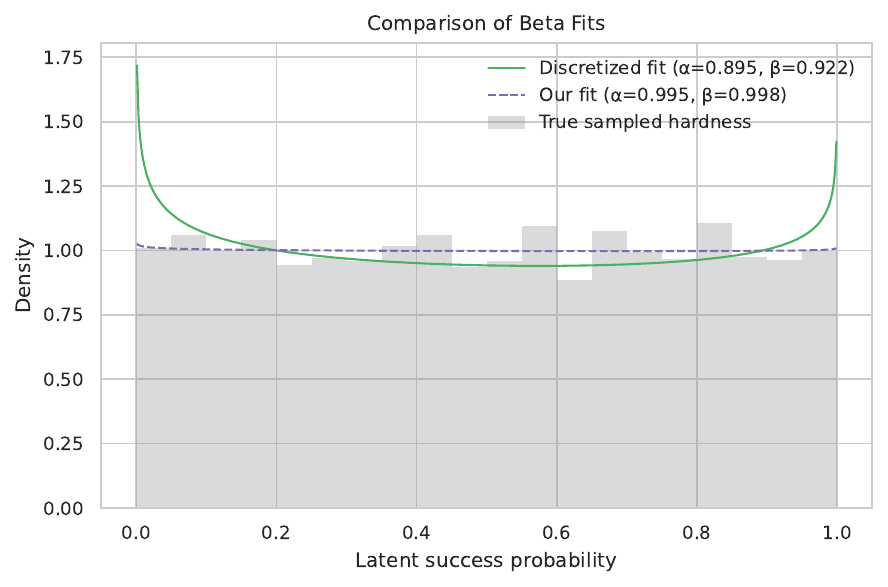}
    \caption{
    \textbf{Comparing Hardness Distribution Fit for Discretized Beta vs. Beta-Bernoulli}. \\
    $m=10\,000$ problem success probabilities are sampled: $\passiat{1} \sim \mathrm{Uniform}([0, 1])$. $b=100$ success/failure samples are drawn for each problem, $s_i \sim \mathrm{Bin}(b, \passiat{1})$.
    }
    \label{fig:beta-fit}
\end{figure}

Denote the underlying distribution of per-problem success probabilities as $\mathrm{pass}_i\text{@}1 \sim \mathcal{U}$, where $\mathcal{U}$ is unknown.
The number of successes $s_i$ on the $i$-th problem out of $b_i$ attempts is then binomially distributed: $s_i \sim \text{Binomial}(b_i, \mathrm{pass}_i\text{@}1)$.
Instead of the biased discretization approach, we model $\mathcal{U}$ as a beta distribution. This allows us to fit a beta-binomial distribution directly to the observed counts of successes and trials $(s_i, b_i)$.
The likelihood for the beta-binomial is given by:
\begin{equation}
\label{eq:beta-binomial}
\Pr\left[s=s_i \mid b=b_i; \alpha, \beta\right] = \binom{b_i}{s_i} \frac{\mathrm{Be}(s_i + \alpha, b_i - s_i + \beta)}{\mathrm{Be}(\alpha, \beta)},
\end{equation}
where $\mathrm{Be}(\cdot, \cdot)$ is the beta function. As shown in Figure~\ref{fig:beta-fit}, the discretized estimator badly fits a uniform distribution because it incorrectly puts excessive weight on the left tail.
Maximizing the beta-binomial likelihood leads to a superior fit which ultimately results in better predictions of $\mathrm{pass}\text{@}k$.

Next, we obtain a maximum likelihood estimate for $\mathcal{U}$:

\begin{equation} \label{eq:mle-beta}
    \hat{\alpha}, \hat{\beta} = \arg \max_{\alpha, \beta > 0} \prod_{i=1}^m \Pr\left[s=s_i \mid b=b_i; \alpha, \beta\right].
\end{equation}

Finally, we retrieve an estimate for $\passat{k}$:

\begin{equation} \label{convert_to_estimate}
    \widehat{\passdat{k}} = \E_{\passiat{1} \sim \mathrm{Beta}(\hat{\alpha}, \hat{\beta})} \left[ 1 - (1 - \passiat{1})^k \right].
\end{equation}

We see in Figure \ref{fig:beta-fit} that our approximate Beta-Bernoulli distribution better fits problem success probabilities sampled from a uniform distribution.

\subsection{More Efficient Sampling Strategies}

It was demonstrated by ~\cite{schaeffer2025largelanguagemonkeyspower} that in the high-$k$ regime, $\passat{k}$ scaling is governed almost exclusively by the shape of the difficulty distribution near $0$.
Distinguishing between an easy problem ($\passiat{1}=0.25$) and a very easy problem ($\passiat{1}=0.75$) provides little to no information.
Therefore, we propose to concentrate our budget, sampling responses from the hardest problem with the least number of attempts so far (see Algorithm~\ref{alg:select-hardest}).

This adaptive approach is not directly applicable to the regression-based estimator, which requires uniform sample count across problems to compute intermediate $\mathrm{pass}_{\mathcal{D}}\text{@}k$ values. It is also inconsistent with the discretized estimator from \cite{schaeffer2025largelanguagemonkeyspower} since direct estimates $\passhat{1} =\frac{s_i}{b_i}$ have different precision with this dynamic sampling method.
However, our distributional fitting method remains valid, as the beta-binomial likelihood (Equation \ref{eq:beta-binomial}) can handle variable numbers of trials ($b_i$) for each problem.
We outline our complete approach in Algorithm \ref{alg:estimator}.

\begin{algorithm}
\caption{Dynamic Sampling + Beta-Binomial Fit for Efficient $\passdat{k}$ Estimation}
\label{alg:estimator}
\begin{algorithmic}
\REQUIRE Dataset $\mathcal{D}$ with $m$ problems, total sample budget $B$, and number of repeated attempts $k$.
\STATE Initialize $\texttt{successes}_i \leftarrow 0$ and $\texttt{attempts}_i \leftarrow 0$ for all $i \in \{1, \ldots, m\}$
\FOR{$t \in \{1, \dots, B\}$}
    \STATE $i_t \leftarrow \texttt{SelectHardestProblem}(\texttt{successes}, \texttt{attempts})$
    \STATE $\texttt{attempts}_{i_t} \leftarrow \texttt{attempts}_{i_t} + 1$
    \STATE $\texttt{successes}_{i_t} \leftarrow \texttt{successes}_{i_t} + \mathbb{1}\left\{\texttt{AttemptProblem}(i_t)\right\}$
\ENDFOR
\STATE $\hat{\alpha}, \hat{\beta} \leftarrow \arg \max_{\alpha, \beta > 0} \prod_{i=1}^m \Pr\left[s=\texttt{successes}_i \mid b=\texttt{attempts}_i; \alpha, \beta\right]$ \hfill Equation~\ref{eq:mle-beta}
\STATE $\passhat{k} \leftarrow \E_{\passiat{1} \sim \mathrm{Beta}(\hat{\alpha}, \hat{\beta})} \left[ 1 - (1 - \passiat{1})^k \right]$ \hfill Equation~\ref{convert_to_estimate}
\STATE \textbf{return} $\passhat{k}$
\end{algorithmic}
\end{algorithm}

\paragraph{On the sample-efficiency of our estimator.}
The decision to select problems dynamically based on estimated problem difficulty is motivated by intuition from the theorems in \cite{schaeffer2025largelanguagemonkeyspower}.
To provide theoretical motivation for our approach, we introduce a natural frequentist estimator, defined below.
Given oracle access to $\passiat{1}$ and control over the number of samples for each problem $b_i$, we show that the variance can be minimized by prioritizing ``harder'' problems with low $\passiat{1}$.

\begin{lemma} \label{thm:optimal-rate}
    Consider the following estimator of $\passat{k}$
    \[
        \passhat{k}_{\text{freq}} \coloneqq 1 - \frac{1}{n} \sum_{i=1}^n (1 - s_i / b_i)^k.
    \]
    In the asymptotic regime as $n \to +\infty$, the sampling budget $b^*$ that minimizes the variance $\mathrm{Var}(\passhat{k}_{\text{freq}})$ is:
    \[
        b^*_i \propto \sqrt{(\passiat{1})(1 - \passiat{1})^{2k-1}}.
    \]
\end{lemma}

\begin{figure*}[ht]
    \centering
    \includegraphics[width=1.0\linewidth]{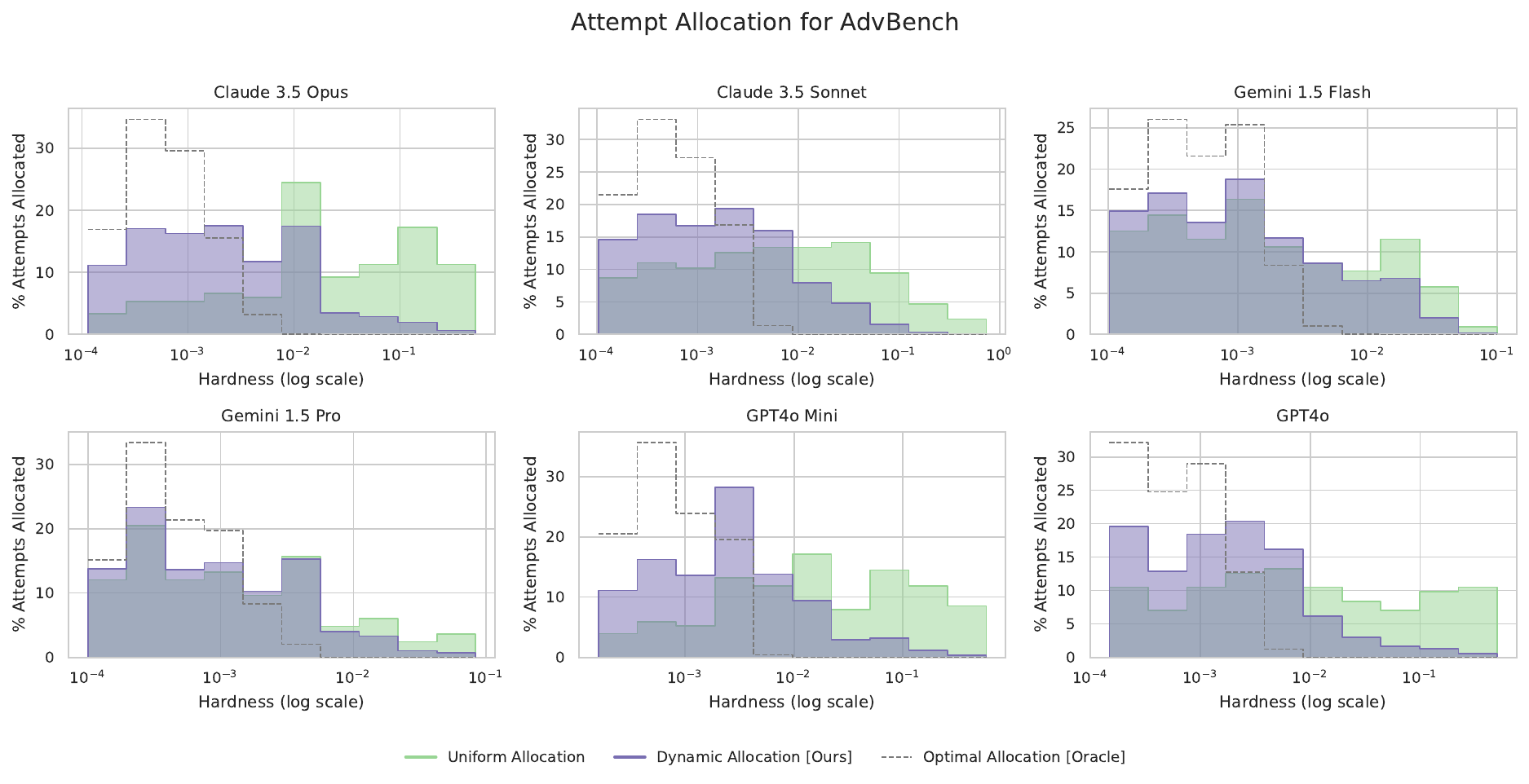}
    \caption{
    \textbf{Budget Allocation by Hardness Relative to the Optimal Allocation from Lemma \ref{thm:optimal-rate}}
    Contrasted distributions of problem success probabilities for the problems selected by dynamic and uniform sampling strategies on AdvBench.
    Note that these probabilities are not immediately available to our estimator but rather approximated given a limited amount of samples for each problem.
    The dotted line represents the distribution of problem success probabilities under the optimal sampling allocation provided in Lemma \ref{thm:optimal-rate}, assuming oracle access to the problem success probabilities.
    The dynamic strategy more closely aligns with this optimal rate.
    }
    \label{fig:allocation-jailbreaking}
\end{figure*}

A proof of Lemma \ref{thm:optimal-rate} is provided in Appendix \ref{sec:opt-distribution}.
For high $k$, the optimal budget prioritizes lower $\passiat{1}$.
We conjecture that such adaptive strategies can also reduce variance in the context of our multi-stage Bayesian approach, but we leave such analysis for future work.
Beyond this, we show in Figure \ref{fig:allocation-jailbreaking} that the distribution of the difficulties of problems selected by our dynamic strategy aligns much more closely with the derived optimal allocation from Lemma \ref{thm:optimal-rate} than that of a uniform strategy.

However, in the sample-count regimes and distributions in our datasets, it is difficult to empirically isolate the benefits of the sampling method alone.
Therefore, we provide some additional empirical support for dynamic sampling on synthetic data in Appendix \ref{sec:opt-distribution}.  We find that when there are many easy problems and a small number of hard outliers, or a uniform distribution of difficulties, the dynamic sampling method outperforms uniform sampling by large margins. On all distributions tested, dynamic sampling performs better than or comparably to uniform sampling.

\begin{figure*}[ht]
    \centering
    \includegraphics[width=1.0\linewidth]{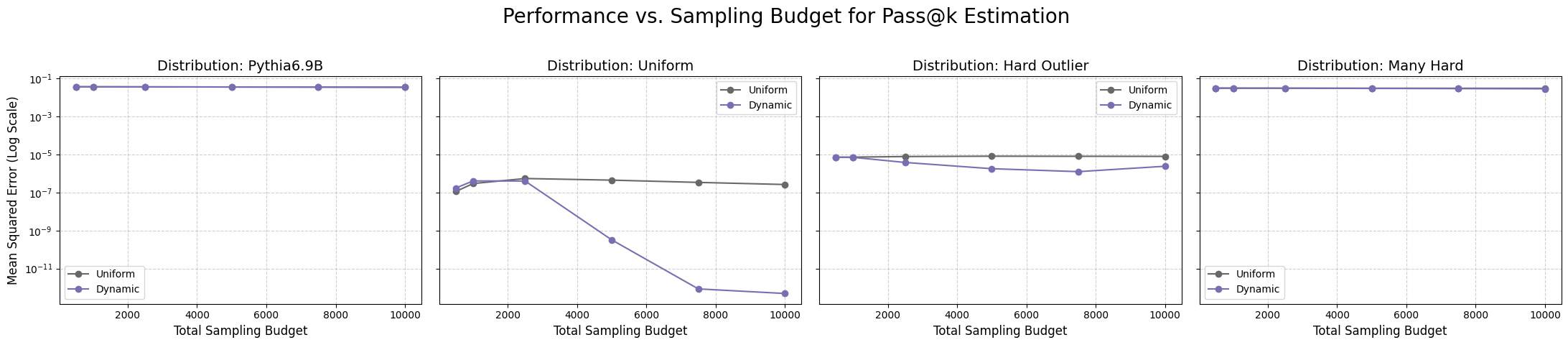}
    \caption{
    \textbf{
    Evaluating Performance Scaling for Uniform vs. Dynamic Allocation Strategies} \\
    Dynamic sampling is most useful when there are a handful of very difficult problems, but many easy problems.
    These distributions allow it to concentrate a large proportion of the budget on difficult problems.
    The ``Hard Outlier" distribution has a single very difficult problem with success probability $1e-4$, and all other problems with difficulties in the range of $0.1$-$0.3$.}
    \label{fig:placeholder}
\end{figure*}

\paragraph{Validity under adaptive sampling.}
We assume that per-problem attempt outcomes represent fresh Bernoulli draws with fixed per-problem probabilities of success $0<\passiat{1}<1$.
This assumption notably \textit{excludes history-dependent methods of repeated sampling}, such as self-reflection~\citep{shinn2023reflexion}, code repair guided by execution feedback~\citep{le2022coderl}, and adaptive jailbreak prompting~\citep{chao2025jailbreaking}.

From here, convergence to the pseudo-true $\passat{k}$
\[
    \passdat{k}^* = \E_{\passiat{1} \sim \mathrm{Beta}(\alpha^*, \beta^*)} \left[ 1 - (1 - \passiat{1})^k \right].
\]
with likelihood-maximizing $\alpha^*, \beta^*$
\[
    \alpha^*, \beta^* = \arg \max_{\alpha, \beta} \prod_{i=1}^m \frac{\passiat{1}^{\alpha-1}(1 - \passiat{1})^{\beta-1}}{\mathrm{Be}(\alpha, \beta)}.
\]
follows from two key properties of our acquisition function: $\texttt{SelectHardestProblem}$ depends only previous outcomes and attempts each problem infinitely often as $B \to \infty$, since the minimum success count increases after every problem obtains another success, which occurs almost surely.

By the first property, the acquisition rule contributes a
parameter-independent factor to the likelihood of the observed
trajectory~\citep[Eq.~(4.7)]{lattimore2020bandit}. Integrating
the Bernoulli outcome factors over independent
$\passiat{1}\sim\mathrm{Beta}(\alpha,\beta)$ gives
\[
L(\alpha,\beta)\propto
\prod_{i=1}^{m}
\frac{\mathrm{Be}(\alpha+s_i,\beta+b_i-s_i)}
     {\mathrm{Be}(\alpha,\beta)}.
\]
The binomial coefficients in Equation~\ref{eq:mle-beta} are also
parameter-independent, so the two objectives have the same maximizer.


By the second property, every problem is sampled infinitely often as $B\to\infty$ with $m$ fixed, so $s_i/b_i \xrightarrow{\mathrm{a.s.}} \passiat{1}$ by the strong law.
For convergence, we take both the fitted and oracle parameters to be global maximizers over the same fixed compact convex domain $\Theta\subset(0,\infty)^2$.
After removing parameter-independent terms, the Beta--Binomial log likelihood converges almost surely and uniformly on $\Theta$ to the oracle Beta log likelihood obtained by observing the true success probabilities directly.
This oracle objective is strictly concave and therefore has a unique maximizer, so $(\hat{\alpha},\hat{\beta}) \xrightarrow{\mathrm{a.s.}} (\alpha^*,\beta^*)$ by the strong-consistency argument of \citep[p.~2122]{newey1994large}.
By continuity, $\passhat{k} \xrightarrow{\mathrm{a.s.}} \passdat{k}^*$ for each fixed $k$.
The limiting prediction need not equal the true $\passdat{k}$, since the fitted Beta distribution may not capture the finite benchmark's distribution of success probabilities.
\section{Results}

In this section, we evaluate the predictive accuracy of our method.
We estimate $\mathrm{pass}_{\mathcal{D}}\text{@}k$ for $k$ in the range $[10^1, 10^3]$ on three real-world datasets and three to six different models for each dataset

\subsection{Experimental Setup}

We source our data from \cite{brown2024largelanguagemonkeysscaling} and \cite{hughes2024bestofnjailbreaking}, which contain $10^4$ sampled successful or failed attempts for each of $100 \sim 200$ problems selected from Code Contests~\citep{code_contests}, MATH~\citep{hendrycksmath2021}, and AdvBench~\citep{zou2023universal}.
We extrapolate from $10^1 \leq B \leq 10^3$ shuffled samples.
Ground truth $\passat{k}$ values are computed using all $10^4$ samples.

\begin{itemize}[leftmargin=6mm]
    \item \textbf{Uniform Sampling} (Log-Log Regression, Discretized Beta): use the first $B/m$ for each problem.
    \item \textbf{Dynamic Sampling} (+ Beta-Binomial Fit): adaptive sampling as defined in Algorithm~\ref{alg:estimator}.
\end{itemize}

\subsection{Discussion}

\begin{figure*}[h]
    \centering
    \includegraphics[width=\linewidth]{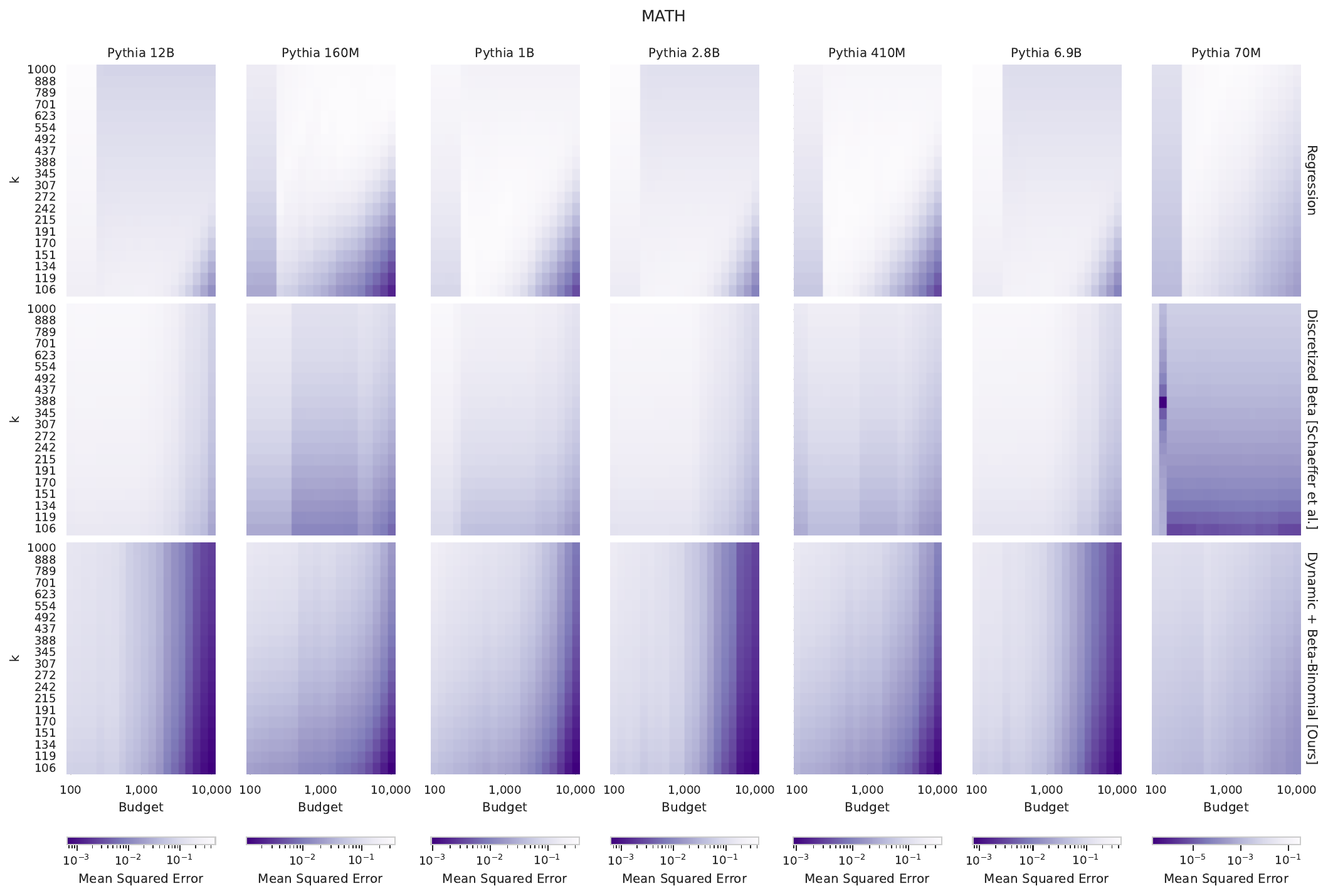}
    \caption{
    Heatmap depicting how predictions of $\passat{k}$ change with the sampling budget and $k$ on MATH.
    Our method minimizes MSE for virtually all values of $k$ and sampling budgets, as evidenced by the darker colors in its heatmap.
    Figures for MATH and Code Contests are in Appendix \ref{additional_figures}.
    }
    \label{fig:math_heatmap}
\end{figure*}

The predictions for AdvBench, MATH, and Code Contests with different sampling budgets are shown in Figure~\ref{fig:prediction_results}.
We observe that \textbf{existing estimators diverge significantly from the true $\passat{k}$ value beyond this threshold}.  Figure \ref{fig:math_heatmap} provides a heat map of errors for different sampling budgets and values of $k$.
Note that, as expected, the error decreases as we increase the sampling budget.  Existing estimators especially struggle with high values of $k$.
We also provide the MSE for each estimator across different sampling budgets in Appendix \ref{additional_figures}.

Across models and datasets, our proposed method provides predictions that are closest to the ground truth.
The predictions from log-log regression are particularly poor, often diverging to predict impossible pass rates greater than 1 (we clip these at 1 for visualization and error computation).
The prior distributional fitting method from \cite{schaeffer2025largelanguagemonkeyspower} performs better than unclipped regression but consistently underestimates $\passat{k}$ for large $k$.  

The gains from our method come from several sources.  Our fitting uses maximum likelihood estimation on an exponential family model, which is known to have properties like asymptotic normality, unbiasedness, and $O(1/\sqrt{n})$-convergence.  We avoid the pitfall of fitting models on top of correlated estimates, as in the regression method.  Finally, we align our sampling budget more closely with the distribution that theoretically minimizes variance.

\section{Conclusion and Future Work}

Predicting the capabilities and vulnerabilities of AI models at scale is a critical challenge.
We provide two core improvements: (1) selecting a more appropriate model for the underlying problem difficulties, and (2) utilizing dynamic sampling to concentrate compute on the most difficult problems.
We demonstrate the impact of these innovations in Figure~\ref{fig:math_heatmap} on mathematical problem-solving.

We achieved large improvements in predictive accuracy by remedying statistical errors in prior methods and improving sampling techniques, without requiring extra sampling compute. These gains suggest that a closer statistical inspection of other scaling-law fitting methodologies could lead to considerable computational savings and, ultimately, better and safer models.


\section{Related Work}

Repeated sampling from LMs improves performance on verifiable tasks, and has become the backbone of reinforcement learning from verified rewards \cite{shao2024deepseekmathpushinglimitsmathematical}.  The HumanEval benchmark first defined pass@$k$ and derived an unbiased estimator for this quantity \citep{chen2021evaluatinglargelanguagemodels}.  Follow-up work on code generation studied sampling as a resource that must be allocated across problems; for instance, \cite{han-etal-2023-sample} adaptively prioritizes sampling solvable instances.  They estimate pass@$k$ with Monte-Carlo, whereas we predict how pass@$k$ varies with $k$ on a fixed budget.  

For unverifiable rewards, repeated sampling must be combined with aggregation techniques such as majority voting \citep{wang2023selfconsistencyimproveschainthought}.
Self-consistency shows that majority voting over multiple chains-of-thought can boost reasoning performance, and adaptive-consistency \citep{aggarwal2023letssamplestepstep} reduces compute by stopping sampling once a clear majority emerges.
\cite{NEURIPS2024_51173cf3} study compound AI systems that repeatedly query an LM, aggregate responses, and describe how performance varies with the number of LM calls.  

Statistically, estimating pass@$k$ involves binomial success probabilities that can be extremely small on difficult benchmark items, making variance control and uncertainty quantification important.  Classical work on binomial proportion intervals compares exact and approximate methods, including the Clopper-Pearson interval and score-based or adjusted Wald intervals \citep{10aa271a-cc2e-38f1-b697-f9f66032ef16}.  Recent analysis in the rare-event regime emphasizes that both coverage and relative margin of error are important when designing estimators and determining sample sizes \cite{mcgrath2024binomialconfidenceintervalsrare}. 
\clearpage

\begin{ack}
This work is partially supported by NSF 2046795, 2205329, and 2504264, NIH, ARPA-H, the MacArthur Foundation, Good Ventures, Schmidt Sciences, the Hasso Plattner Foundation and Stanford HAI.
\end{ack}

\bibliography{neurips2026/references}

@misc{wang2023selfconsistencyimproveschainthought,
      title={Self-Consistency Improves Chain of Thought Reasoning in Language Models}, 
      author={Xuezhi Wang and Jason Wei and Dale Schuurmans and Quoc Le and Ed Chi and Sharan Narang and Aakanksha Chowdhery and Denny Zhou},
      year={2023},
      eprint={2203.11171},
      archivePrefix={arXiv},
      primaryClass={cs.CL},
      url={https://arxiv.org/abs/2203.11171}, 
}

@misc{mcgrath2024binomialconfidenceintervalsrare,
      title={Binomial confidence intervals for rare events: importance of defining margin of error relative to magnitude of proportion}, 
      author={Owen McGrath and Kevin Burke},
      year={2024},
      eprint={2109.02516},
      archivePrefix={arXiv},
      primaryClass={stat.AP},
      url={https://arxiv.org/abs/2109.02516}, 
}

@article{10aa271a-cc2e-38f1-b697-f9f66032ef16,
 ISSN = {08834237, 21688745},
 URL = {http://www.jstor.org/stable/2676784},
 author = {Lawrence D. Brown and T. Tony Cai and Anirban DasGupta},
 journal = {Statistical Science},
 number = {2},
 pages = {101--117},
 publisher = {Institute of Mathematical Statistics},
 title = {Interval Estimation for a Binomial Proportion},
 urldate = {2025-11-17},
 volume = {16},
 year = {2001}
}

@inproceedings{NEURIPS2024_51173cf3,
 author = {Chen, Lingjiao and Davis, Jared and Hanin, Boris and Bailis, Peter and Stoica, Ion and Zaharia, Matei and Zou, James},
 booktitle = {Advances in Neural Information Processing Systems},
 doi = {10.52202/079017-1455},
 editor = {A. Globerson and L. Mackey and D. Belgrave and A. Fan and U. Paquet and J. Tomczak and C. Zhang},
 pages = {45767--45790},
 publisher = {Curran Associates, Inc.},
 title = {Are More LLM Calls All You Need? Towards the Scaling Properties of Compound AI Systems},
 url = {https://proceedings.neurips.cc/paper_files/paper/2024/file/51173cf34c5faac9796a47dc2fdd3a71-Paper-Conference.pdf},
 volume = {37},
 year = {2024}
}

@misc{aggarwal2023letssamplestepstep,
      title={Let's Sample Step by Step: Adaptive-Consistency for Efficient Reasoning and Coding with LLMs}, 
      author={Pranjal Aggarwal and Aman Madaan and Yiming Yang and Mausam},
      year={2023},
      eprint={2305.11860},
      archivePrefix={arXiv},
      primaryClass={cs.CL},
      url={https://arxiv.org/abs/2305.11860}, 
}

@inproceedings{han-etal-2023-sample,
    title = "On Sample-Efficient Code Generation",
    author = "Han, Hojae  and
      Kim, Yu Jin  and
      Kim, Byoungjip  and
      Lee, Youngwon  and
      Lee, Kyungjae  and
      Lee, Kyungmin  and
      Lee, Moontae  and
      Bae, Kyunghoon  and
      Hwang, Seung-won",
    editor = "Wang, Mingxuan  and
      Zitouni, Imed",
    booktitle = "Proceedings of the 2023 Conference on Empirical Methods in Natural Language Processing: Industry Track",
    month = dec,
    year = "2023",
    address = "Singapore",
    publisher = "Association for Computational Linguistics",
    url = "https://aclanthology.org/2023.emnlp-industry.73/",
    doi = "10.18653/v1/2023.emnlp-industry.73",
    pages = "783--791"
}

@misc{shao2024deepseekmathpushinglimitsmathematical,
      title={DeepSeekMath: Pushing the Limits of Mathematical Reasoning in Open Language Models}, 
      author={Zhihong Shao and Peiyi Wang and Qihao Zhu and Runxin Xu and Junxiao Song and Xiao Bi and Haowei Zhang and Mingchuan Zhang and Y. K. Li and Y. Wu and Daya Guo},
      year={2024},
      eprint={2402.03300},
      archivePrefix={arXiv},
      primaryClass={cs.CL},
      url={https://arxiv.org/abs/2402.03300}, 
}

@article{
zellinger2025rational,
title={Rational Tuning of {LLM} Cascades via Probabilistic Modeling},
author={Michael J. Zellinger and Matt Thomson},
journal={Transactions on Machine Learning Research},
issn={2835-8856},
year={2025},
url={https://openreview.net/forum?id=YCBVcGSZeR},
note={}
}

@misc{levi2024simplemodelinferencescaling,
      title={A Simple Model of Inference Scaling Laws}, 
      author={Noam Levi},
      year={2024},
      eprint={2410.16377},
      archivePrefix={arXiv},
      primaryClass={stat.ML},
      url={https://arxiv.org/abs/2410.16377}, 
}

@misc{howe2025scalingtrendslanguagemodel,
      title={Scaling Trends in Language Model Robustness}, 
      author={Nikolaus Howe and Ian McKenzie and Oskar Hollinsworth and Michał Zajac and Tom Tseng and Aaron Tucker and Pierre-Luc Bacon and Adam Gleave},
      year={2025},
      eprint={2407.18213},
      archivePrefix={arXiv},
      primaryClass={cs.LG},
      url={https://arxiv.org/abs/2407.18213}, 
}

@misc{glazer2024frontiermathbenchmarkevaluatingadvanced,
      title={FrontierMath: A Benchmark for Evaluating Advanced Mathematical Reasoning in AI}, 
      author={Elliot Glazer and Ege Erdil and Tamay Besiroglu and Diego Chicharro and Evan Chen and Alex Gunning and Caroline Falkman Olsson and Jean-Stanislas Denain and Anson Ho and Emily de Oliveira Santos and Olli Järviniemi and Matthew Barnett and Robert Sandler and Matej Vrzala and Jaime Sevilla and Qiuyu Ren and Elizabeth Pratt and Lionel Levine and Grant Barkley and Natalie Stewart and Bogdan Grechuk and Tetiana Grechuk and Shreepranav Varma Enugandla and Mark Wildon},
      year={2024},
      eprint={2411.04872},
      archivePrefix={arXiv},
      primaryClass={cs.AI},
      url={https://arxiv.org/abs/2411.04872}, 
}

@misc{panfilov2025capabilitybasedscalinglawsllm,
      title={Capability-Based Scaling Laws for LLM Red-Teaming}, 
      author={Alexander Panfilov and Paul Kassianik and Maksym Andriushchenko and Jonas Geiping},
      year={2025},
      eprint={2505.20162},
      archivePrefix={arXiv},
      primaryClass={cs.AI},
      url={https://arxiv.org/abs/2505.20162}, 
}

@inproceedings{jimenez2024swe,
  title={SWE-bench: Can Language Models Resolve Real-world Github Issues?},
  author={Jimenez, Carlos E and Yang, John and Wettig, Alexander and Yao, Shunyu and Pei, Kexin and Press, Ofir and Narasimhan, Karthik R},
  year={2024},
  booktitle={The Twelfth International Conference on Learning Representations}
}

@misc{kazdan2025nocourseican,
      title={No, of Course I Can! Deeper Fine-Tuning Attacks That Bypass Token-Level Safety Mechanisms}, 
      author={Joshua Kazdan and Abhay Puri and Rylan Schaeffer and Lisa Yu and Chris Cundy and Jason Stanley and Sanmi Koyejo and Krishnamurthy Dvijotham},
      year={2025},
      eprint={2502.19537},
      archivePrefix={arXiv},
      primaryClass={cs.CR},
      url={https://arxiv.org/abs/2502.19537}, 
}

@misc{schaeffer2025largelanguagemonkeyspower,
      title={How Do Large Language Monkeys Get Their Power (Laws)?}, 
      author={Rylan Schaeffer and Joshua Kazdan and John Hughes and Jordan Juravsky and Sara Price and Aengus Lynch and Erik Jones and Robert Kirk and Azalia Mirhoseini and Sanmi Koyejo},
      year={2025},
      eprint={2502.17578},
      archivePrefix={arXiv},
      primaryClass={cs.AI},
      url={https://arxiv.org/abs/2502.17578}, 
}

@misc{hughes2024bestofnjailbreaking,
      title={Best-of-N Jailbreaking}, 
      author={John Hughes and Sara Price and Aengus Lynch and Rylan Schaeffer and Fazl Barez and Sanmi Koyejo and Henry Sleight and Erik Jones and Ethan Perez and Mrinank Sharma},
      year={2024},
      eprint={2412.03556},
      archivePrefix={arXiv},
      primaryClass={cs.CL},
      url={https://arxiv.org/abs/2412.03556}, 
}

@article{
  code_contests,
  author = {Yujia Li  and David Choi  and Junyoung Chung  and Nate Kushman  and Julian Schrittwieser  and R{\'e}mi Leblond  and Tom Eccles  and James Keeling  and Felix Gimeno  and Agustin Dal Lago  and Thomas Hubert  and Peter Choy  and Cyprien de Masson d’Autume  and Igor Babuschkin  and Xinyun Chen  and Po-Sen Huang  and Johannes Welbl  and Sven Gowal  and Alexey Cherepanov  and James Molloy  and Daniel J. Mankowitz  and Esme Sutherland Robson  and Pushmeet Kohli  and Nando de Freitas  and Koray Kavukcuoglu  and Oriol Vinyals },
  title = {Competition-level code generation with AlphaCode},
  journal = {Science},
  volume = {378},
  number = {6624},
  pages = {1092-1097},
  year = {2022},
  doi = {10.1126/science.abq1158},
  URL = {https://www.science.org/doi/abs/10.1126/science.abq1158},
  eprint = {https://www.science.org/doi/pdf/10.1126/science.abq1158}}

@misc{kwok2025robomonkeyscalingtesttimesampling,
      title={RoboMonkey: Scaling Test-Time Sampling and Verification for Vision-Language-Action Models}, 
      author={Jacky Kwok and Christopher Agia and Rohan Sinha and Matt Foutter and Shulu Li and Ion Stoica and Azalia Mirhoseini and Marco Pavone},
      year={2025},
      eprint={2506.17811},
      archivePrefix={arXiv},
      primaryClass={cs.RO},
      url={https://arxiv.org/abs/2506.17811}, 
}

@article{hendrycksmath2021,
  title={Measuring Mathematical Problem Solving With the MATH Dataset},
  author={Dan Hendrycks and Collin Burns and Saurav Kadavath and Akul Arora and Steven Basart and Eric Tang and Dawn Song and Jacob Steinhardt},
  journal={NeurIPS},
  year={2021}
}

@misc{zou2023universal,
      title={Universal and Transferable Adversarial Attacks on Aligned Language Models}, 
      author={Andy Zou and Zifan Wang and J. Zico Kolter and Matt Fredrikson},
      year={2023},
      eprint={2307.15043},
      archivePrefix={arXiv},
      primaryClass={cs.CL}
}

@inproceedings{kulal2019spoc,
 author = {Kulal, Sumith and Pasupat, Panupong and Chandra, Kartik and Lee, Mina and Padon, Oded and Aiken, Alex and Liang, Percy S},
 booktitle = {Advances in Neural Information Processing Systems},
 editor = {H. Wallach and H. Larochelle and A. Beygelzimer and F. d\textquotesingle Alch\'{e}-Buc and E. Fox and R. Garnett},
 pages = {},
 publisher = {Curran Associates, Inc.},
 title = {SPoC: Search-based Pseudocode to Code},
 url = {https://proceedings.neurips.cc/paper_files/paper/2019/file/7298332f04ac004a0ca44cc69ecf6f6b-Paper.pdf},
 volume = {32},
 year = {2019}
}

@inproceedings{du2024evaluatingcode,
author = {Du, Xueying and Liu, Mingwei and Wang, Kaixin and Wang, Hanlin and Liu, Junwei and Chen, Yixuan and Feng, Jiayi and Sha, Chaofeng and Peng, Xin and Lou, Yiling},
title = {Evaluating Large Language Models in Class-Level Code Generation},
year = {2024},
isbn = {9798400702174},
publisher = {Association for Computing Machinery},
address = {New York, NY, USA},
url = {https://doi.org/10.1145/3597503.3639219},
doi = {10.1145/3597503.3639219},
booktitle = {Proceedings of the IEEE/ACM 46th International Conference on Software Engineering},
articleno = {81},
numpages = {13},
location = {Lisbon, Portugal},
series = {ICSE '24}
}

@misc{chen2021evaluatinglargelanguagemodels,
      title={Evaluating Large Language Models Trained on Code}, 
      author={Mark Chen and Jerry Tworek and Heewoo Jun and Qiming Yuan and Henrique Ponde de Oliveira Pinto and Jared Kaplan and Harri Edwards and Yuri Burda and Nicholas Joseph and Greg Brockman and Alex Ray and Raul Puri and Gretchen Krueger and Michael Petrov and Heidy Khlaaf and Girish Sastry and Pamela Mishkin and Brooke Chan and Scott Gray and Nick Ryder and Mikhail Pavlov and Alethea Power and Lukasz Kaiser and Mohammad Bavarian and Clemens Winter and Philippe Tillet and Felipe Petroski Such and Dave Cummings and Matthias Plappert and Fotios Chantzis and Elizabeth Barnes and Ariel Herbert-Voss and William Hebgen Guss and Alex Nichol and Alex Paino and Nikolas Tezak and Jie Tang and Igor Babuschkin and Suchir Balaji and Shantanu Jain and William Saunders and Christopher Hesse and Andrew N. Carr and Jan Leike and Josh Achiam and Vedant Misra and Evan Morikawa and Alec Radford and Matthew Knight and Miles Brundage and Mira Murati and Katie Mayer and Peter Welinder and Bob McGrew and Dario Amodei and Sam McCandlish and Ilya Sutskever and Wojciech Zaremba},
      year={2021},
      eprint={2107.03374},
      archivePrefix={arXiv},
      primaryClass={cs.LG},
      url={https://arxiv.org/abs/2107.03374}, 
}

@book{l1983theory,
  title={Theory of Point Estimation},
  author={Lehmann, L.E.},
  series={A Wiley publication in mathematical statistics},
  url={https://books.google.com/books?id=VcXdngEACAAJ},
  year={1983},
  publisher={Wiley}
}

@misc{brown2024largelanguagemonkeysscaling,
      title={Large Language Monkeys: Scaling Inference Compute with Repeated Sampling}, 
      author={Bradley Brown and Jordan Juravsky and Ryan Ehrlich and Ronald Clark and Quoc V. Le and Christopher Ré and Azalia Mirhoseini},
      year={2024},
      eprint={2407.21787},
      archivePrefix={arXiv},
      primaryClass={cs.LG},
      url={https://arxiv.org/abs/2407.21787}, 
}

@inproceedings{anil2024manyshot,
title={Many-shot Jailbreaking},
author={Cem Anil and Esin Durmus and Nina Rimsky and Mrinank Sharma and Joe Benton and Sandipan Kundu and Joshua Batson and Meg Tong and Jesse Mu and Daniel J Ford and Francesco Mosconi and Rajashree Agrawal and Rylan Schaeffer and Naomi Bashkansky and Samuel Svenningsen and Mike Lambert and Ansh Radhakrishnan and Carson Denison and Evan J Hubinger and Yuntao Bai and Trenton Bricken and Timothy Maxwell and Nicholas Schiefer and James Sully and Alex Tamkin and Tamera Lanham and Karina Nguyen and Tomasz Korbak and Jared Kaplan and Deep Ganguli and Samuel R. Bowman and Ethan Perez and Roger Baker Grosse and David Duvenaud},
booktitle={The Thirty-eighth Annual Conference on Neural Information Processing Systems},
year={2024},
url={https://openreview.net/forum?id=cw5mgd71jW}
}

@book{lattimore2020bandit,
  author    = {Lattimore, Tor and Szepesv{\'a}ri, Csaba},
  title     = {Bandit Algorithms},
  publisher = {Cambridge University Press},
  year      = {2020},
  doi       = {10.1017/9781108571401}
}

@incollection{newey1994large,
  author    = {Newey, Whitney K. and McFadden, Daniel},
  title     = {Large Sample Estimation and Hypothesis Testing},
  booktitle = {Handbook of Econometrics},
  editor    = {Engle, Robert F. and McFadden, Daniel L.},
  volume    = {4},
  chapter   = {36},
  pages     = {2111--2245},
  publisher = {Elsevier},
  year      = {1994},
  doi       = {10.1016/S1573-4412(05)80005-4}
}

@inproceedings{shinn2023reflexion,
  author    = {Shinn, Noah and Cassano, Federico and Gopinath, Ashwin
               and Narasimhan, Karthik and Yao, Shunyu},
  title     = {{Reflexion}: Language Agents with Verbal Reinforcement Learning},
  booktitle = {Advances in Neural Information Processing Systems},
  volume    = {36},
  year      = {2023},
  doi       = {10.52202/075280-0377}
}

@inproceedings{le2022coderl,
  author    = {Le, Hung and Wang, Yue and Gotmare, Akhilesh Deepak
               and Savarese, Silvio and Hoi, Steven Chu Hong},
  title     = {{CodeRL}: Mastering Code Generation through Pretrained
               Models and Deep Reinforcement Learning},
  booktitle = {Advances in Neural Information Processing Systems},
  volume    = {35},
  year      = {2022},
  doi       = {10.52202/068431-1549}
}

@inproceedings{chao2025jailbreaking,
  author    = {Chao, Patrick and Robey, Alexander and Dobriban, Edgar
               and Hassani, Hamed and Pappas, George J. and Wong, Eric},
  title     = {Jailbreaking Black Box Large Language Models in Twenty Queries},
  booktitle = {2025 IEEE Conference on Secure and Trustworthy Machine Learning
               ({SaTML})},
  pages     = {23--42},
  publisher = {IEEE},
  year      = {2025},
  doi       = {10.1109/SaTML64287.2025.00010}
}
\bibliographystyle{plain}

\newpage


\appendix

\section{Limitations}

While we predict pass@$k$ on a scale of $\textrm{pass}@10\,000$ using dozens to hundreds of samples, frontier labs face the challenge of predicting $\textrm{pass}@k$ for $k$ several orders of magnitude higher.  These frontier labs also have the capability to generate far more samples to aid their predictions than we can.  Although we hope that the lessons of how to cross orders of magnitude when predicting pass@$k$ transfer, power law fitting might be more viable when the number of samples available is higher.  In the same vein, because we are limited in the number of samples that we can draw, we are unable to estimate the precision of pass@$1$ beyond $1e-4$, which creates hard limits for us when testing the accuracy of our models.

\section{Impact Statement}
This paper presents work whose goal is to advance the field of machine learning. There are many potential societal consequences of our work, none of which we feel must be specifically highlighted.

\section{Adaptive Sampling Criterion}

\begin{algorithm}
\caption{$\texttt{SelectHardestProblem}$}
\label{alg:select-hardest}
\begin{algorithmic}
\REQUIRE Dataset $\mathcal{D}$ with $m$ problems and per-problem counts of successful and total attempts: \texttt{successes} and \texttt{attempts}, respectively.

\STATE $s^* \leftarrow \min_i \texttt{successes}_i$
\STATE $H \leftarrow \arg\min_{\{i \,:\,\texttt{successes}_i = s^*\}} \texttt{attempts}_i$
\STATE $i^* \sim \mathrm{Uniform}(H)$
\STATE \textbf{return} $i^*$
\end{algorithmic}
\end{algorithm}

\section{Pitfalls of Linear Regression} \label{regression_proofs}

In this section, we precisely quantify the statements made in Section \ref{regression_issues}.

\textbf{The estimates $\widehat{\passat{k}}$ are not independent for different $k$:} Recall that one of the assumptions of the linear regression model is that the observations are independent.  The following lemma characterizes this non-independence on a per-problem basis:
\begin{lemma}
    Recall that $s_i$ is the number of successes observed out of $b$ attempts on the $i$th problem of $\mathcal{D}$. If $k \geq l$, and $0 < s_i < b$ then there exists an invertible function $f$ such that
    \begin{align} \widehat{\text{pass}_i@k}=f\left(\widehat{\text{pass}_i@l}\right).\end{align}

    This invertible function takes the form:
     \begin{equation} f\left(\widehat{\text{pass}_i@l}\right) = \widehat{\text{pass}_i@l} + s_i \sum_{m=l}^{k-1} \frac{\binom{b - s_i}{m}}{(b-m)\binom{b}{m}}.\end{equation}
\end{lemma}

\begin{proof}
\begin{align*}
\text{Let} \quad g(m) = \frac{\binom{b - s_i}{m}}{\binom{b}{m}}, \quad \text{then} \quad \widehat{\text{pass}_i@m} = 1 - g(m).\\
\end{align*}
\begin{align*}
\text{Now,} \quad \frac{g(m+1)}{g(m)} 
&= \frac{\binom{b - s_i}{m+1}}{\binom{b}{m+1}} \cdot \frac{\binom{b}{m}}{\binom{b - s_i}{m}} \\
&= \frac{\binom{b - s_i}{m+1}}{\binom{b - s_i}{m}} \cdot \frac{\binom{b}{m}}{\binom{b}{m+1}} \\
&= \frac{b - s_i - m}{m+1} \cdot \frac{m+1}{b-m} \\
&= \frac{b - s_i - m}{b - m}.\\\\
\Rightarrow 1 - g(m+1) &= 1 - \frac{b - s_i - m}{b - m} g(m) \\
\Rightarrow 1 - g(m+1) &= (1 - g(m)) + g(m) \left(1 - \frac{b - s_i - m}{b - m}\right) \\
&= (1 - g(m)) + g(m) \cdot \frac{s_i}{b - m}.\\\\
\Rightarrow \widehat{\text{pass}_i@(m+1)} 
&= \widehat{\text{pass}_i@m} + g(m) \cdot \frac{s_i}{b - m} \\
\Rightarrow \widehat{\text{pass}_i@k} 
&= \widehat{\text{pass}_i@l} + s_i \sum_{m=l}^{k-1} \frac{1}{b - m} g(m) \quad \text{as desired}.
\end{align*}
\end{proof}

    This lemma implies that given $\passiat{k}$ for any $k$, $\passiat{j}$ for $j\neq k$ is uniquely determined. 

    \textbf{The estimates of $\widehat{\passat{k}}$ have different variances for different values of $k$:} A second assumption of the linear regression model is that the noise in the model is homoscedastic, i.e. the noise is the same for all $k$.  This is again not the case for the estimators $\widehat{\passat{k}}$.  The following lemma gives one instance in which these estimators are not homoscedastic:

    \begin{lemma}
        Suppose that we have $n$ samples from a language model on problem $i$, and the language model has true probability $p$ of getting problem $i$ correct.  Then \begin{equation} \textrm{Var}\left(\widehat{\passiat{n}}\right) = (1-p)^n - (1-p)^{2n}, \end{equation} and \begin{equation} \textrm{Var}\left(\widehat{\passiat{1}}\right) = p(1-p)/n.\end{equation}
    \end{lemma}

\begin{proof}

Let \(c\sim\text{Binomial}(n,p)\) be the number of correct completions
obtained from \(n\) i.i.d.\ samples of a fixed problem~\(i\).  
For each \(k\in\{0,1,\dots,n\}\) define the empirical
\emph{pass@\(\!k\)} estimator
\[
  \passhat{k}=f_k(c), \text{ where }
  f_k(c)
  \;=\;
  1-\frac{\binom{\,n-c\,}{\,k\,}}{\binom{\,n\,}{\,k\,}}
\]

Our goal is to show that the variances of \(\passhat{k}\) are not
constant in \(k\).  We begin with the variance in its raw
definition:
\[
  \operatorname{Var}\bigl[f_k(c)\bigr]
  \;=\;
  \underbrace{\mathbb{E}\!\bigl[f_k(c)^2\bigr]}_{(a)}
  \;-\;
  \underbrace{\Bigl(\mathbb{E}\!\bigl[f_k(c)\bigr]\Bigr)^{2}}_{(b)}.
  \tag{$\star$}
\]
Both expectations can be written as finite sums over the
binomial probability‐mass function:
\[
\begin{aligned}
(a)\;=\;
  \sum_{c=0}^{n}
  \Bigl(1-\tfrac{\binom{n-c}{k}}{\binom{n}{k}}\Bigr)^{\!2}
  \binom{n}{c}\,p^{c}(1-p)^{\,n-c},
\quad
(b)\;=\;
  \Bigl(
  \sum_{c=0}^{n}
  \Bigl(1-\tfrac{\binom{n-c}{k}}{\binom{n}{k}}\Bigr)
  \binom{n}{c}\,p^{c}(1-p)^{\,n-c}
  \Bigr)^{\!2}.
\end{aligned}
\]

We now specialize to two extreme choices of \(k\).

\subsection*{Case \(\boldsymbol{k=n}\)}

Because
\(\binom{n-c}{\,n\,}=1\) if \(c=0\) and \(0\) otherwise,
\[
  f_{n}(c)
  = 1-\binom{n-c}{n}
  = \mathbf 1_{\{c\ge 1\}}
  \;\in\;\{0,1\},
  \quad\text{hence } f_{n}(c)^{2}=f_{n}(c).
\]

Next we compute the first and second moments.
\[
\begin{aligned}
\mathbb{E}[f_{n}(c)] = \mathbb{E}[f_{n}(c)^{2}]
  &= \sum_{c=0}^{n}
     \mathbf 1_{\{c\ge 1\}}
     \binom{n}{c}\,p^{c}(1-p)^{\,n-c}\\
  &= \sum_{c=1}^{n}\binom{n}{c}\,p^{c}(1-p)^{\,n-c}\\
  &= 1-
     \binom{n}{0}\,p^{0}(1-p)^{\,n}, \quad \text{Since the binomial PMF is normalized}\\
     &= 1-(1-p)^{n}\\
\end{aligned}
\]

Plugging the two moments into (\(\star\)),
\[
  \operatorname{Var}\bigl[f_{n}(c)\bigr]
  \;=\;
  \bigl[1-(1-p)^{n}\bigr]
  \;-\;
  \bigl[1-(1-p)^{n}\bigr]^{2}
  \;=\;
  (1-p)^{\,n}-(1-p)^{\,2n}.
\]

\subsection*{Case \(\boldsymbol{k=1}\)}
\[
  f_{1}(c)
  = 1-\frac{n-c}{n}
  = \frac{c}{n}.
\]

Because \(\mathbb{E}[c]=np\) and \(\operatorname{Var}[c]=np(1-p)\),
\[
\begin{aligned}
\mathbb{E}[f_{1}(c)]
  &= \frac{1}{n}\,\mathbb{E}[c] = p, \quad \text{and} \\
\mathbb{E}\!\bigl[f_{1}(c)^{2}\bigr]
  &= \frac{1}{n^{2}}\,\mathbb{E}[c^{2}]                                     \\[2pt]
  &= \frac{1}{n^{2}}\bigl(\operatorname{Var}[c]+\mathbb{E}[c]^{2}\bigr)      \\[2pt]
  &= \frac{1}{n^{2}}\bigl(np(1-p)+n^{2}p^{2}\bigr)                           \\[2pt]
  &= \frac{p(1-p)}{n}+p^{2}.
\end{aligned}
\]

finally,
\[
  \operatorname{Var}\bigl[f_{1}(c)\bigr]
  \;=\;
  \Bigl(\frac{p(1-p)}{n}+p^{2}\Bigr) - p^{2}
  \;=\;
  \frac{p(1-p)}{n}.
\]
\end{proof}

\section{More Flexible Fitting Methods}\label{more_flexible}

\cite{schaeffer2025largelanguagemonkeyspower} claimed that a standard beta distribution was not flexible enough to fit the distribution of $\passiat{1}$, leading them to model the distribution of $\passiat{k}$ as a scaled beta-binomial rather than a beta-binomial distribution.  The authors developed the discretized fitting method described in Section \ref{rylan_distributional} because they could not find a tractable likelihood for the three-parameter beta-binomial distribution.  

In this section, we derive a tractable likelihood for the scaled beta-binomial distribution, allowing us to avoid estimating $\hat{\theta}$ from \eqref{scaled_beta} using the unprincipled estimator from \eqref{scale_estimator}.  A tractable likelihood also allows us to fit the scaled beta-binomial distribution directly to $n, k_i$ rather than first estimating $\passiat{k}$ and fitting the scaled beta distribution to these estimates.  

We first rewrite the expression for the likelihood of the scaled beta-binomial distribution to remove the integral in the following lemma:

\begin{lemma}
    The likelihood for the scaled beta-binomial distribution is given by 

    \begin{align}
        \frac{1}{\mathrm{Be}(\alpha, \beta)} {\binom{n}{k}} \int_0^\theta p^k(1-p)^{n-k} \left(\frac{p}{\theta}\right)^{\alpha-1}\left(1-\frac{p}{\theta}\right)^{\beta-1} \frac{1}{\theta}dp \\ = \frac{1}{\mathrm{Be}(\alpha, \beta)} {\binom{n}{k}} \sum_{i=0}^{n-k} {\binom{n-k}{i}} (-1)^i \theta^{k+i} \mathrm{Be}(k+i+\alpha, \beta).
    \end{align}
\end{lemma}

The proof can be found in Appendix \ref{scaled_bb_likelihood}.  

Although the resulting likelihood no longer contains an integral, it involves an alternating sum of potentially large terms.  Define \begin{align}
    W_i =  {\binom{n-k}{i}} \theta^{k+i} \mathrm{Be}(k+i+\alpha, \beta).
\end{align}

In terms of $W_i$, our optimization objective is \begin{align}
-\log\left(\sum_{i=0}^{n-k} (-1)^i W_i\right).  
\end{align}

To compute this as stably as possible, we use an alternating log-sum-exp function.  Letting $W_m = \max\{W_0,...,W_{n-k}\}$, our log likelihood becomes:

\begin{align}
    -\log\left(\sum_{i=0}^{n-k} (-1)^i \exp(\log(W_i)  - \log(W_m))\right) - \log(W_m). 
\end{align}

\section{Scaled Beta-Binomial Likelihood} \label{scaled_bb_likelihood}

\begin{align}
    \frac{1}{\mathrm{Be}(\alpha, \beta)} {\binom{n}{k}} \int_0^\theta p^k(1-p)^{n-k} \left(\frac{p}{\theta}\right)^{\alpha-1}\left(1-\frac{p}{\theta}\right)^{\beta-1} \frac{1}{\theta}dp
    \\ = \frac{1}{\mathrm{Be}(\alpha, \beta)} {\binom{n}{k}} \theta^k  \int_0^\theta \left(\frac{p}{\theta}\right)^k(1-p)^{n-k} \left(\frac{p}{\theta}\right)^{\alpha-1}\left(1-\frac{p}{\theta}\right)^{\beta-1} \frac{1}{\theta}dp \\
    = \frac{1}{\mathrm{Be}(\alpha, \beta)} {\binom{n}{k}} \theta^k \sum_{i=0}^{n-k} {\binom{n-k}{i}} \int_0^\theta  (-1)^i p^i \left(\frac{p}{\theta}\right)^{k+\alpha-1}\left(1-\frac{p}{\theta}\right)^{\beta-1} \frac{1}{\theta}dp \\
    = \frac{1}{\mathrm{Be}(\alpha, \beta)} {\binom{n}{k}} \sum_{i=0}^{n-k} {\binom{n-k}{i}} \int_0^\theta  \theta^{k+i} (-1)^i  \left(\frac{p}{\theta}\right)^{k+i+\alpha-1}\left(1-\frac{p}{\theta}\right)^{\beta-1} \frac{1}{\theta}dp \\ = \frac{1}{\mathrm{Be}(\alpha, \beta)} {\binom{n}{k}} \sum_{i=0}^{n-k} {\binom{n-k}{i}} (-1)^i \theta^{k+i} \mathrm{Be}(k+i+\alpha, \beta)  \\ = \frac{1}{\mathrm{Be}(\alpha, \beta)} {\binom{n}{k}} \sum_{i=0}^{n-k} {\binom{n-k}{i}} (-1)^i \theta^{k+i} \mathrm{Be}(k+i+\alpha, \beta)
\end{align}

Define \begin{align}
    W_i =  {\binom{n-k}{i}} \theta^{k+i} \mathrm{Be}(k+i+\alpha, \beta).
\end{align}

Our optimization objective is \begin{align}
-\log\left(\sum_{i=0}^{n-k} (-1)^i W_i\right).  
\end{align}

To compute this as stably as possible, we use an alternating log-sum-exp function.  Letting $W_m = \max\{W_0,...,W_{n-k}\}$, our log likelihood becomes:

\begin{align}
    -\log\left(\sum_{i=0}^{n-k} (-1)^i \exp(\log(W_i)  - \log(W_m))\right) - \log(W_m). 
\end{align}

\begin{align*}
    \passiat{1} \sim \textrm{Beta}(\alpha, \beta, \theta)\\
    k_i \sim \textrm{Binomial}(n, \passiat{1})
\end{align*}

\section{Optimal Distribution of Samples} \label{sec:opt-distribution}

\subsection{Proofs}

\begin{lemma}[Variance in the Asymptotic Regime] \label{lem:asympt-var}
    For a sequence of random random variables $\{x_n\}$ such that $x_n = y_n/n$ where $y_n \sim \text{Bin}(n, p)$, we have the following:
    \[
        \sqrt{n} ((1 - x_n)^k - (1 - p)^k) \dto \N(0, pk^2(1-p)^{2k-1})
    \]
\end{lemma}

\begin{proof}
    By the Central Limit Theorem,
    \begin{align}
        \sqrt{n} ((1 - x_n) - (1 - p)) &\dto \N(0, p(1-p))
    \end{align}
    Let $g: \R \rightarrow \R$ be defined as follows:
    \[
        g(t) = t^k
    \]
    Applying the delta method:
    \begin{align}
        \sqrt{n} ((1 - x_n)^k - (1 - p)^k) &\dto \N(0, g'(1 - p)^2p(1-p)) \\
         &\dto \N(0, (k(1 - p)^{k-1})^2p(1-p)) \\
         &\dto \N(0, pk^2(1-p)^{2k-1})
    \end{align}    
\end{proof}

\begin{lemma}[Variance-Minimizing Budget] \label{lem:varmin-budget}
    Consider a random variable $X = \sum_{i=1}^m X_i$ where each $X_i$ is an independent random variable with variance $\mathrm{Var}(X_i) = v_i / b_i$.
    
    Consider the positive scaled simplex $B = \{ b : b_i > 0 \And \sum_j^m b_j = B \}$.
    We have the following:
    \begin{align}
        b^* &= \arg\min_{b \in B} \mathrm{Var}(X; b) \\
        b^*_i &= \frac{\sqrt{v_i}}{\sum_j^m \sqrt{v_j}}
    \end{align}
\end{lemma}

\begin{proof}
    Our objective is this:
    \[
        \min_{b_i > 0} \sum_{i=1}^m v_i/b_i \, \text{ s.t. } \, \sum_{i=1}^m b_j = B
    \]
    This objective is convex as a sum of convex functions, meaning we can use the Lagrange method:
    \begin{align}
        \mathcal{L}(b, \lambda) = \sum_{i=1}^m v_i/b_i + \lambda\left(\sum_{i=1}^m b_i - B\right)
    \end{align}
    Applying first order conditions we get the following:
    \begin{align}
        \frac{\partial \mathcal{L}}{\partial b_i} &= -v_i/b_i^2 + \lambda \\
        0 &= -v_i/b_i^2 + \lambda \\
        b_i &= \sqrt{v_i / \lambda} \\
        b_i &\propto \sqrt{v_i}
    \end{align}
\end{proof}

Combining Lemma \ref{lem:asympt-var} and Lemma \ref{lem:varmin-budget}, we have Lemma \ref{thm:optimal-rate}.


\subsection{Synthetic Comparison of Uniform and Dynamic Sampling}\label{sampling_ablations}

We demonstrate the gains possible with dynamic sampling via the following contrived scenario: half of the problems are ``easy'' ($\passiat{1} = 0.3$) and half of the problems are ``impossible`` ($\passiat{1} = 0$).
In this instance, we expect $\passat{k} \rightarrow 1/2$ as $k \rightarrow \infty$.
However, without a sufficient allocation of samples to the ``impossible'' problems, the uniform sampling strategy prevents our estimator from determining whether these problems are impossible or just hard (i.e., still likely to be solved in $k$ attempts).
This results in an upwards-biased estimate and relatively slow improvement of MSE as the budget grows.

\begin{figure}[H]
    \centering
    \includegraphics[width=0.7\linewidth]{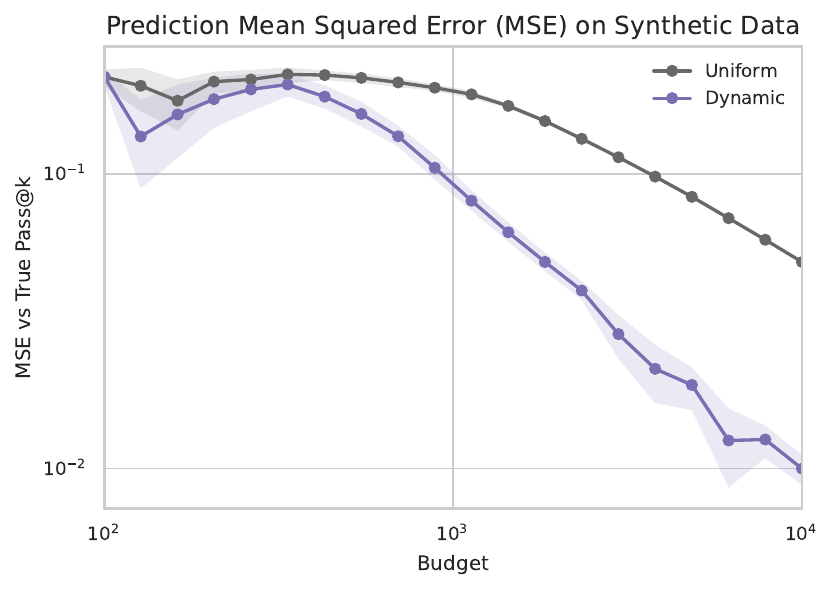}
    \caption{
    The MSE of our estimator with both dynamic and uniform sampling strategies given the described synthetic problem success probabilities, $n=100$ problems and $k=1\,000$.
    By focusing on the most difficult problems, the dynamic strategy allows our estimator to converge rapidly to the true $\passat{k}$ value.
    }
    \label{fig:fig:synthetic-mse-main}
\end{figure}

We also provide some insight into the distributions for which dynamic sampling has advantages over uniform.  We find that for uniform difficulty distributions or distributions that contain a handful of very hard outlier problems, dynamic sampling provides the most advantage.  For distributions with many (or mostly) difficult problems, dynamic sampling holds little to no advantage over uniform sampling, since in these cases, uniform and dynamic sampling distribute the budget very similarly.  If only a handful of problems are quickly solved, then dynamic sampling has very little extra samples to allocate to the more difficult problems.

\begin{table}[h]
\centering
\begin{tabular}{lcc}
\hline
\textbf{Method} & \textbf{Uniform Sampling} & \textbf{Dynamic Sampling} \\
\hline
Beta-Binomial   & 0.000017 [0.000006, 0.000030] & 0.000006 [0.000001, 0.000012] \\
Discretized Beta & 0.030204 [0.023471, 0.037151] & 0.002609 [0.001329, 0.004156] \\
\hline
\end{tabular}
\caption{
The MSE of our estimator and the discretized-Beta estimator with both dynamic and uniform sampling strategies given synthetic problem success probabilities sampled from Uniform$([0,1])$ for $n=64$ problems, $k=1\,000$, and the budget fixed at $256$.
}
\end{table}

\begin{figure}[H]
    \centering
    \includegraphics[width=0.7\linewidth]{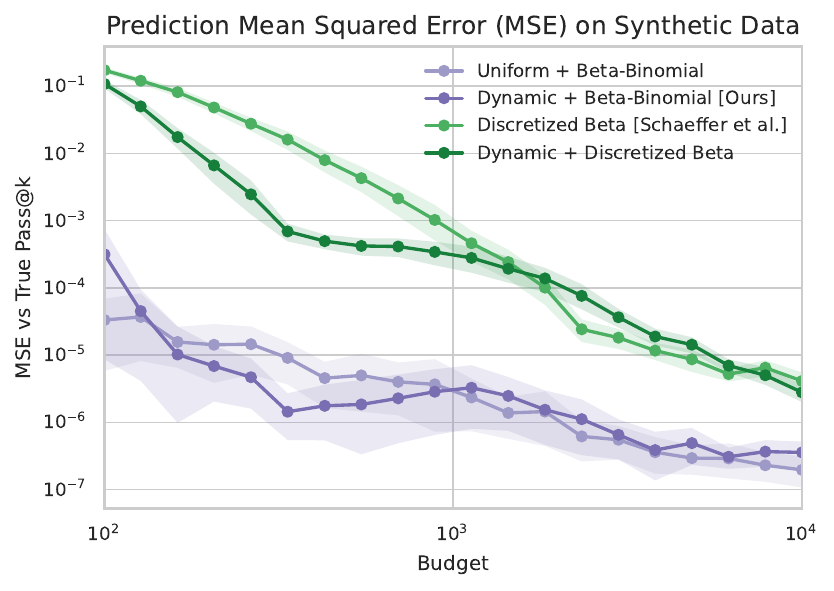}
    \caption{
    The MSE of our estimator and the discretized-Beta estimator with both dynamic and uniform sampling strategies given synthetic problem success probabilities sampled from Uniform$([0,1])$ for $n=64$ problems and $k=1\,000$.
    We observe that dynamic sampling leads to modest improvements in both estimators.
    }
    \label{fig:synthetic-mse-full}
\end{figure}

\section{Additional Figures} \label{additional_figures}

We provide matching figures from the main paper for the benchmarks that were omitted due to lack of space.
Additionally, we include plots that track the scaling of mean squared error (MSE) as budget increases for fixed k, and figures containing the performance of a mixed continuous-discrete model with a discrete lump of probability at $pass@1 = 0$

\clearpage

\begin{figure}[p]
    \centering
    \includegraphics[width=\linewidth,height=0.82\textheight,keepaspectratio]{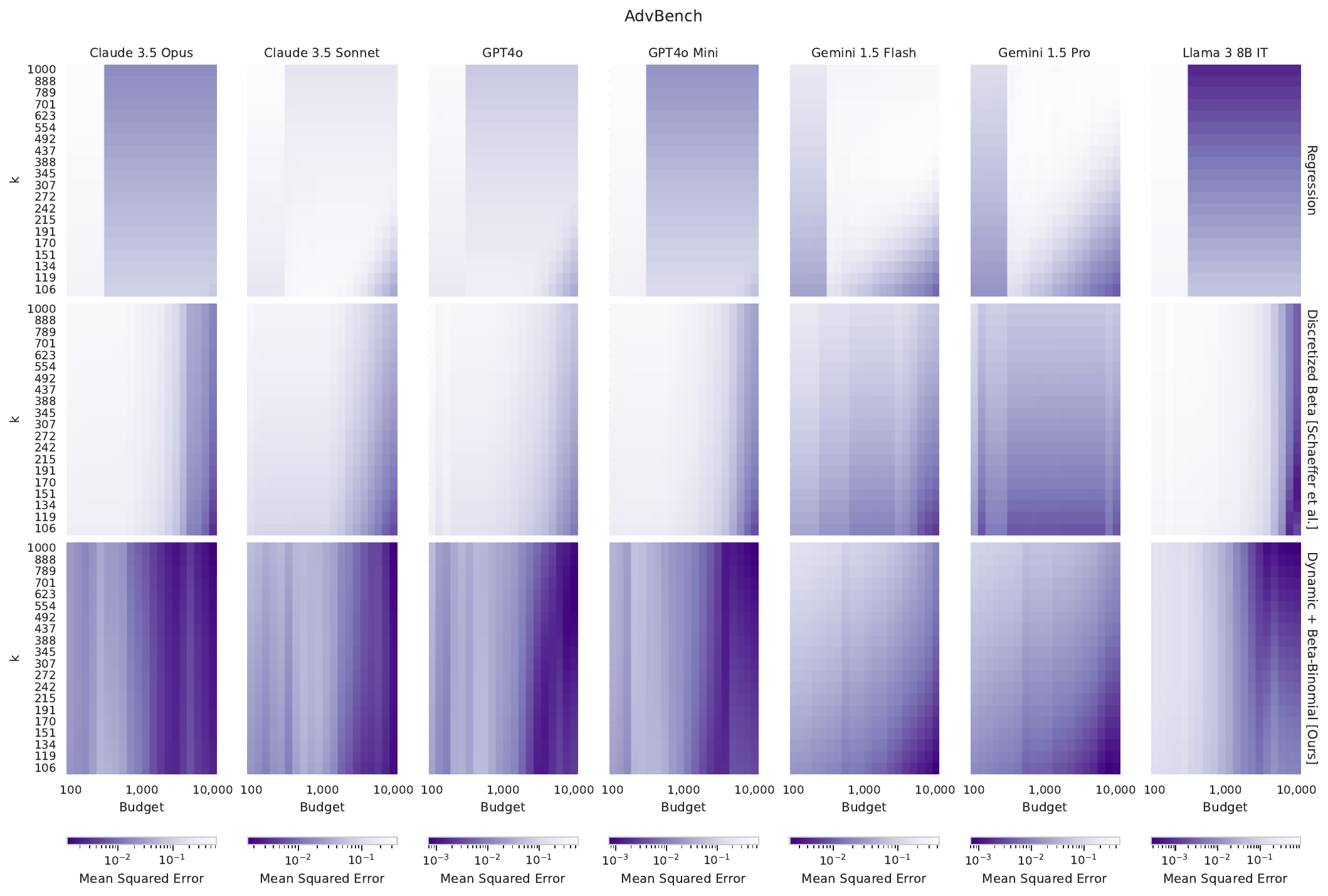}
    \caption{Heatmap depicting how predictions of $\passat{k}$ change with the sampling budget and $k$ for the AdvBench benchmark.
    Darker colors indicate lower mean squared error.}
    \label{fig:jailbreaking_heatmap}
\end{figure}
\clearpage

\begin{figure}[p]
    \centering
    \includegraphics[width=\linewidth,height=0.82\textheight,keepaspectratio]{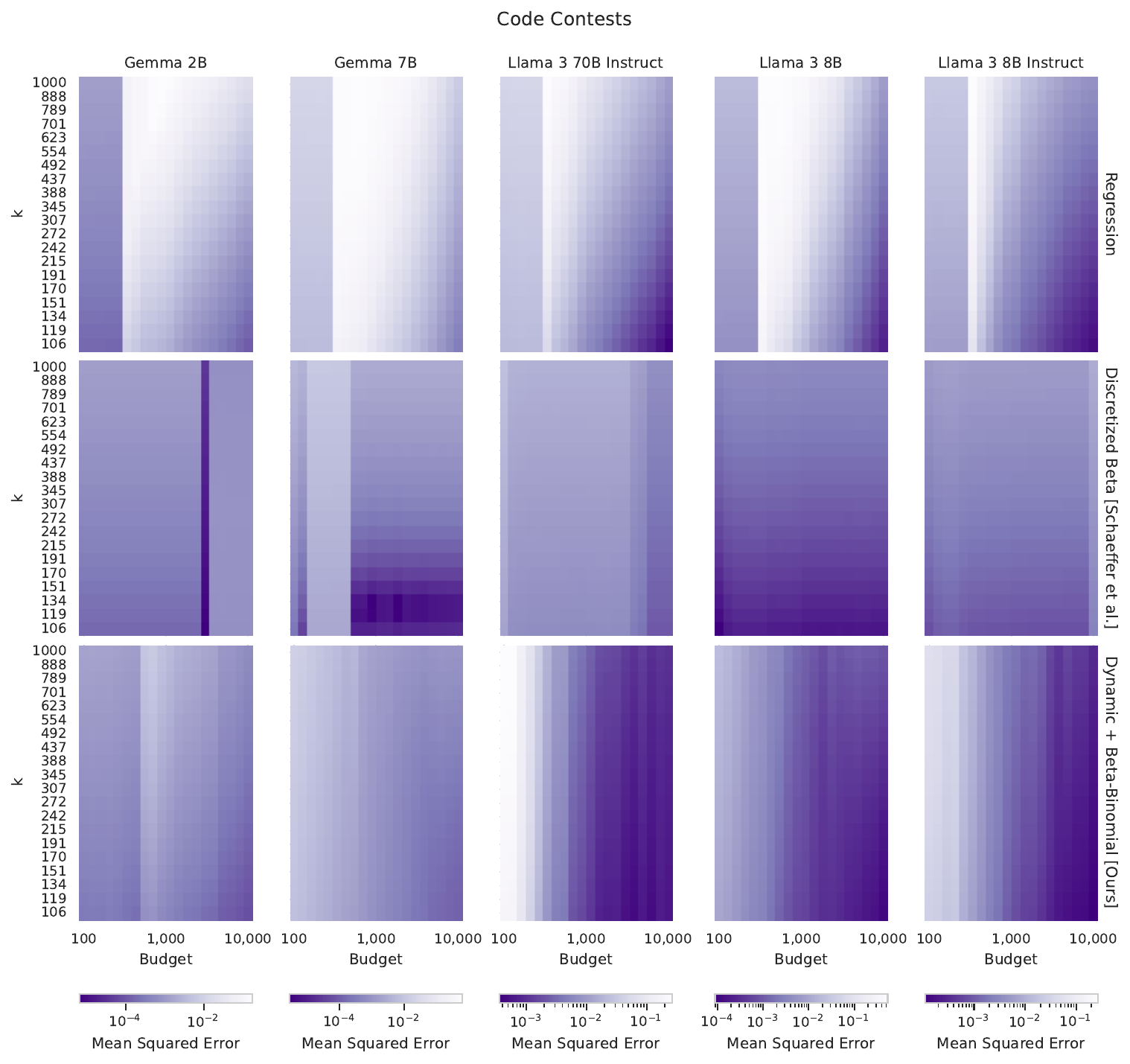}
    \caption{Heatmap depicting how predictions of $\passat{k}$ change with the sampling budget and $k$ for the Code Contests benchmark.
    Darker colors indicate lower mean squared error.}
    \label{fig:code_contests_heatmap}
\end{figure}
\clearpage

\begin{figure}[p]
    \centering
    \includegraphics[width=\linewidth,height=0.82\textheight,keepaspectratio]{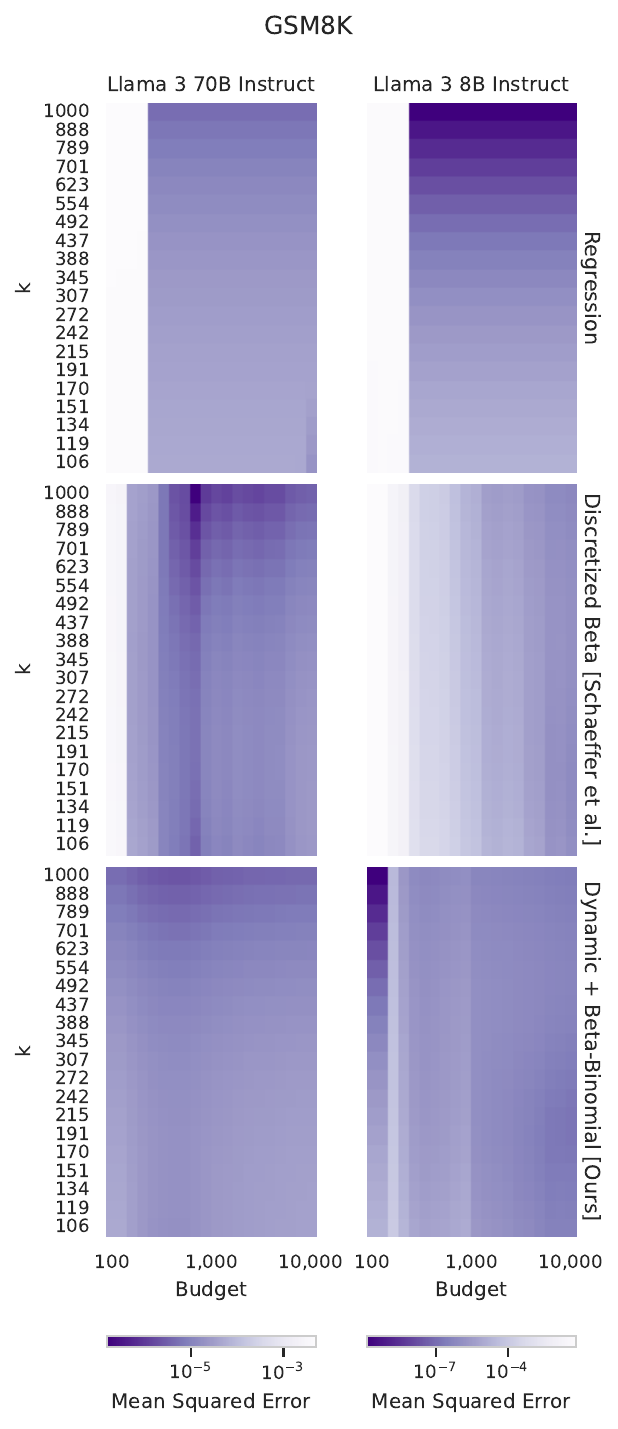}
    \caption{Heatmap depicting how predictions of $\passat{k}$ change with the sampling budget and $k$ for the GSM8K benchmark.
    Darker colors indicate lower mean squared error.}
    \label{fig:gsm8k_heatmap}
\end{figure}
\clearpage

\begin{figure}[p]
    \centering
    \includegraphics[width=\linewidth,height=0.82\textheight,keepaspectratio]{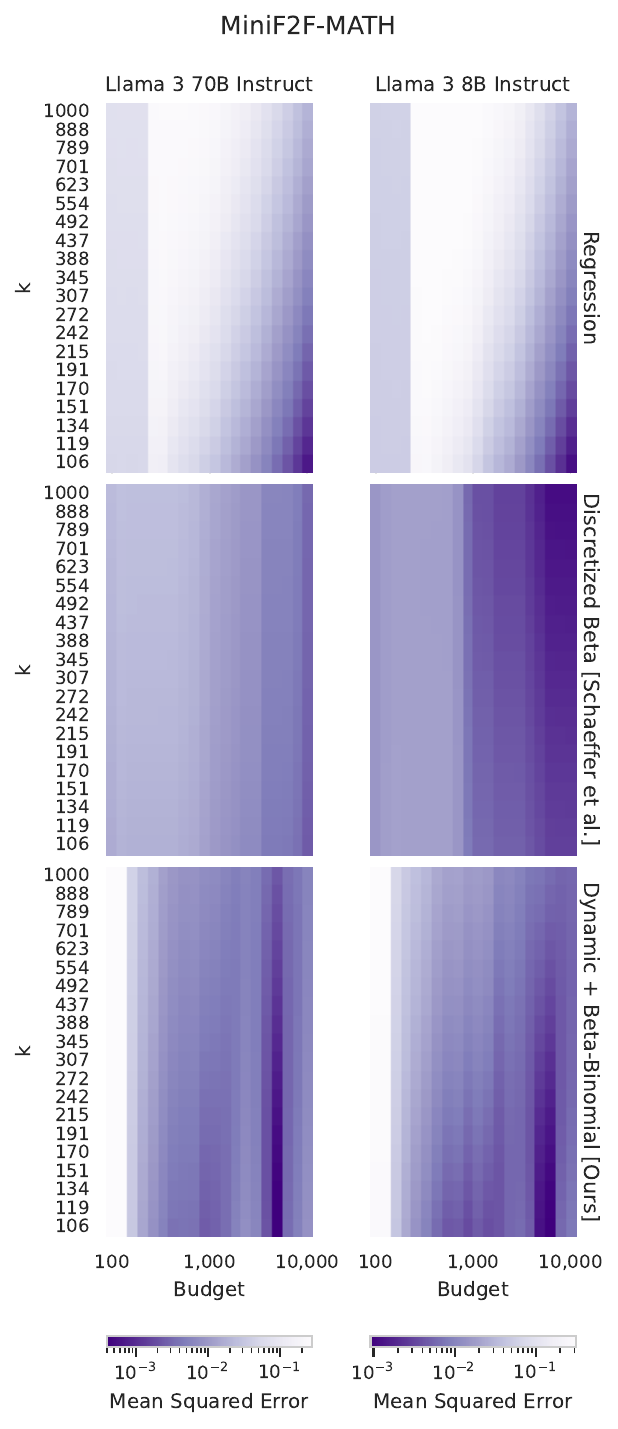}
    \caption{Heatmap depicting how predictions of $\passat{k}$ change with the sampling budget and $k$ for the MiniF2F-MATH benchmark.
    Darker colors indicate lower mean squared error.}
    \label{fig:minif2f_heatmap}
\end{figure}
\clearpage
\begin{figure}[!htbp]
    \centering
    \begin{subfigure}[t]{0.8\linewidth}
        \centering
        \includegraphics[width=0.8\linewidth]{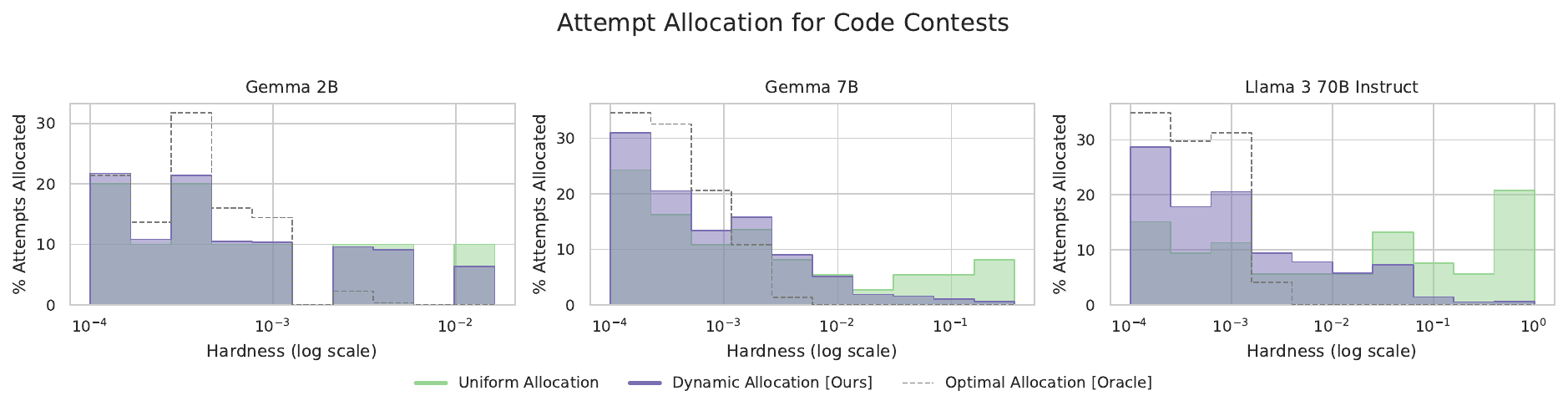}
        \label{fig:allocation-code}
    \end{subfigure}

    \vspace{0.5em}

    \begin{subfigure}[t]{0.8\linewidth}
        \centering
        \includegraphics[width=0.8\linewidth]{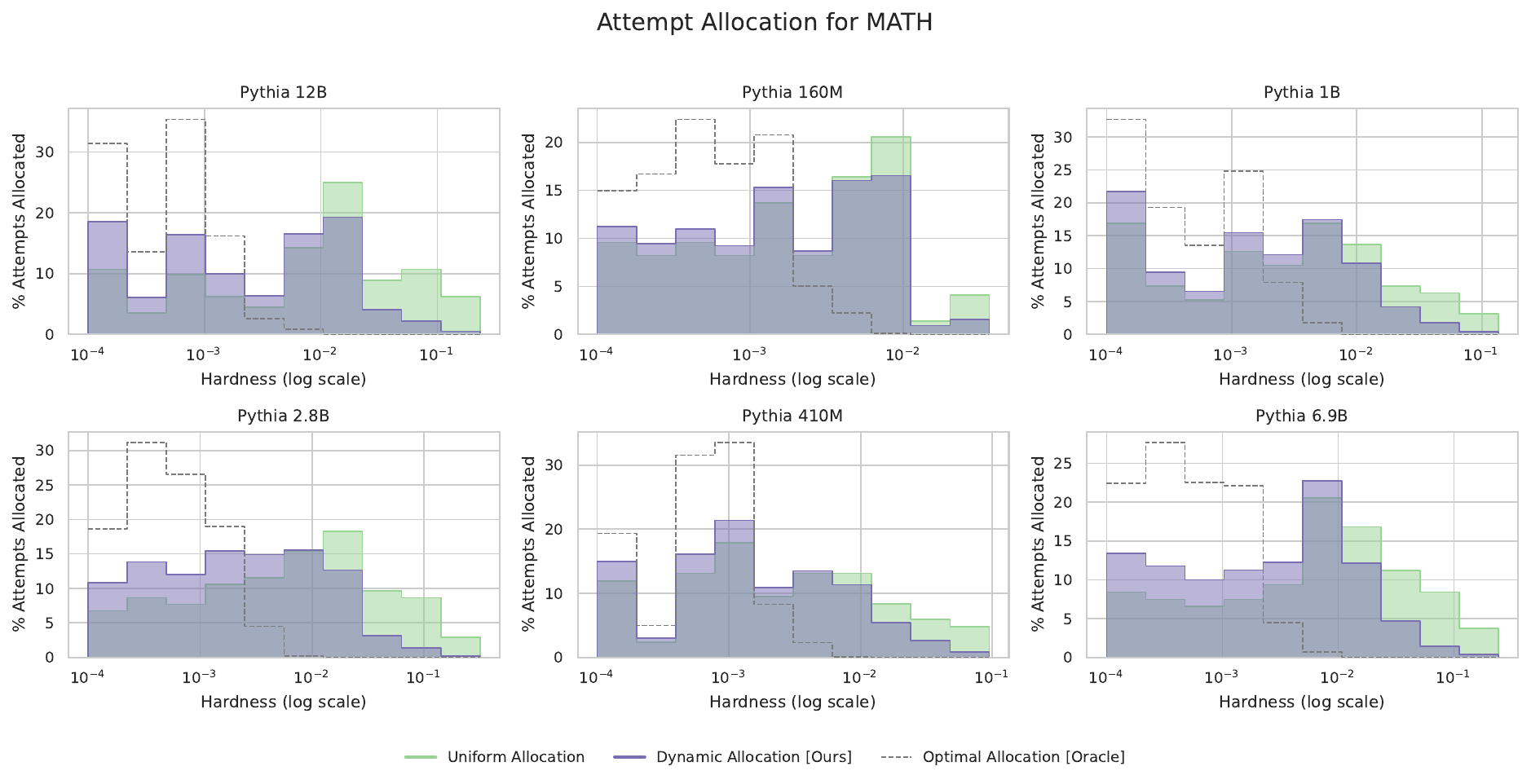}
        \label{fig:allocation-math}
    \end{subfigure}

    \caption{Contrasted distributions of problem success probabilities for the problems selected by dynamic and uniform sampling strategies on Code Contests and MATH.}
\end{figure}

\clearpage
\begin{figure}[!htbp]
    \centering
    \begin{subfigure}[t]{0.8\linewidth}
        \centering
        \includegraphics[width=0.8\linewidth]{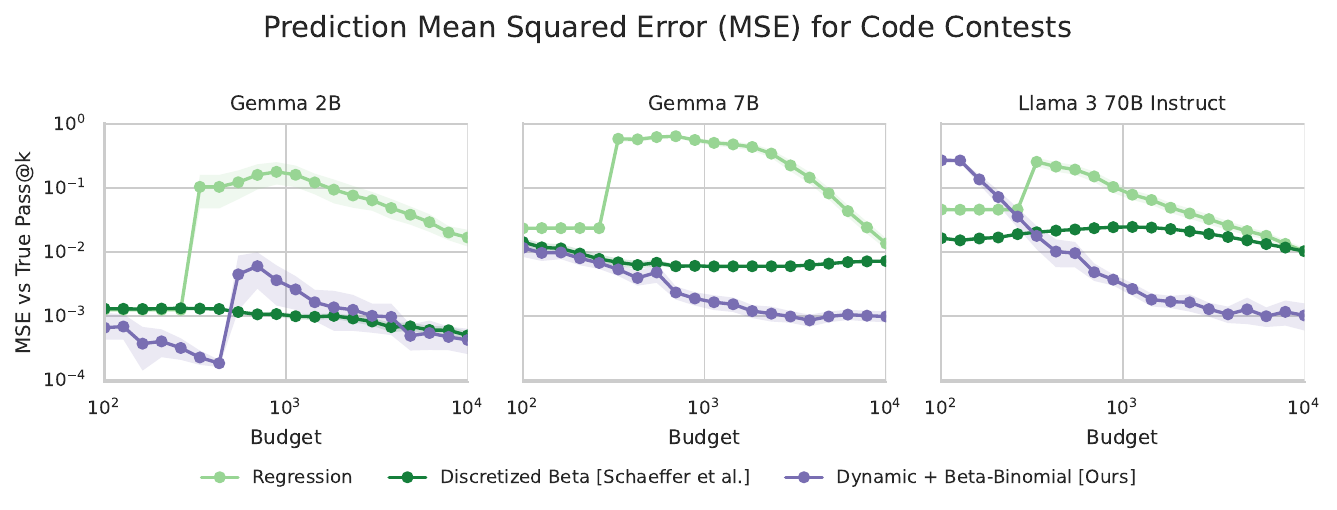}
        \label{fig:mse-code}
    \end{subfigure}

    \vspace{0.5em}

    \begin{subfigure}[t]{0.8\linewidth}
        \centering
        \includegraphics[width=0.8\linewidth]{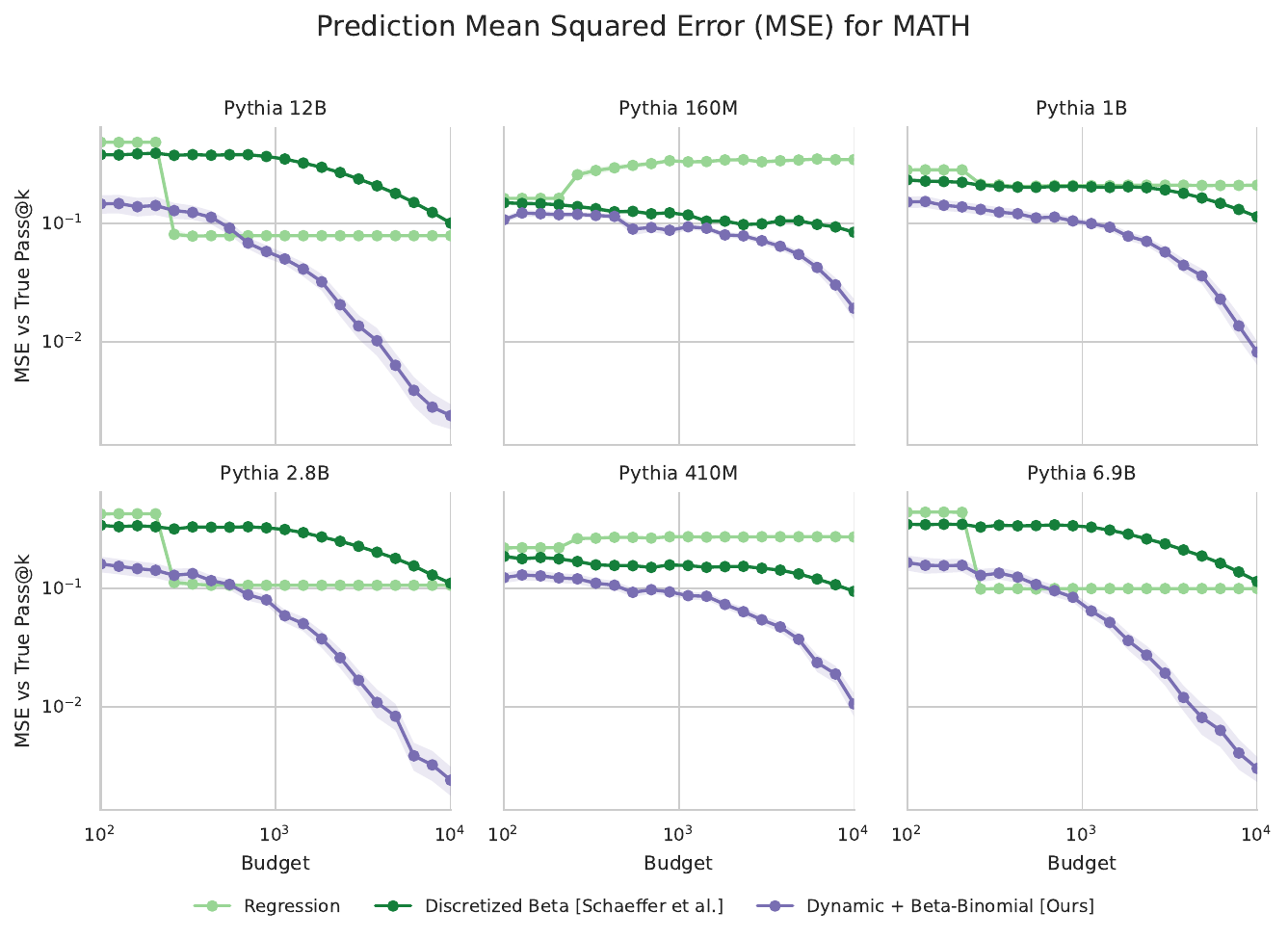}
        \label{fig:mse-math}
    \end{subfigure}

    \vspace{0.5em}

    \begin{subfigure}[t]{0.8\linewidth}
        \centering
        \includegraphics[width=0.8\linewidth]{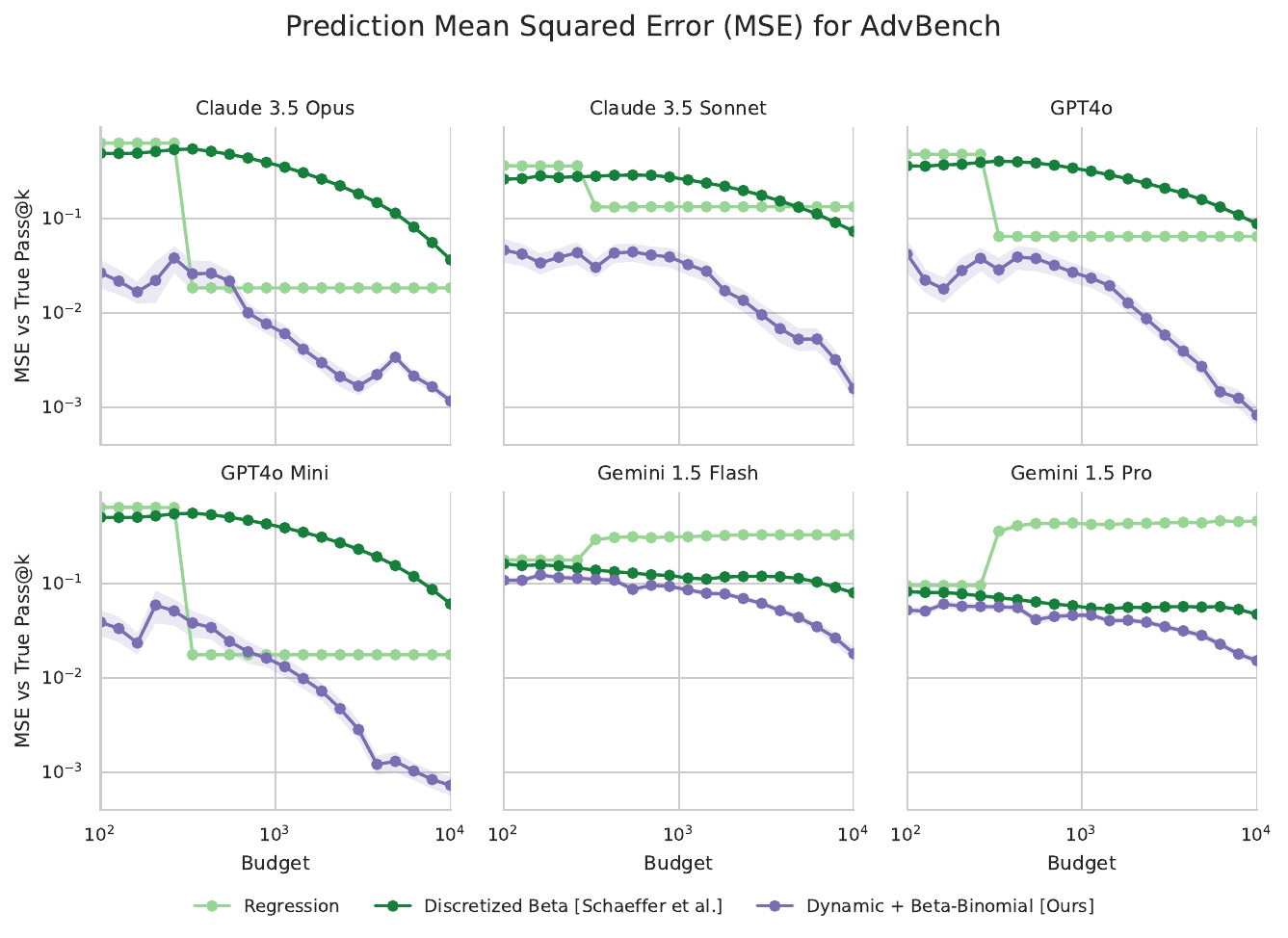}
        \label{fig:mse-jailbreaking}
    \end{subfigure}

    \caption{MSE scaling with increasing budget.
        As expected, more samples generally leads to a reduction in MSE across all approaches.
        For some models our approach reaches MSE more than 10x lower than its counterparts.}
\end{figure}

\begin{figure}
    \centering
    \begin{subfigure}[t]{1.0\textwidth}
        \centering
        \includegraphics[width=1.0\linewidth]{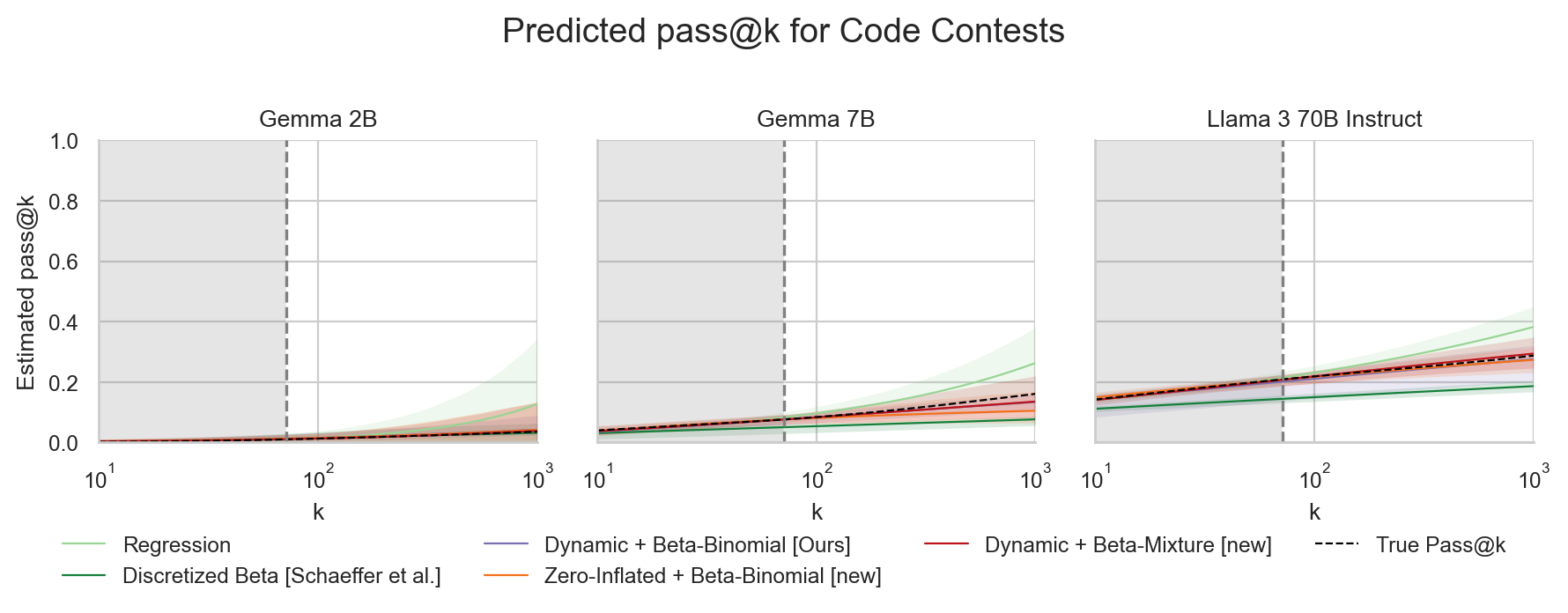}
        \label{fig:p_at_k_code}
    \end{subfigure}

    \caption{
    $pass@k$ plots comparing the performance of the mixed continuous-discrete model with a discrete lump of probability at $pass@1 = 0$ (dynamic sampling) to other methods on code contest data.
    }
\end{figure}
\begin{figure}
    \centering
    \begin{subfigure}[t]{1.0\textwidth}
        \centering
        \includegraphics[width=1.0\linewidth]{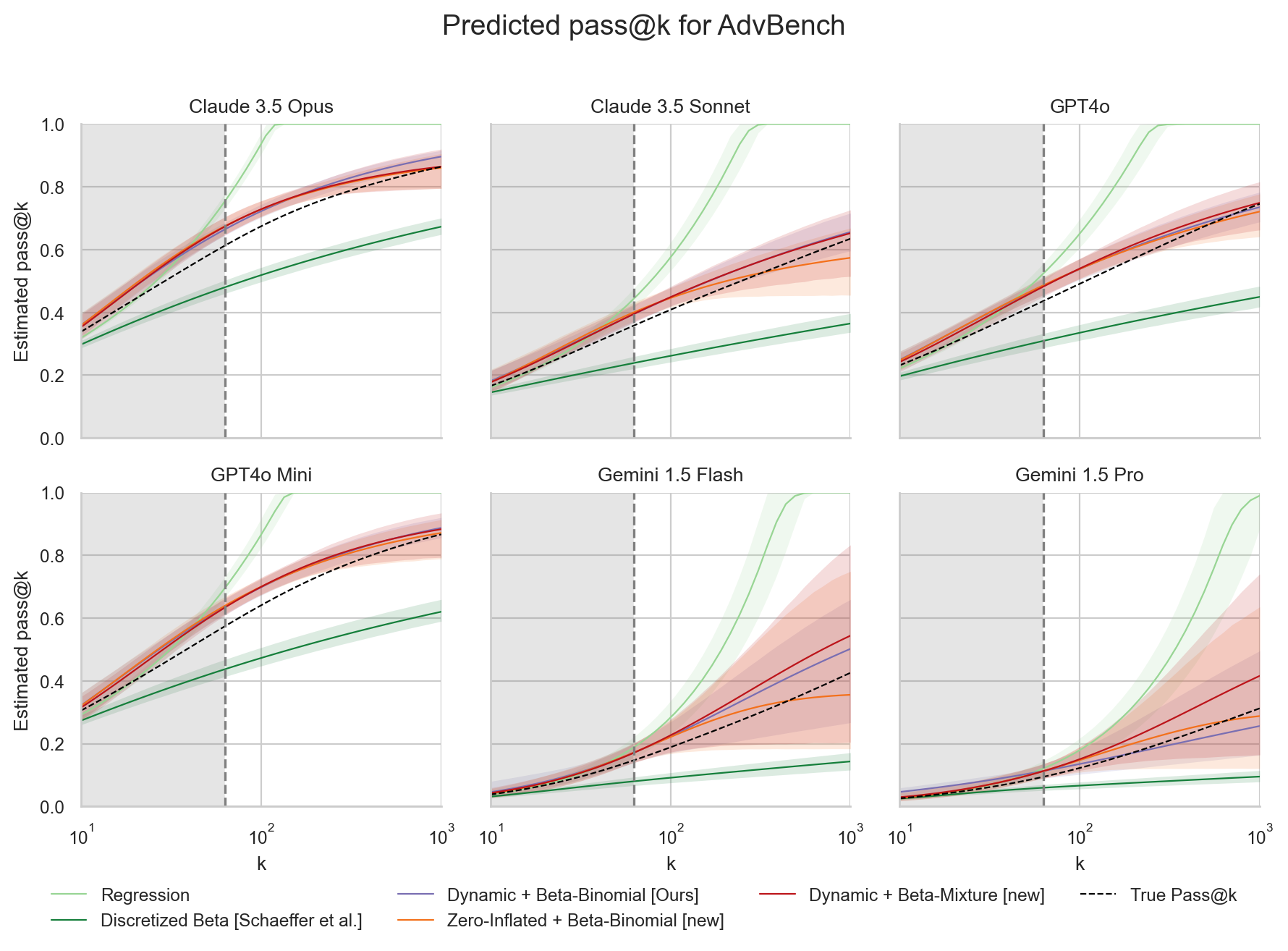}
        \label{fig:p_at_k_jail}
    \end{subfigure}

    \caption{
    $pass@k$ plots comparing the performance of dynamic sampling applied to the mixed continuous-discrete model with a discrete lump of probability at $pass@1 = 0$ \citep{zellinger2025rational} to other methods on BON Jailbreaking data. 
    }
\end{figure}
\begin{figure}
    \centering
    \begin{subfigure}[t]{1.0\textwidth}
        \centering
        \includegraphics[width=1.0\linewidth]{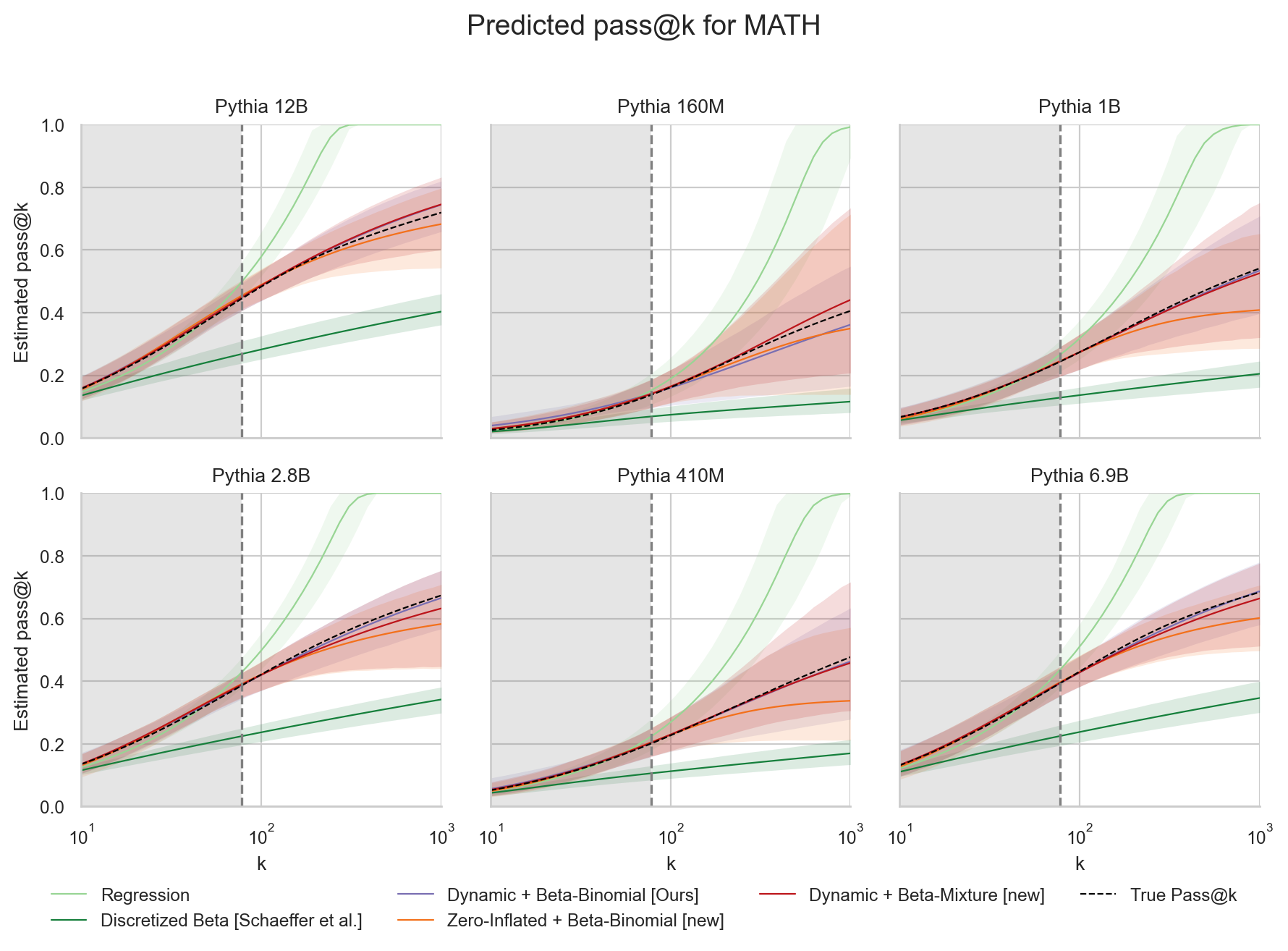}
        \label{fig:p_at_k_math}
    \end{subfigure}

    \caption{
    $pass@k$ plots comparing the performance of dynamic sampling applied to the mixed continuous-discrete model with a discrete lump of probability at $pass@1 = 0$ \citep{zellinger2025rational} to other methods on MATH data.    }
\end{figure}

\begin{figure}
    \centering
    \begin{subfigure}[t]{1.0\textwidth}
        \centering
        \includegraphics[width=1.0\linewidth]{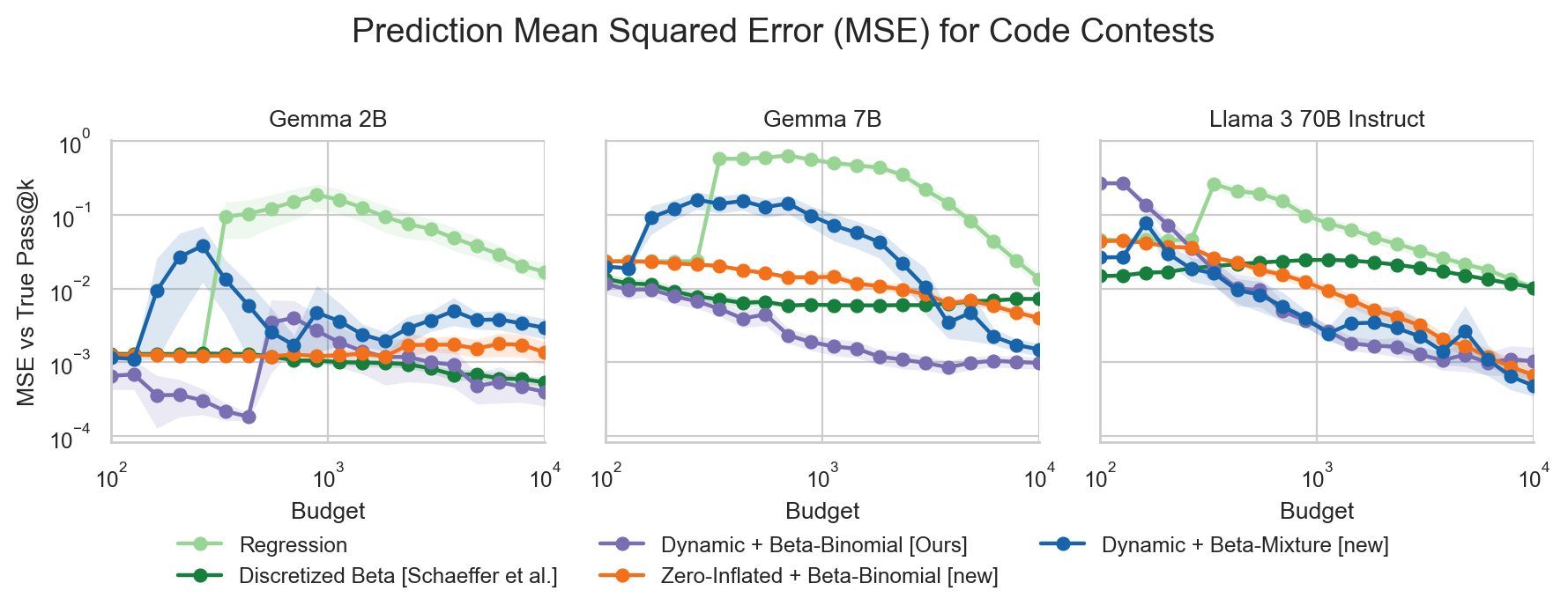}
        \label{fig:mse_code}
    \end{subfigure}

    \caption{
    MSE plots comparing the performance of the mixed continuous-discrete model with a discrete lump of probability at $pass@1 = 0$ (dynamic sampling) to other methods on code contest data.
    }
\end{figure}
\begin{figure}
    \centering
    \begin{subfigure}[t]{1.0\textwidth}
        \centering
        \includegraphics[width=1.0\linewidth]{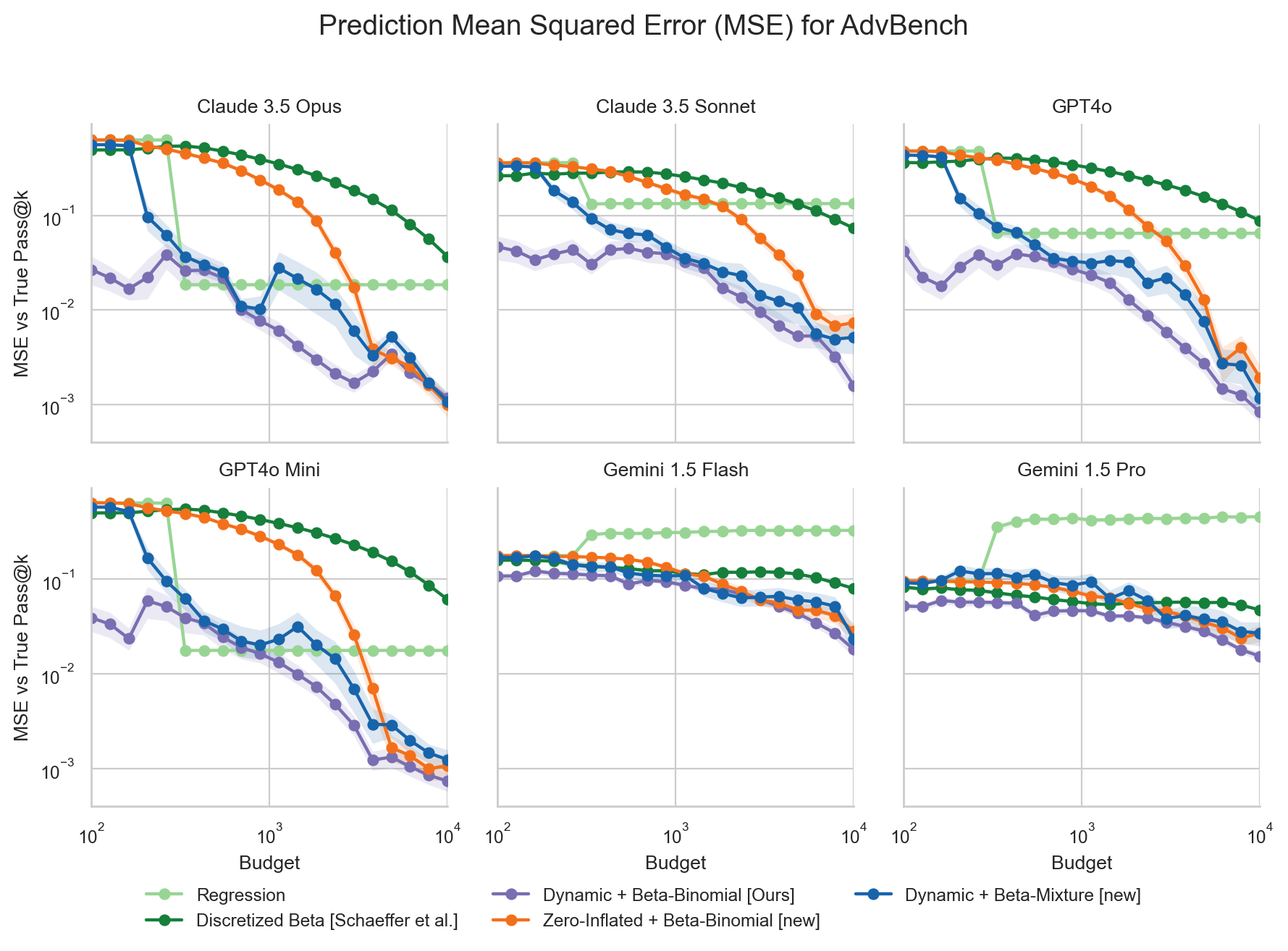}
        \label{fig:mse_jail}
    \end{subfigure}

    \caption{
    MSE plots comparing the performance of the mixed continuous-discrete model with a discrete lump of probability at $pass@1 = 0$ (dynamic sampling) to other methods on jail-breaking data.
    }
\end{figure}
\begin{figure}
    \centering
    \begin{subfigure}[t]{1.0\textwidth}
        \centering
        \includegraphics[width=1.0\linewidth]{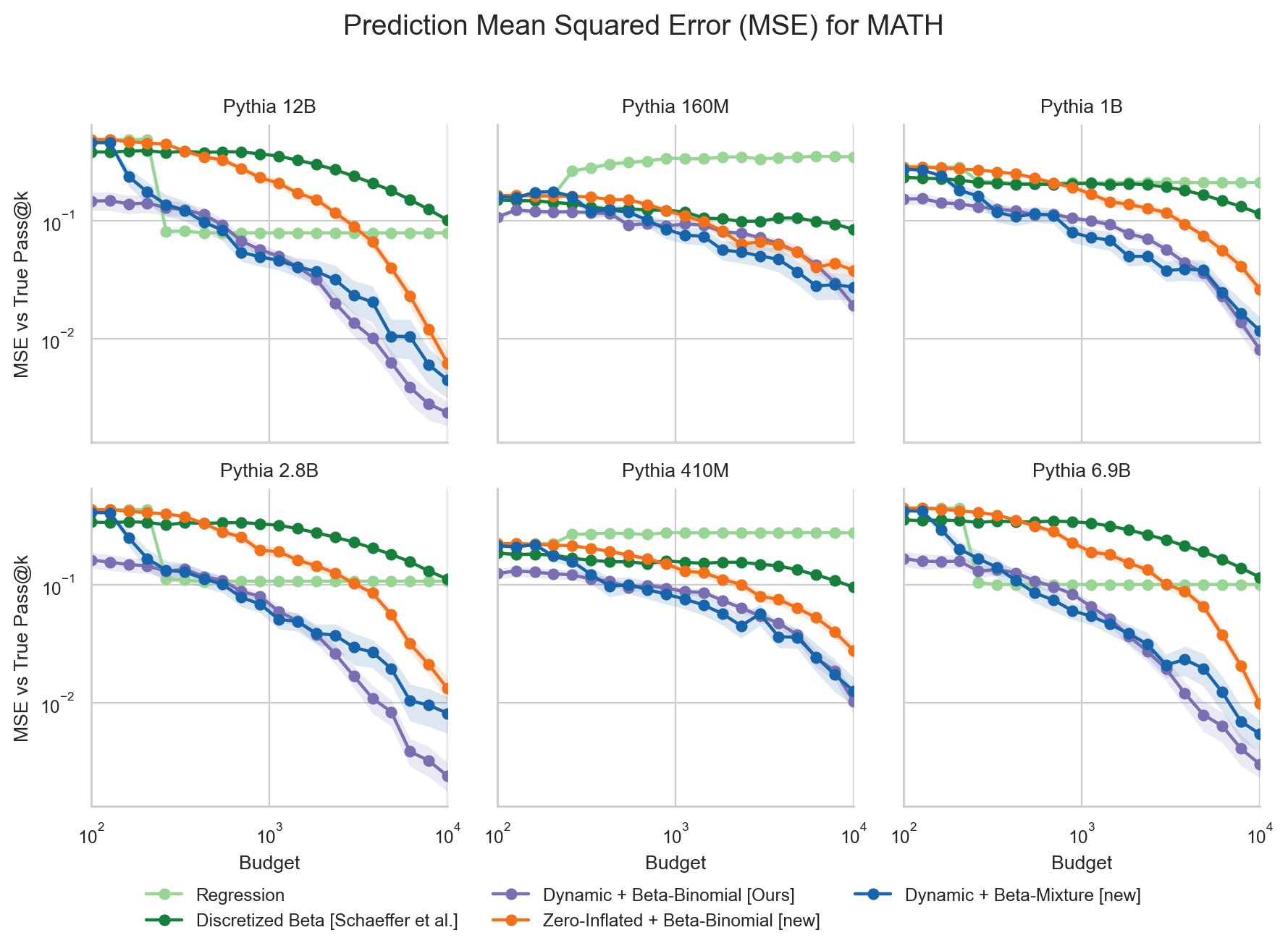}
        \label{fig:mse_math}
    \end{subfigure}

    \caption{
    MSE plots comparing the performance of the mixed continuous-discrete model with a discrete lump of probability at $pass@1 = 0$ (dynamic sampling) to other methods on math data.
    }
\end{figure}

\section{Datasets}
\label{sec:benchmarks}

We draw our evaluation data from two recent sources: \cite{brown2024largelanguagemonkeysscaling} and \cite{hughes2024bestofnjailbreaking}.
They record, for each of 128 prompt samples, the \textbf{number of successful outcomes out of $10\,000$ trials}.
These prompts are sampled from three benchmark suites:

\begin{itemize}
  \item \textbf{CodeContests}~\citep{code_contests}: A competitive programming benchmark which collects description-to-code tasks from contest platforms such as AtCoder, CodeChef, Codeforces, and HackerEarth. Models are evaluated on precise correctness via test cases. Later refinements (e.g.\ CodeContests+) improve test case generation and validation to reduce false positives in evaluation.  
  \item \textbf{MATH}~\citep{hendrycksmath2021}: A mathematical reasoning dataset of 12,500 high school competition problems (e.g.\ AMC, AIME). Each problem comes with a full solution path and final answer. The benchmark evaluates model proficiency in multi-step reasoning across domains such as algebra, number theory, geometry, and combinatorics.  
  \item \textbf{AdvBench}~\citep{zou2023universal}: An adversarial NLP benchmark oriented toward security tasks. It emphasizes realistic attacker goals and evaluates models’ success or failure under adversarial prompting strategies.  
\end{itemize}

This combination lets us evaluate the efficacy of our estimator on problem success probability distributions extracted from \textbf{coding}, \textbf{mathematical reasoning}, and \textbf{adversarial robustness} domains.

\clearpage
\section{Schaeffer et al., Comparison Experiments}
\label{sec:schaeffer-comparisons}

We compare the estimators~\citep{schaeffer2025largelanguagemonkeyspower} under uniform and dynamic sampling. Each success probability is drawn from a Beta or Kumaraswamy distribution with shape $(a,b)$, then multiplied by $s$.
All 32 settings use $k=1000$ and 30 seeds; shading shows 95\% bootstrap intervals for mean squared error.

\begin{figure}[!htbp]
    \centering
    \textbf{Scaled Beta: $a=0.15$}\par\smallskip
    \begin{tabular}{@{}cc@{}}
        $m=64$ & $m=128$ \\[-2pt]
        \includegraphics[width=0.47\linewidth,height=0.19\textheight,keepaspectratio]{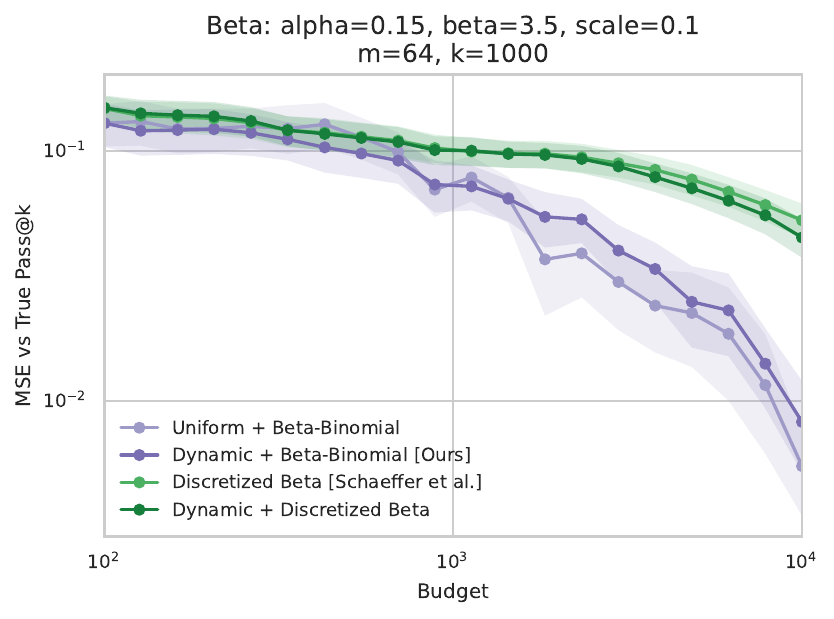} &
        \includegraphics[width=0.47\linewidth,height=0.19\textheight,keepaspectratio]{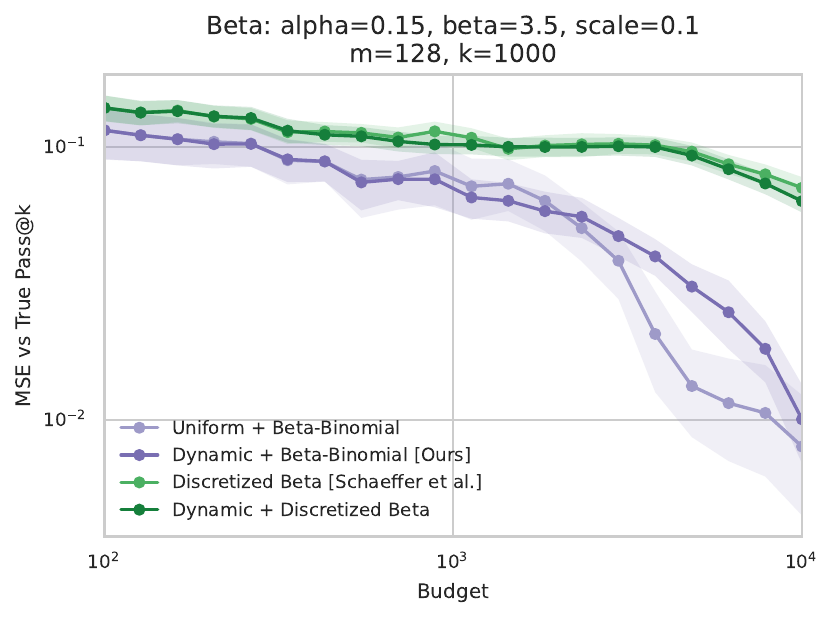} \\[2pt]
        \includegraphics[width=0.47\linewidth,height=0.19\textheight,keepaspectratio]{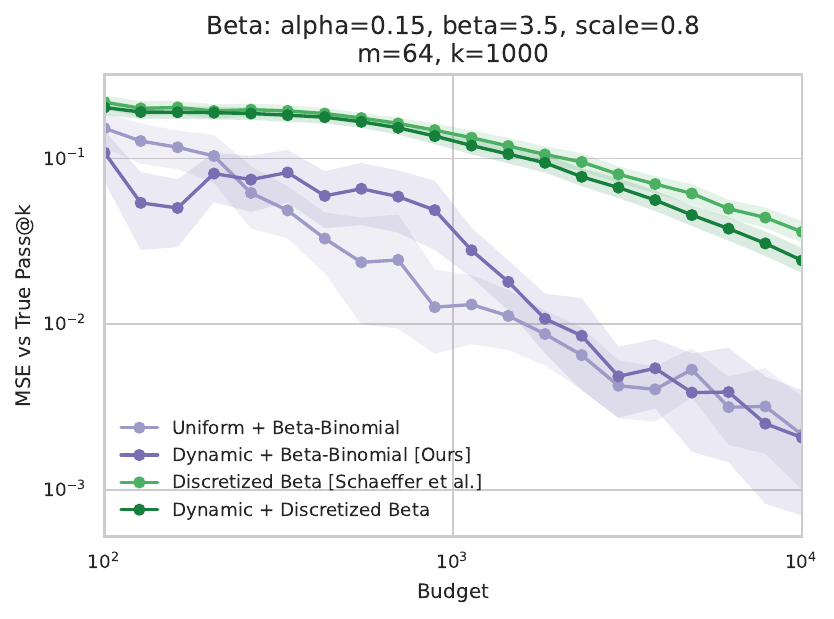} &
        \includegraphics[width=0.47\linewidth,height=0.19\textheight,keepaspectratio]{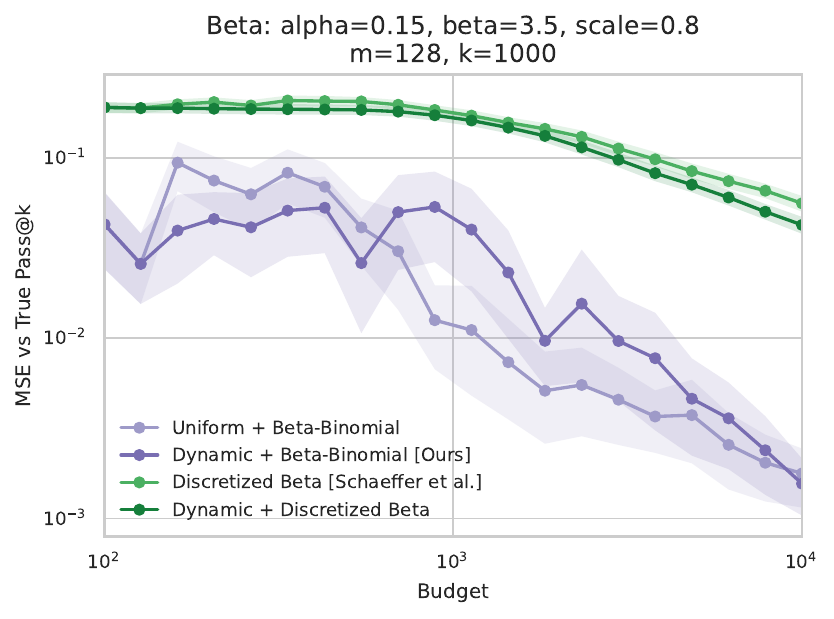} \\[2pt]
        \includegraphics[width=0.47\linewidth,height=0.19\textheight,keepaspectratio]{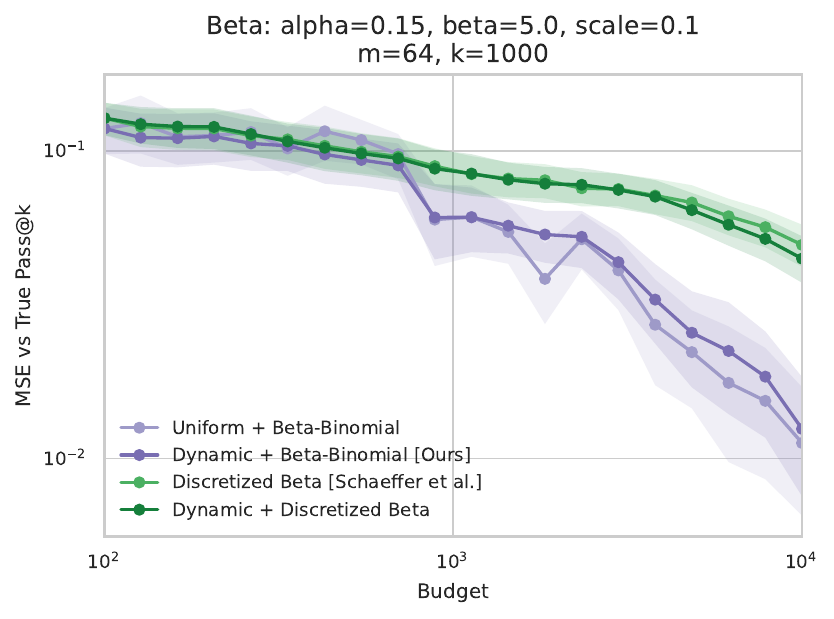} &
        \includegraphics[width=0.47\linewidth,height=0.19\textheight,keepaspectratio]{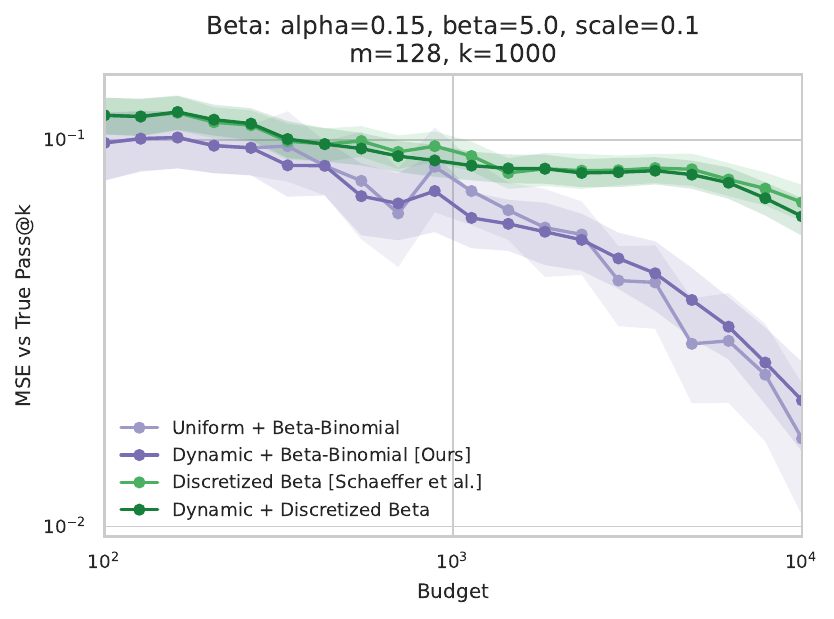} \\[2pt]
        \includegraphics[width=0.47\linewidth,height=0.19\textheight,keepaspectratio]{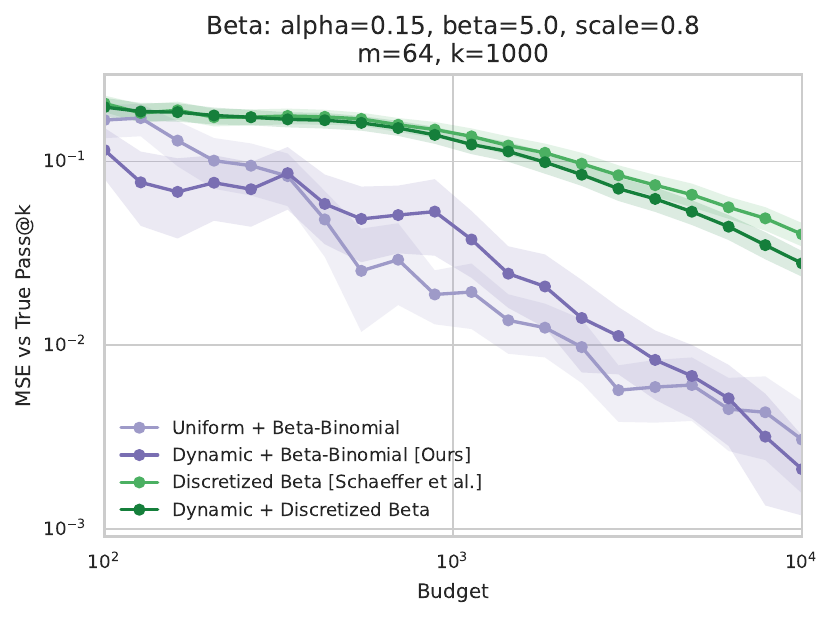} &
        \includegraphics[width=0.47\linewidth,height=0.19\textheight,keepaspectratio]{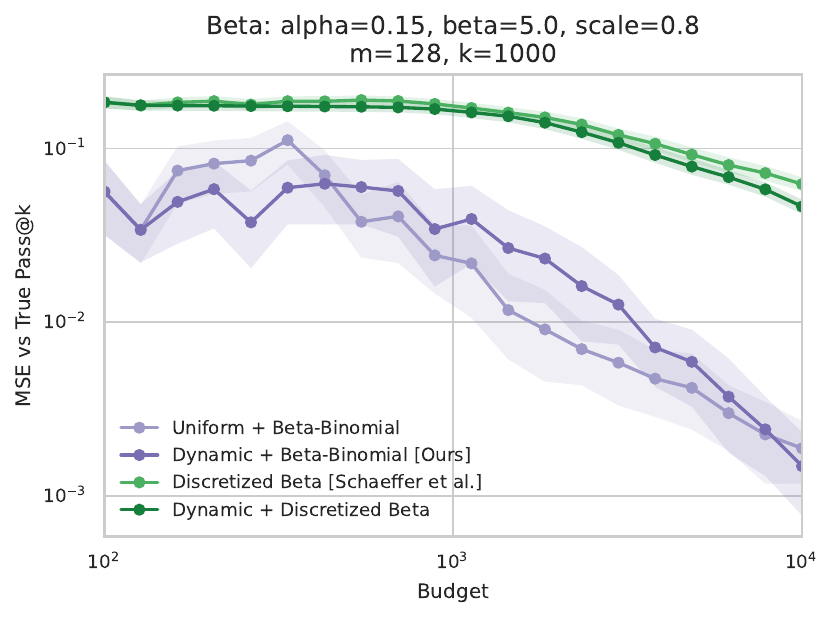} \\[2pt]
    \end{tabular}
    \caption{MSE comparisons for scaled Beta success probabilities with $a=0.15$ at $k=1000$. Columns: $m=64$ and $m=128$ problems. Rows, from top to bottom: $(b,s)=(3.5,0.1)$, $(3.5,0.8)$, $(5.0,0.1)$, and $(5.0,0.8)$.}
    \label{fig:schaeffer-beta-a015}
\end{figure}
\clearpage

\begin{figure}[!htbp]
    \centering
    \textbf{Scaled Beta: $a=0.4$}\par\smallskip
    \begin{tabular}{@{}cc@{}}
        $m=64$ & $m=128$ \\[-2pt]
        \includegraphics[width=0.47\linewidth,height=0.19\textheight,keepaspectratio]{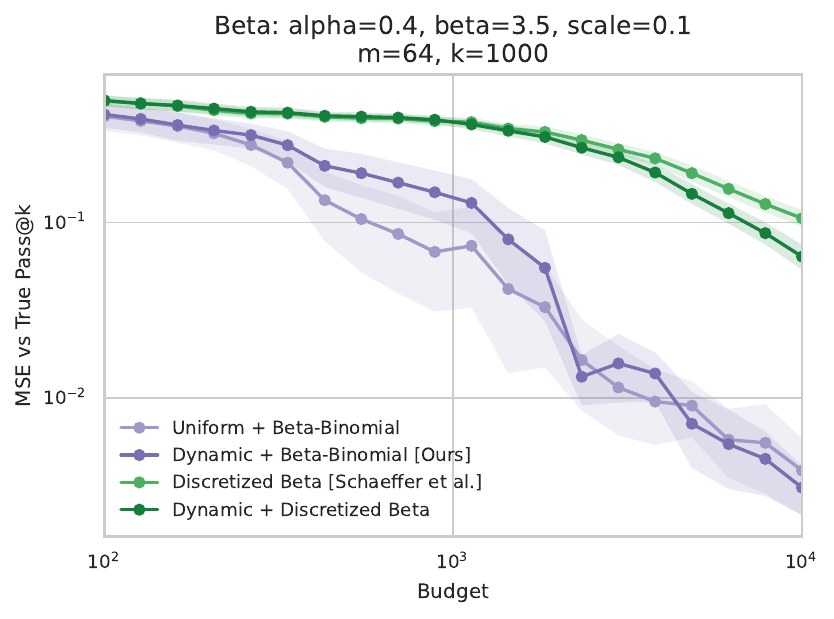} &
        \includegraphics[width=0.47\linewidth,height=0.19\textheight,keepaspectratio]{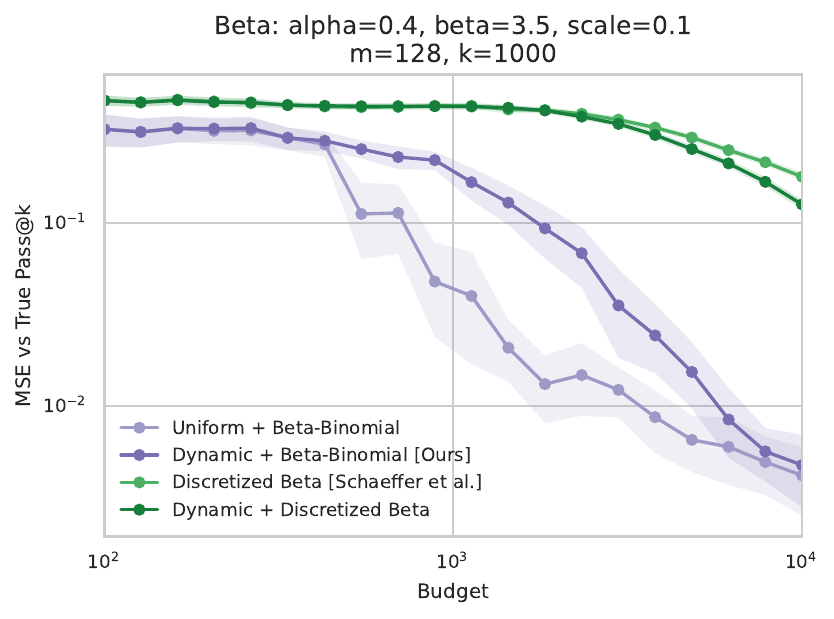} \\[2pt]
        \includegraphics[width=0.47\linewidth,height=0.19\textheight,keepaspectratio]{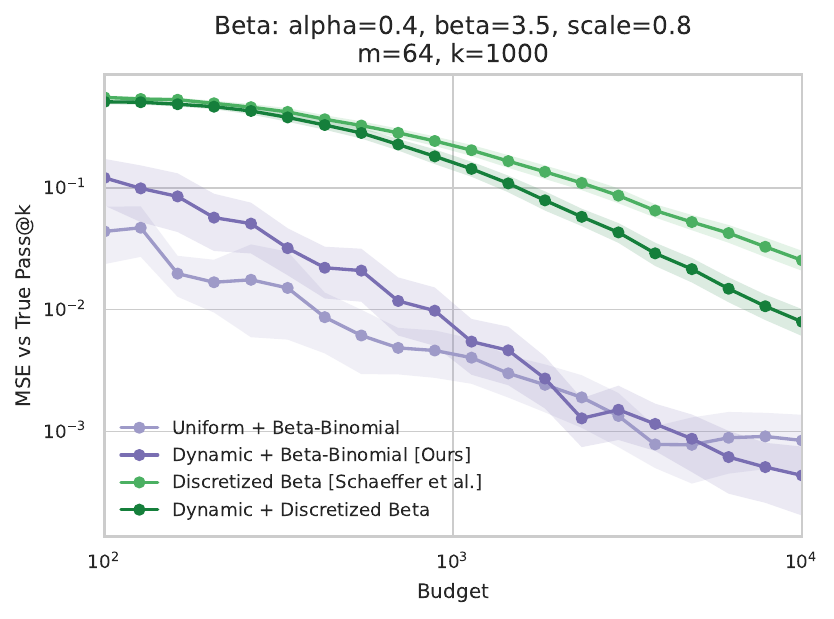} &
        \includegraphics[width=0.47\linewidth,height=0.19\textheight,keepaspectratio]{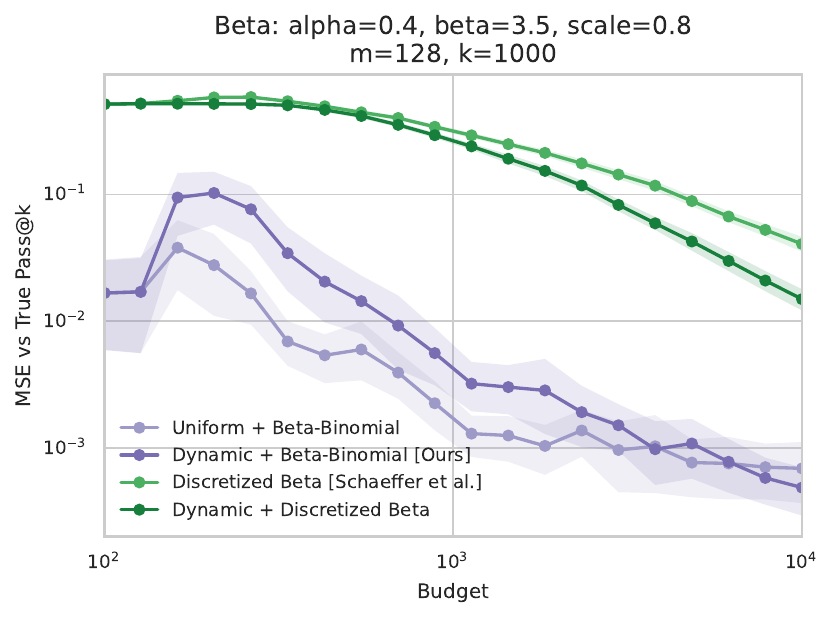} \\[2pt]
        \includegraphics[width=0.47\linewidth,height=0.19\textheight,keepaspectratio]{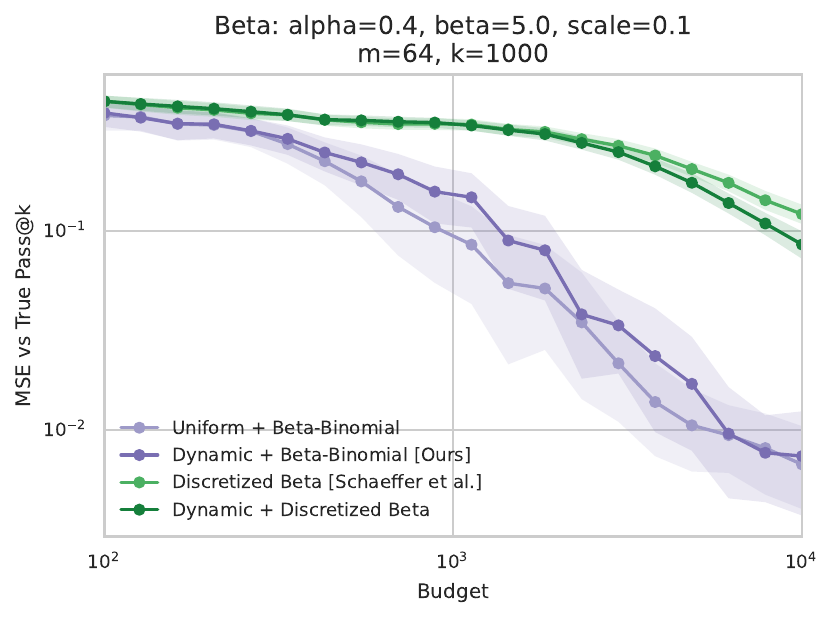} &
        \includegraphics[width=0.47\linewidth,height=0.19\textheight,keepaspectratio]{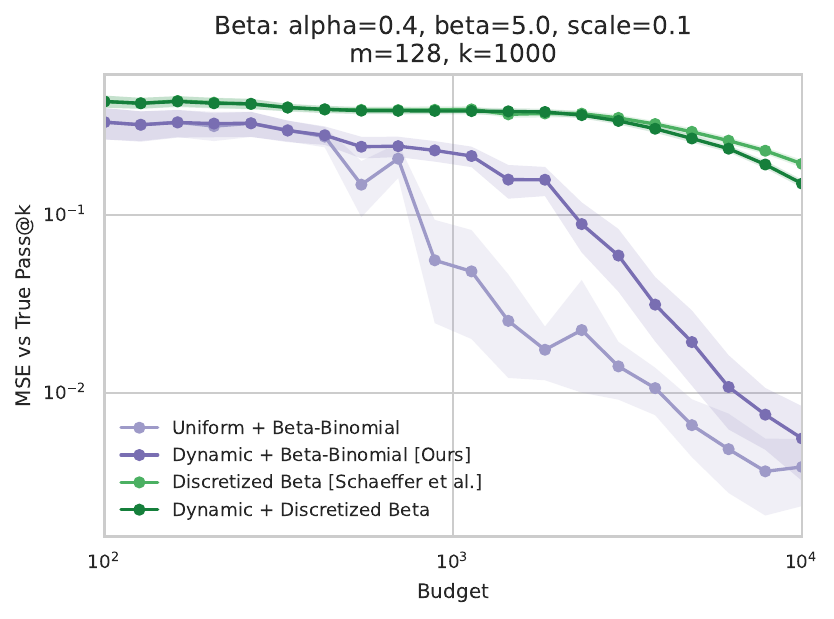} \\[2pt]
        \includegraphics[width=0.47\linewidth,height=0.19\textheight,keepaspectratio]{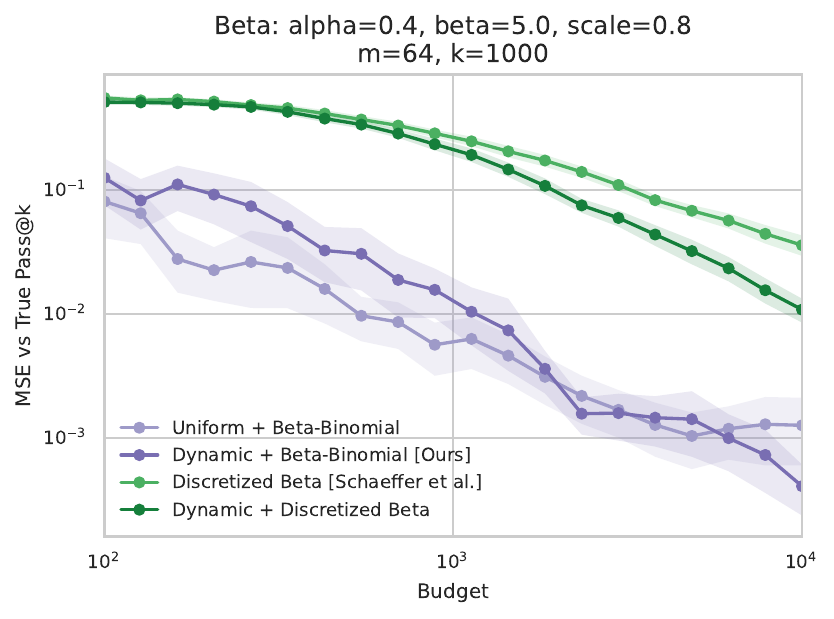} &
        \includegraphics[width=0.47\linewidth,height=0.19\textheight,keepaspectratio]{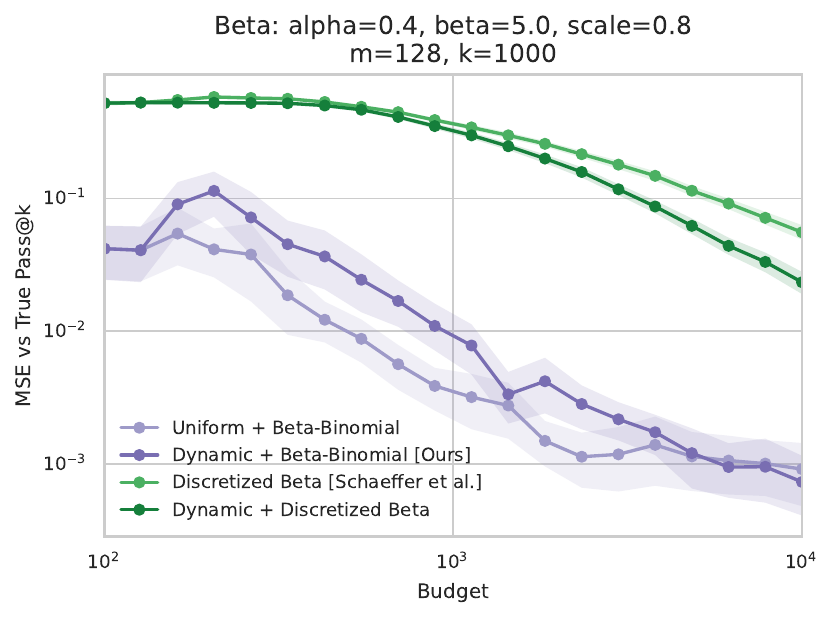} \\[2pt]
    \end{tabular}
    \caption{MSE comparisons for scaled Beta success probabilities with $a=0.4$ at $k=1000$. Columns: $m=64$ and $m=128$ problems. Rows, from top to bottom: $(b,s)=(3.5,0.1)$, $(3.5,0.8)$, $(5.0,0.1)$, and $(5.0,0.8)$.}
    \label{fig:schaeffer-beta-a04}
\end{figure}
\clearpage

\begin{figure}[!htbp]
    \centering
    \textbf{Scaled Kumaraswamy: $a=0.15$}\par\smallskip
    \begin{tabular}{@{}cc@{}}
        $m=64$ & $m=128$ \\[-2pt]
        \includegraphics[width=0.47\linewidth,height=0.19\textheight,keepaspectratio]{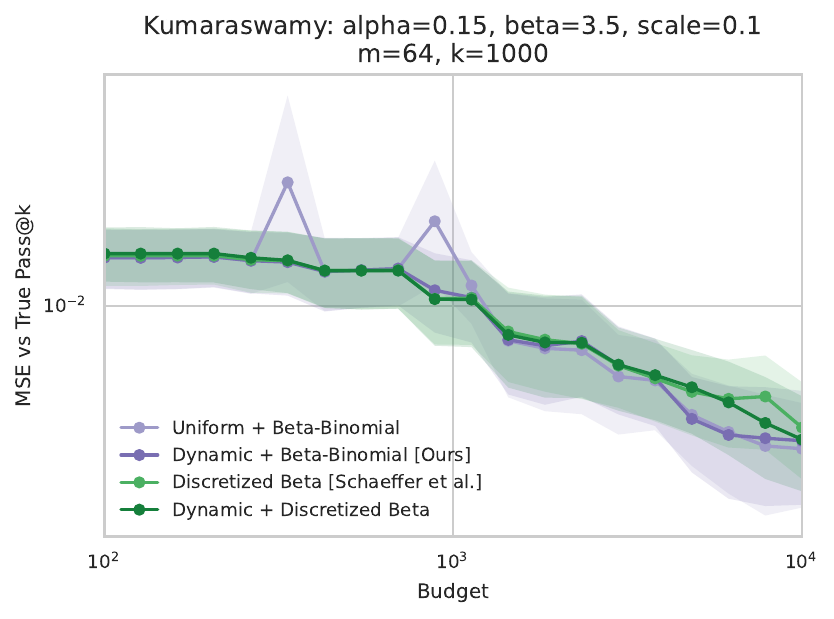} &
        \includegraphics[width=0.47\linewidth,height=0.19\textheight,keepaspectratio]{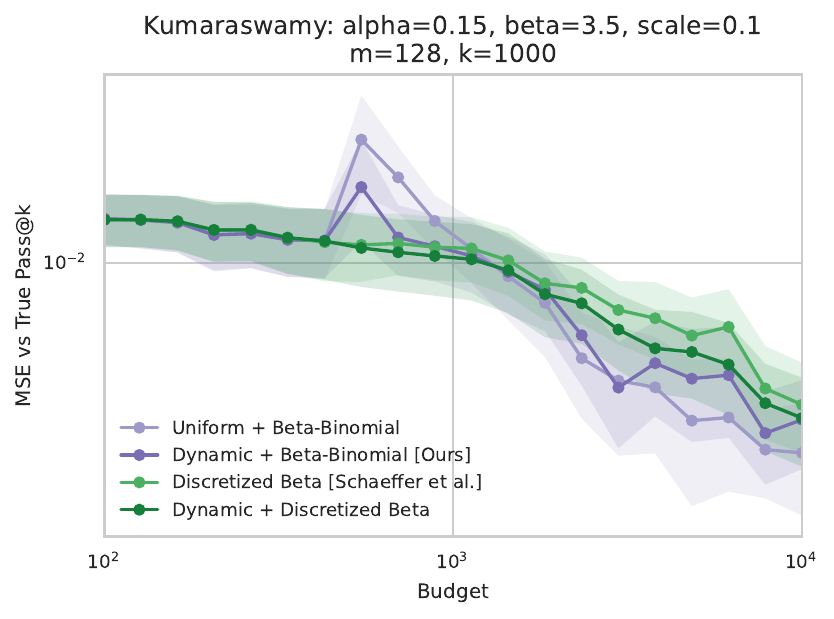} \\[2pt]
        \includegraphics[width=0.47\linewidth,height=0.19\textheight,keepaspectratio]{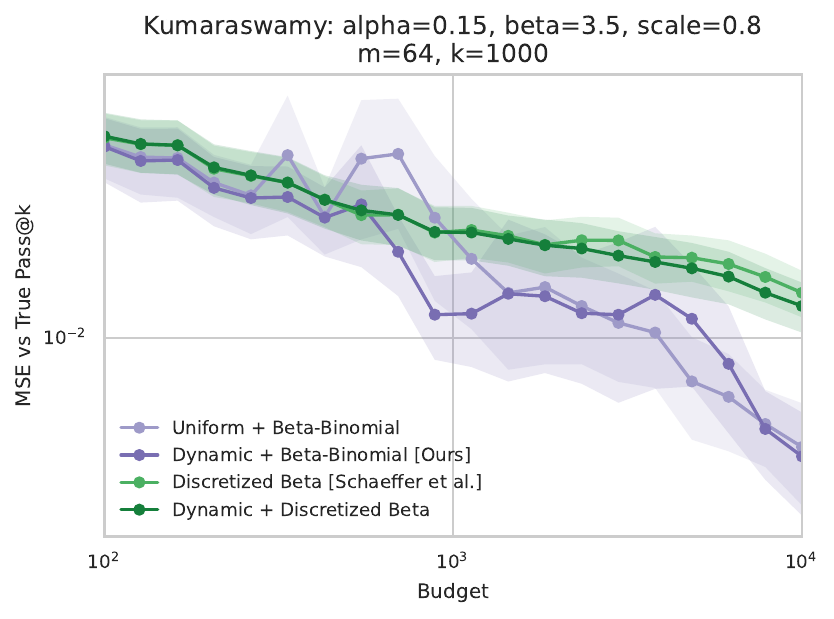} &
        \includegraphics[width=0.47\linewidth,height=0.19\textheight,keepaspectratio]{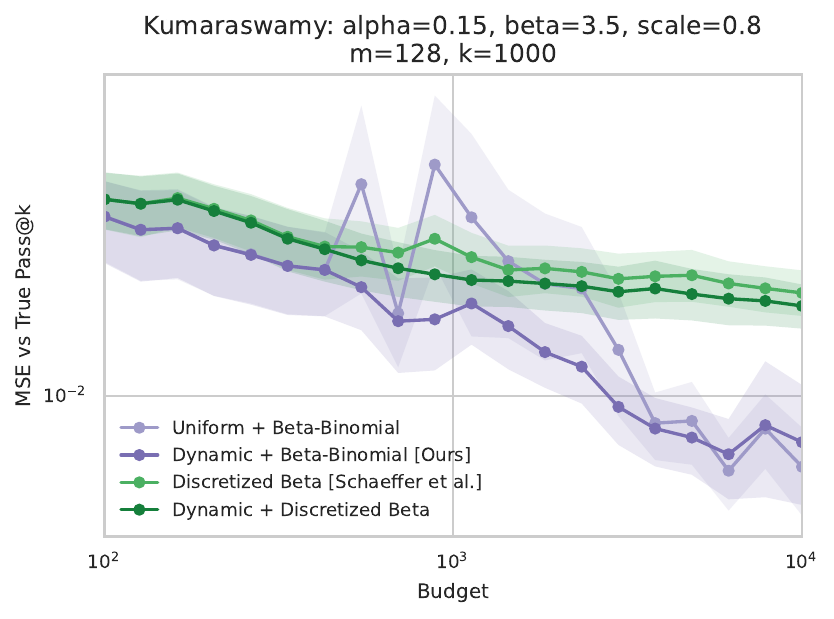} \\[2pt]
        \includegraphics[width=0.47\linewidth,height=0.19\textheight,keepaspectratio]{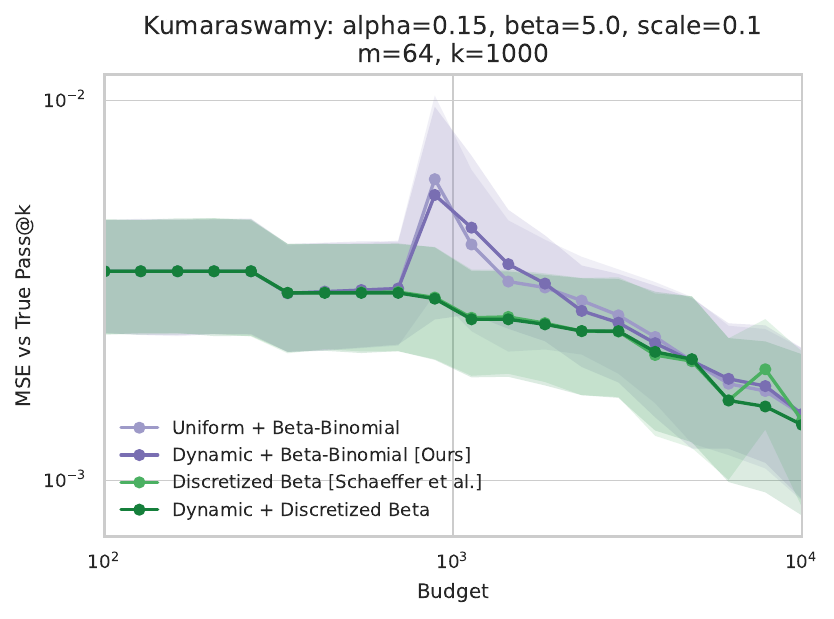} &
        \includegraphics[width=0.47\linewidth,height=0.19\textheight,keepaspectratio]{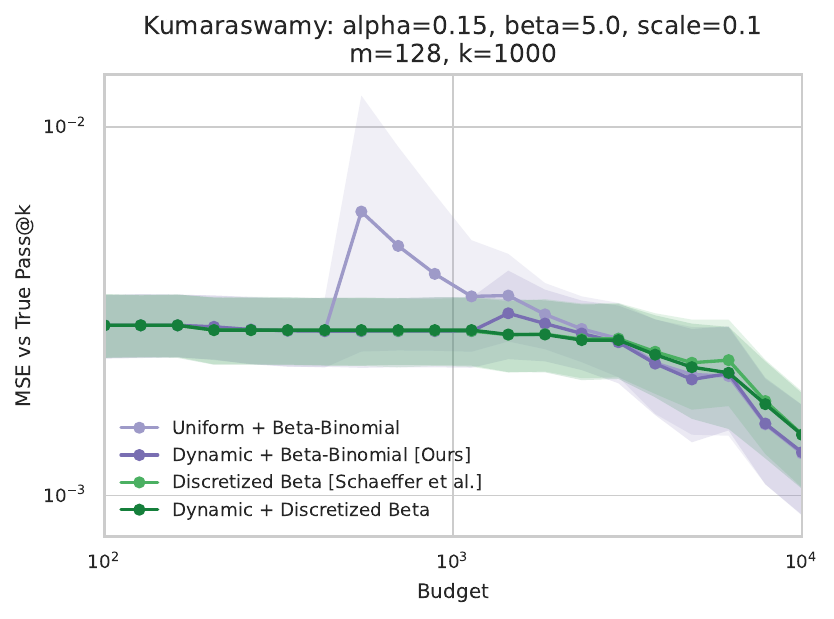} \\[2pt]
        \includegraphics[width=0.47\linewidth,height=0.19\textheight,keepaspectratio]{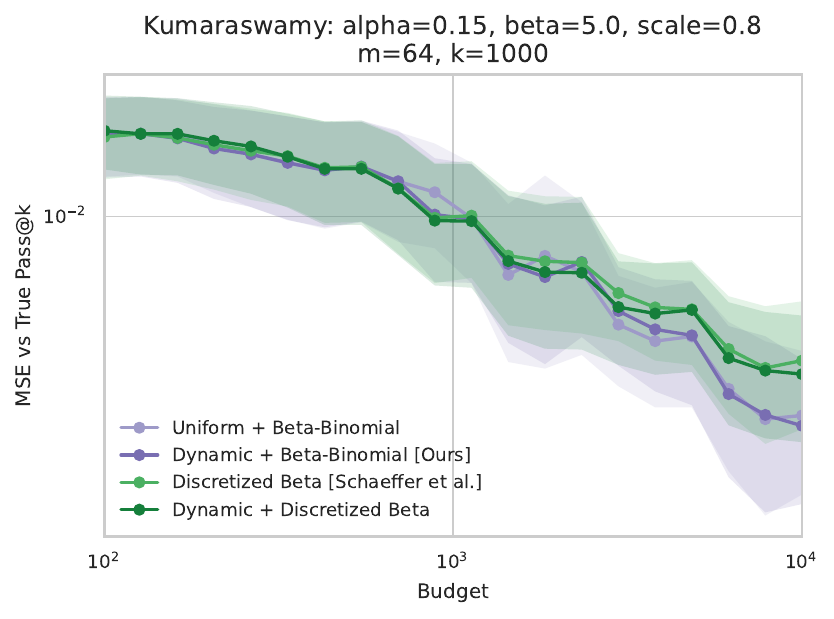} &
        \includegraphics[width=0.47\linewidth,height=0.19\textheight,keepaspectratio]{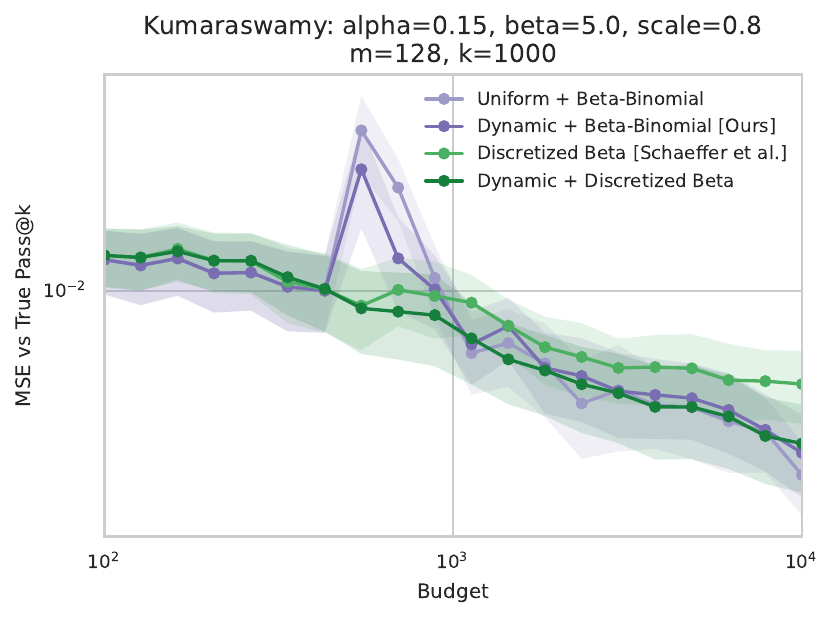} \\[2pt]
    \end{tabular}
    \caption{MSE comparisons for scaled Kumaraswamy success probabilities with $a=0.15$ at $k=1000$. Columns: $m=64$ and $m=128$ problems. Rows, from top to bottom: $(b,s)=(3.5,0.1)$, $(3.5,0.8)$, $(5.0,0.1)$, and $(5.0,0.8)$.}
    \label{fig:schaeffer-kumaraswamy-a015}
\end{figure}
\clearpage

\begin{figure}[!htbp]
    \centering
    \textbf{Scaled Kumaraswamy: $a=0.4$}\par\smallskip
    \begin{tabular}{@{}cc@{}}
        $m=64$ & $m=128$ \\[-2pt]
        \includegraphics[width=0.47\linewidth,height=0.19\textheight,keepaspectratio]{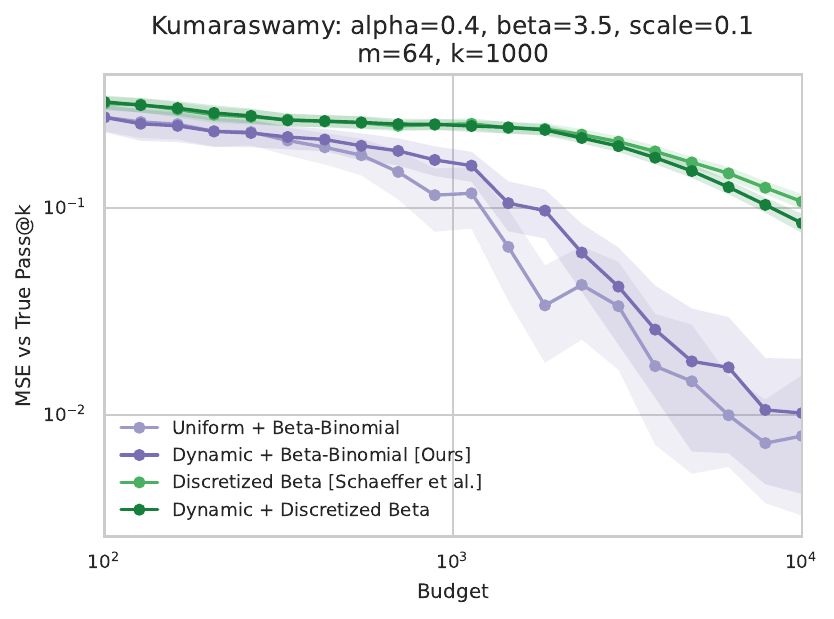} &
        \includegraphics[width=0.47\linewidth,height=0.19\textheight,keepaspectratio]{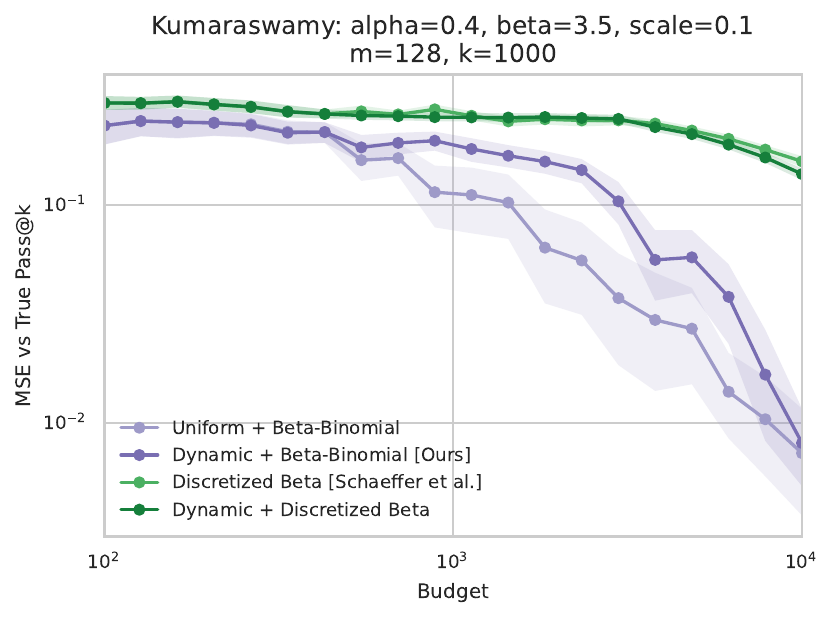} \\[2pt]
        \includegraphics[width=0.47\linewidth,height=0.19\textheight,keepaspectratio]{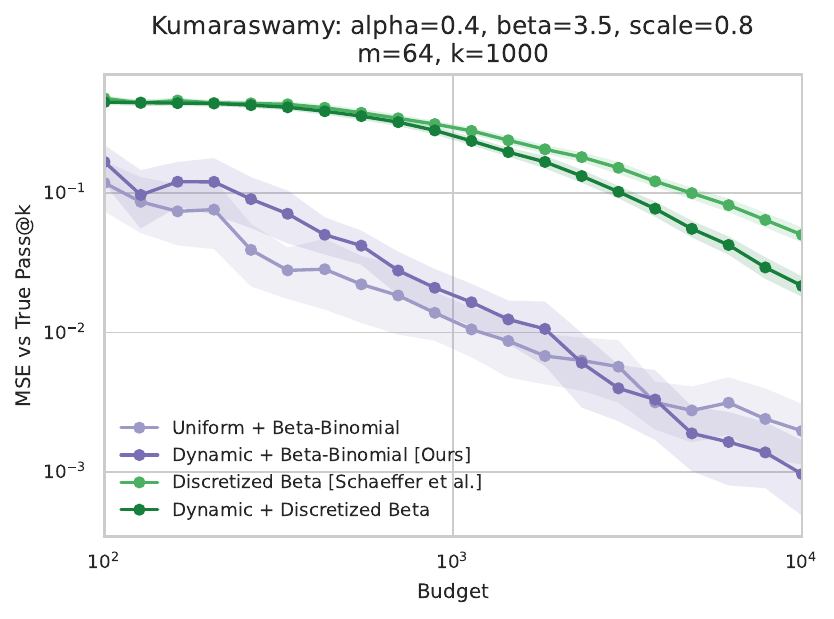} &
        \includegraphics[width=0.47\linewidth,height=0.19\textheight,keepaspectratio]{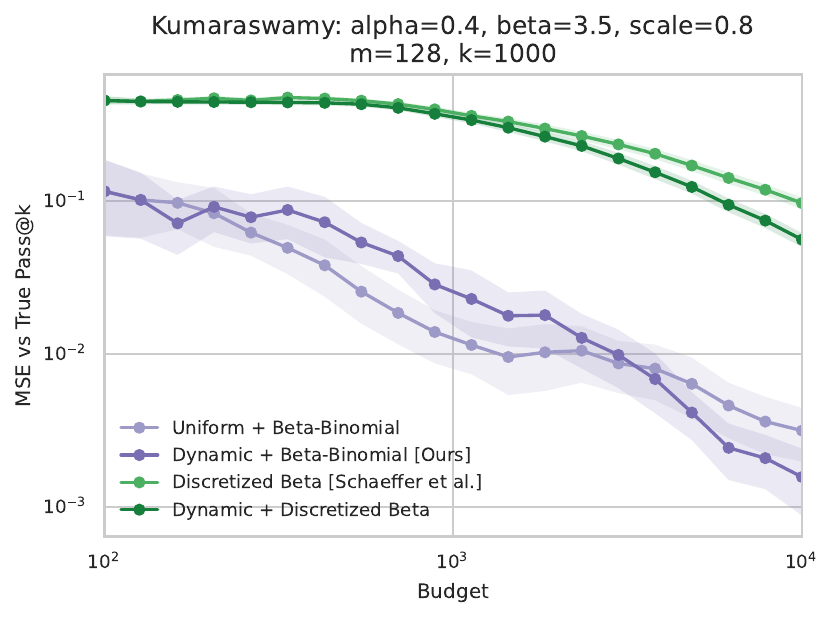} \\[2pt]
        \includegraphics[width=0.47\linewidth,height=0.19\textheight,keepaspectratio]{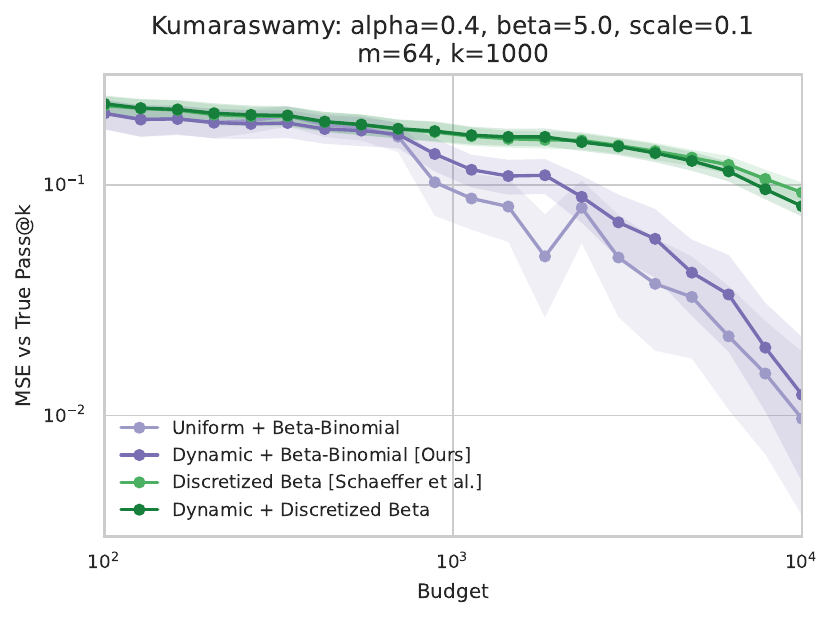} &
        \includegraphics[width=0.47\linewidth,height=0.19\textheight,keepaspectratio]{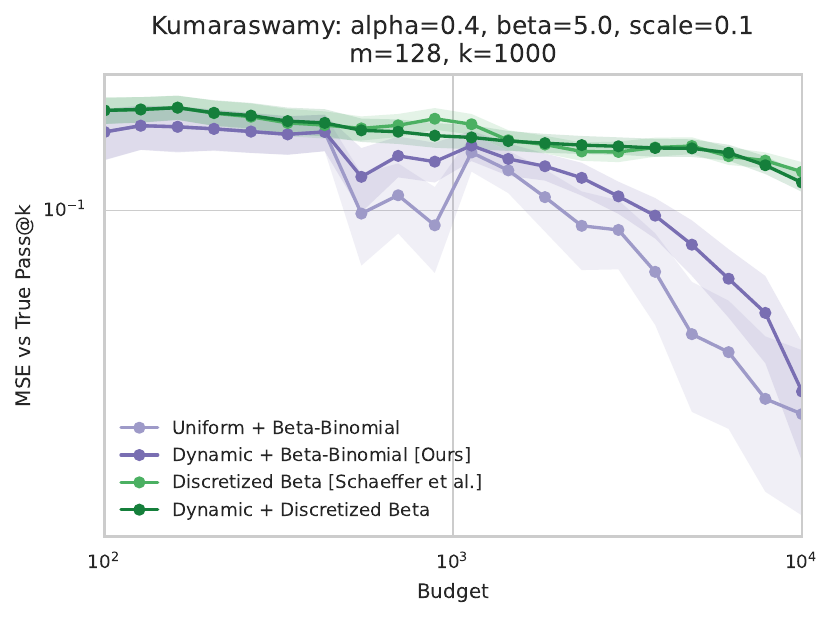} \\[2pt]
        \includegraphics[width=0.47\linewidth,height=0.19\textheight,keepaspectratio]{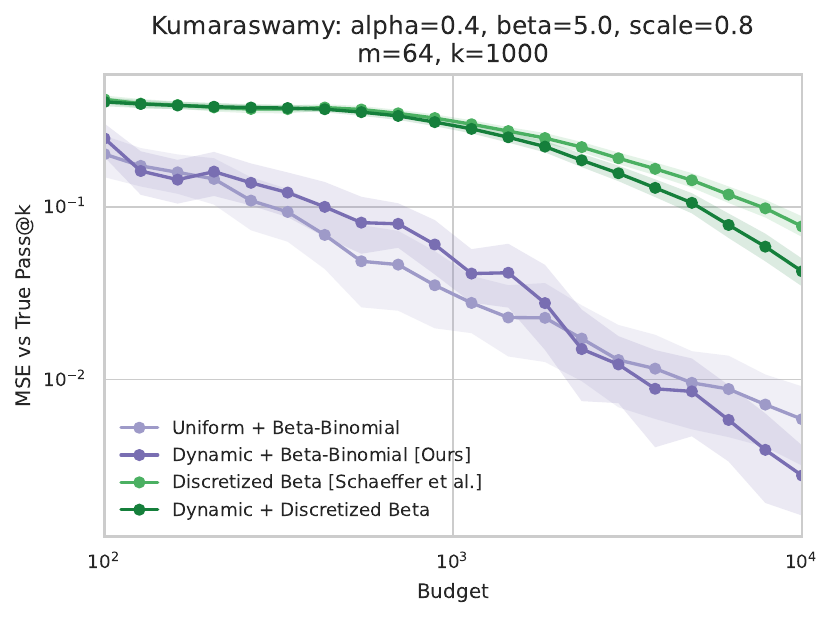} &
        \includegraphics[width=0.47\linewidth,height=0.19\textheight,keepaspectratio]{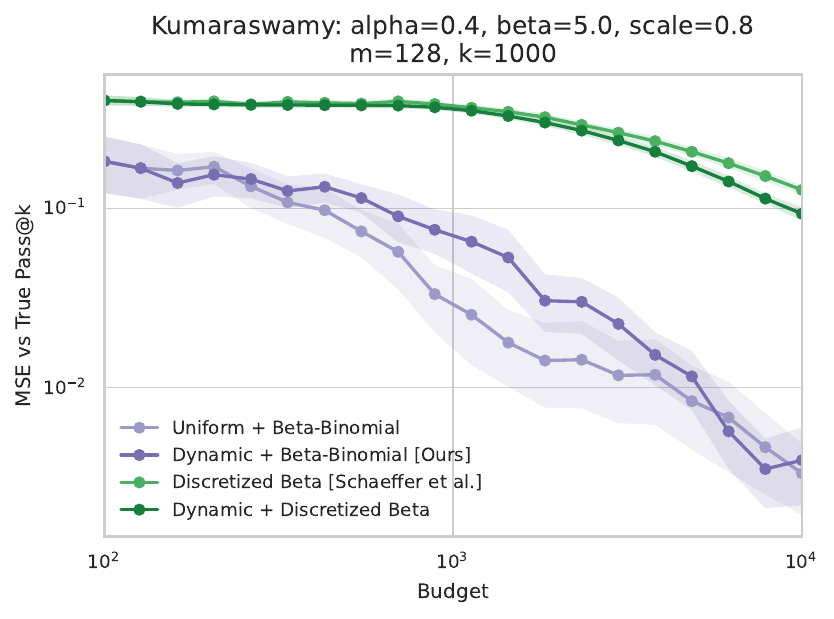} \\[2pt]
    \end{tabular}
    \caption{MSE comparisons for scaled Kumaraswamy success probabilities with $a=0.4$ at $k=1000$. Columns: $m=64$ and $m=128$ problems. Rows, from top to bottom: $(b,s)=(3.5,0.1)$, $(3.5,0.8)$, $(5.0,0.1)$, and $(5.0,0.8)$.}
    \label{fig:schaeffer-kumaraswamy-a04}
\end{figure}
\clearpage


\newpage
\section*{NeurIPS Paper Checklist}

\begin{enumerate}

\item {\bf Claims}
    \item[] Question: Do the main claims made in the abstract and introduction accurately reflect the paper's contributions and scope?
    \item[] Answer: \answerYes{} 
    \item[] Justification: As claimed, we test our proposed $\passat{k}$ extrapolation in the sample budget range $[10^1, 10^3]$ and show capacity to predict well across orders of magnitude. 

\item {\bf Limitations}
    \item[] Question: Does the paper discuss the limitations of the work performed by the authors?
    \item[] Answer: \answerYes{} 
    \item[] Justification: We include a limitations section in the appendix.

\item {\bf Theory assumptions and proofs}
    \item[] Question: For each theoretical result, does the paper provide the full set of assumptions and a complete (and correct) proof?
    \item[] Answer: \answerYes{} 
    \item[] Justification: Yes. Proofs with assumptions are included in the appendix.

    \item {\bf Experimental result reproducibility}
    \item[] Question: Does the paper fully disclose all the information needed to reproduce the main experimental results of the paper to the extent that it affects the main claims and/or conclusions of the paper (regardless of whether the code and data are provided or not)?
    \item[] Answer: \answerYes{}{} 
    \item[] Justification: Yes, we provide a python package with a one-line script to reproduce results as well as an environment specified by a Docker image.

\item {\bf Open access to data and code}
    \item[] Question: Does the paper provide open access to the data and code, with sufficient instructions to faithfully reproduce the main experimental results, as described in supplemental material?
    \item[] Answer: \answerYes{}{} 
    \item[] Justification: Yes, see previous.

\item {\bf Experimental setting/details}
    \item[] Question: Does the paper specify all the training and test details (e.g., data splits, hyperparameters, how they were chosen, type of optimizer) necessary to understand the results?
    \item[] Answer: \answerYes{}{} 
    \item[] Justification: Important details are provided in the paper. The remainder necessary to reproduce results are present in the provided code. 

\item {\bf Experiment statistical significance}
    \item[] Question: Does the paper report error bars suitably and correctly defined or other appropriate information about the statistical significance of the experiments?
    \item[] Answer: \answerYes{} 
    \item[] Justification: Yes. Error bars are provided for each of the non-synthetic $\passat{k}$ predictions.

\item {\bf Experiments compute resources}
    \item[] Question: For each experiment, does the paper provide sufficient information on the computer resources (type of compute workers, memory, time of execution) needed to reproduce the experiments?
    \item[] Answer: \answerYes{} 
    \item[] Justification: The experiments require 2-4 hours to complete. This is specified in the code. Gathering samples requires much greater time and compute, but we provide this as part of the package.
    
\item {\bf Code of ethics}
    \item[] Question: Does the research conducted in the paper conform, in every respect, with the NeurIPS Code of Ethics \url{https://neurips.cc/public/EthicsGuidelines}?
    \item[] Answer: \answerYes{} 
    \item[] Justification: We have reviewed the code of ethics to confirm.

\item {\bf Broader impacts}
    \item[] Question: Does the paper discuss both potential positive societal impacts and negative societal impacts of the work performed?
    \item[] Answer: \answerYes{} 
    \item[] Justification: We provide an impacts statement in the appendix and highlight the importance of extrapolated repeated sampling performance throughout the paper.
    
\item {\bf Safeguards}
    \item[] Question: Does the paper describe safeguards that have been put in place for responsible release of data or models that have a high risk for misuse (e.g., pre-trained language models, image generators, or scraped datasets)?
    \item[] Answer: \answerNA{} 
    \item[] Justification: The paper poses no such risks.

\item {\bf Licenses for existing assets}
    \item[] Question: Are the creators or original owners of assets (e.g., code, data, models), used in the paper, properly credited and are the license and terms of use explicitly mentioned and properly respected?
    \item[] Answer: \answerNA{}
    \item[] Justification: We gathered data and designed models ourselves.

\item {\bf New assets}
    \item[] Question: Are new assets introduced in the paper well documented and is the documentation provided alongside the assets?
    \item[] Answer: \answerYes{}{} 
    \item[] Justification: Yes. The code is well-organized, documented, and legible.

\item {\bf Crowdsourcing and research with human subjects}
    \item[] Question: For crowdsourcing experiments and research with human subjects, does the paper include the full text of instructions given to participants and screenshots, if applicable, as well as details about compensation (if any)? 
    \item[] Answer: \answerNA{} 
    \item[] Justification: We do not use crowdsourcing or perform research with human subjects.

\item {\bf Institutional review board (IRB) approvals or equivalent for research with human subjects}
    \item[] Question: Does the paper describe potential risks incurred by study participants, whether such risks were disclosed to the subjects, and whether Institutional Review Board (IRB) approvals (or an equivalent approval/review based on the requirements of your country or institution) were obtained?
    \item[] Answer: \answerNA{} 
    \item[] Justification: We do not use crowdsourcing or perform research with human subjects.

\item {\bf Declaration of LLM usage}
    \item[] Question: Does the paper describe the usage of LLMs if it is an important, original, or non-standard component of the core methods in this research? Note that if the LLM is used only for writing, editing, or formatting purposes and does \emph{not} impact the core methodology, scientific rigor, or originality of the research, declaration is not required.
    \item[] Answer: \answerNA{} 
    \item[] Justification: The core method development in this research does not involve LLMs.

\end{enumerate}

\end{document}